%% file: latex/acl_latex.tex
\documentclass[11pt]{article}

\usepackage[final]{acl}

\usepackage{times}
\usepackage{latexsym}

\usepackage[T1]{fontenc}

\usepackage[utf8]{inputenc}

\usepackage{microtype}

\usepackage{inconsolata}

\usepackage{graphicx}

\usepackage{booktabs}
\usepackage{subcaption}
\usepackage{amsmath}
\usepackage{mathrsfs}

\usepackage{amsthm}

\usepackage[capitalise,english,nameinlink]{cleveref} %
\usepackage{multirow}
\usepackage[table]{xcolor}

\newtheorem{dfn}{Definition}

\newcommand{\gc}[1]{#1}

\crefname{figure}{Figure}{Figures}
\crefname{table}{Table}{Tables}
\crefname{section}{Section}{Sections}
\crefname{equation}{Equation}{Equations}
\crefname{appendix}{Appendix}{Appendices}

\Crefname{figure}{Figure}{Figures}
\Crefname{table}{Table}{Tables}
\Crefname{section}{Section}{Sections}
\Crefname{equation}{Equation}{Equations}
\Crefname{appendix}{Appendix}{Appendices}

\usepackage{amsfonts}   %
\usepackage{amssymb}    %

\usepackage{bm}

\def\va{{\bm{a}}}
\def\vb{{\bm{b}}}

\def\vh{{\bm{h}}}

\def\vl{{\bm{l}}}

\def\vp{{\bm{p}}}

\def\vw{{\bm{w}}}

\def\vy{{\bm{y}}}

\def\mW{{\bm{W}}}

\title{Suppressing Final Layer Hidden State Jumps in Transformer Pretraining}

\author{
 \textbf{Keigo Shibata\textsuperscript{1}},
 \textbf{Kazuki Yano\textsuperscript{1}},
 \textbf{Ryosuke Takahashi\textsuperscript{1,2}},
 \textbf{Jaesung Lee\textsuperscript{1}},
\\
 \textbf{Wataru Ikeda\textsuperscript{1}},
 \textbf{Jun Suzuki\textsuperscript{1,2,3}}.
\\
 \textsuperscript{1}Tohoku University,
 \textsuperscript{2}RIKEN,
 \textsuperscript{3}NII LLMC,
\\
\texttt{is-failab-research@grp.tohoku.ac.jp}
}

\begin{document}
\maketitle
\begin{abstract}
This paper discusses the internal behavior of Transformer language models.
Many recent pre-trained models have been reported to exhibit only slight changes in the angular distance between the input and output hidden state vectors in the middle Transformer layers, despite a disproportionately large ``jump'' in the angular distance occurring in or around the final Transformer layer.
To characterize this, we first introduce a quantitative metric for the jump strength around the final layer, and then demonstrate its prevalence across many open-weight models, as well as its amplification throughout pre-training.
Assuming such jumps indicate an undesirable property, we propose the jump-suppressing regularizer (JREG) which penalizes this jump during pre-training, thereby encouraging more balanced capability usage across the middle layers.
Empirical evaluations of three model sizes of Llama-based models, trained with the proposed JREG method, reveal improved task performance compared to the baseline without altering the model architecture.
\end{abstract}

\section{Introduction}
Transformer-based language models (Transformer-LMs) have demonstrated outstanding performance across a broad spectrum of artificial intelligence (AI) tasks~\citep{achiam2023gpt, grattafiori2024llama}. 
This success has motivated research to elucidate the internal mechanisms that enable these models to generate appropriate and fluent responses to a wide range of instructions~\citep{tigges2024llm,stolfo2024confidence}. 
As a result, studies interpreting various aspects of Transformer-LMs have attracted considerable attention and emerged as one of the most vibrant areas in recent AI research~\citep{belrose2023eliciting, kobayashi2024analyzing}.

\begin{figure}[t]
    \centering
    \includegraphics[width=\linewidth]{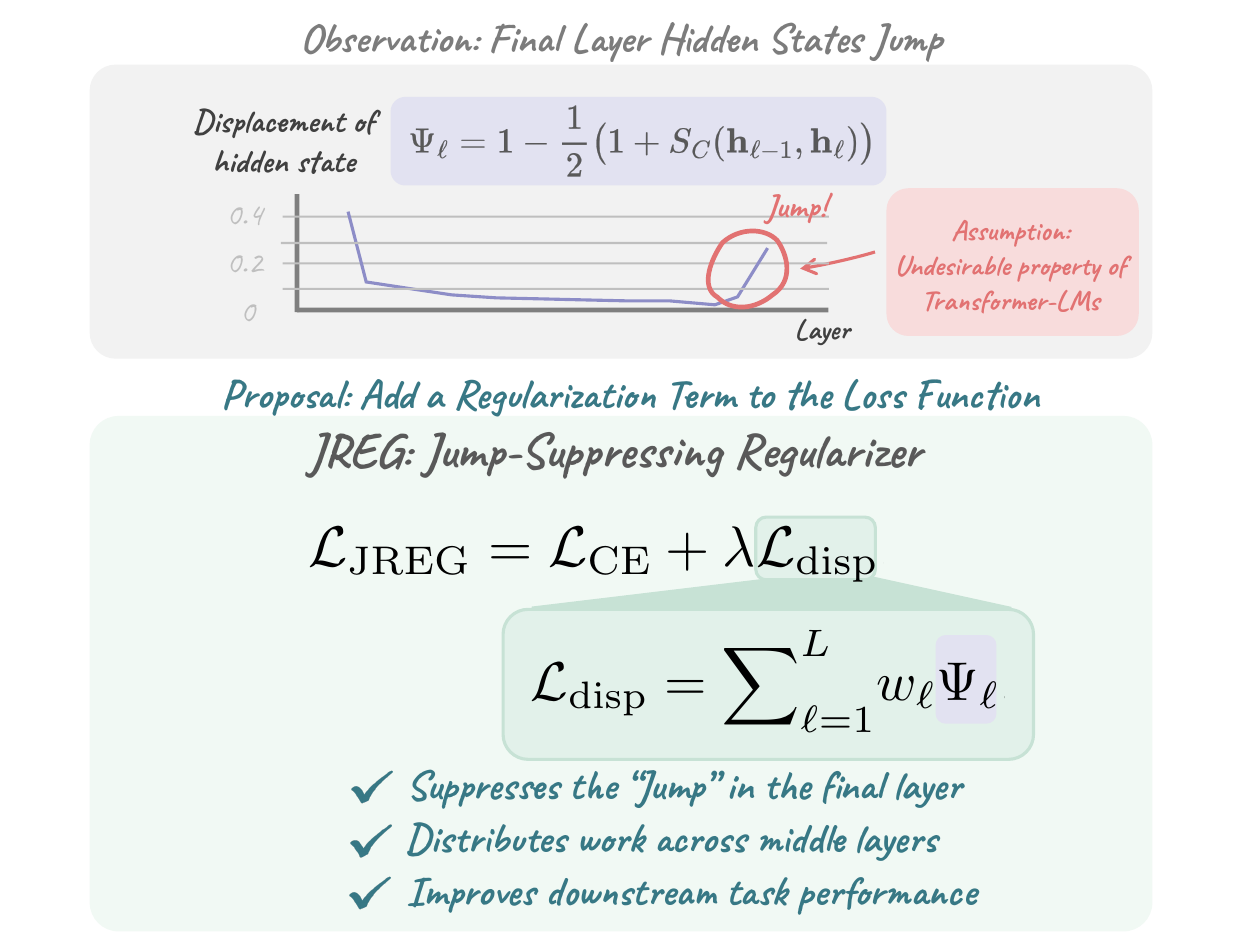}
    \caption{Many recent open-weight language models exhibit minimal hidden state displacement in their middle layers, but exhibit a pronounced “jump” at their final layer. Suppressing this jump during pre-training could improve overall performance by fostering a more balanced use of capabilities across the middle layers.}
    \label{fig:1} 
\end{figure}

Regarding the model structure of Transformer-LMs, many studies have reported the existence of redundant and ineffective parameters in pre-trained LMs~\citep{DBLP:journals/corr/abs-2407-15516, DBLP:journals/corr/abs-2403-03853,DBLP:conf/aaai/SunPNJ25}, while other studies have suggested that such redundancy plays a critical role in enabling effective and efficient pre-training~\citep{aghajanyan-etal-2021-intrinsic,song2024unraveling,DBLP:journals/corr/abs-2406-19384}.
In particular, pre-trained Llama models~\citep{grattafiori2024llama} have been observed to contain many redundant and ineffective Transformer layers in the middle of the models~\citep{DBLP:conf/aaai/SunPNJ25}. 
Specifically, this observation can be rewritten to state that these middle layers produce output hidden state vectors that are highly similar to their input vectors in terms of angular distance.
In contrast to such behavior in the middle layers, several studies~\citep{DBLP:conf/aaai/SunPNJ25, DBLP:journals/corr/abs-2407-15516} have also recently observed and reported the ``jump'' behavior, which we define as a pronounced large change in angular distance, occurring in (and, in some cases, around) the final layer(s).
We hypothesize that a small angular distance in the middle layers leads the final layer to exhibit a disproportionately large angular distance change because the relative representational load on those middle layers is reduced, causing the final layer to bear more of the workload to compensate for their less effective contribution.
This imbalance in layer contributions likely limits the overall capacity of the model and increases parameter redundancy.

This paper focuses on investigating this jump behavior.
\cref{fig:1} shows an overview of the study presented in this paper.
We first introduce a metric to quantitatively compare the jump strength in and around the final layers (\cref{Difinition_of_Displacement_and_Jump_rate}).
We then demonstrate that such a jump behavior is often observed in many open-weight pre-trained models, including well-known and widely used ones (\cref{Analysis of open-weight models}), and tends to become increasingly pronounced as pre-training progresses (\cref{Formation condition of the jump}).
We assume that this jump indicates an undesirable property of pre-trained models and thus propose a regularizer, referred to as \textbf{Jump-Suppressing Regularizer (JREG)}, that strongly penalizes the emergence of the jumps in and around the final layer during pre-training (\cref{Proposed Method}), and indirectly encourages more balanced usage of the model's capabilities across the middle layers.

To empirically investigate our research question, that is, \emph{whether mitigating the jump behavior in and around the final layer of Transformer-LMs and encouraging more even utilization across the middle layers can improve overall model capacity and leads to performance improvements}, we conduct experiments on three sizes of Llama‐based architectures (170M, 1B and 3.4B parameters) under two pre-training regimes: (1) using the standard cross-entropy loss only, which serves as the baseline for comparison, and (2) using the standard cross-entropy loss combined with JREG.
We evaluate each model based on both (a) the jumping behaviors around the final layers and (b) performance on widely used downstream tasks.
The empirical results and our analyses reveal that incorporating JREG can improve downstream task performance and enhance the capabilities of models without altering the model architecture, by strongly penalizing jump behaviors around the final layers, which also mitigates ineffective middle layers in Transformer-LMs.

\section{Related Work}
\paragraph{Layer redundancy.}
Transformer-LMs often exhibit layer redundancy, where multiple layers may learn similar or redundant operations.
Models like Universal Transformer~\citep{DBLP:conf/iclr/DehghaniGVUK19} and ALBERT~\citep{DBLP:conf/iclr/LanCGGSS20}, which share parameters across layers, achieve high parameter efficiency.
The linear relationship between the layer input and the output vectors~\citep{DBLP:conf/acl/RazzhigaevMGGOD24}, as well as the small angular distance between layers~\citep{DBLP:conf/aaai/SunPNJ25, DBLP:journals/corr/abs-2407-15516}, indicate that pre-trained models exhibit layer redundancy.
These findings substantiate the hypothesis that Transformer layers can learn nearly identical operations. 
The location of these redundant layers appears to be architecture dependent. 
For example, redundancy may appear in shallower layers in some models~\citep{DBLP:journals/csl/SajjadDDN23}, while in others, such as Llama and GPT, removing deeper layers (except for the final layers) causes only a slight performance degradation~\citep{DBLP:journals/corr/abs-2407-15516, DBLP:journals/corr/abs-2403-03853,DBLP:conf/aaai/SunPNJ25} as the output magnitude in Pre-LN Transformers grows with depth~\citep{DBLP:conf/icml/KediaZKJGL24, DBLP:journals/corr/abs-2304-14802, DBLP:journals/corr/abs-2502-05795}
Consequently, methods to reduce such redundancy have been proposed, including architectural modifications like Mix-LN~\citep{DBLP:journals/corr/abs-2412-13795}, approaches suppressing deep-layer hidden state norms~\citep{DBLP:journals/corr/abs-2502-05795}, and training techniques such as LayerDrop~\citep{DBLP:conf/nips/ZhangH20, DBLP:conf/iclr/FanGJ20, DBLP:conf/acl/ElhoushiSLHWL0A24} and LayerShuffle~\citep{DBLP:journals/corr/abs-2502-05795}. 
Unlike these previous approaches, this paper introduces a regularization technique that suppresses final layer displacement, which also addresses the common issue of underutilized middle layers and improves downstream task performance without changing the model architecture.

\paragraph{Early exit.}
To reduce the per‐token latency of Transformer-LMs, a large body of work equips middle layers with the ability to produce final predictions and halt computation once a satisfactory confidence threshold is met.
This idea was first popularized in convolutional nets by \textsc{BranchyNet}~\citep{DBLP:conf/icpr/Teerapittayanon16} and later inspiring a series of BERT‐style architectures such as \textsc{PABEE}~\citep{DBLP:conf/nips/ZhouXGM0W20}, \textsc{DeeBERT}~\citep{DBLP:conf/acl/XinTLYL20}, and \textsc{FastBERT}~\citep{DBLP:conf/acl/LiuZWZDJ20}.
These methods attach lightweight classifiers to each layer, train them with knowledge distillation to approximate the final output, and decide at run time whether to exit early according to some criteria.
For autoregressive language models, early exit must preserve sequence consistency and hidden state caching.
\textsc{CALM}~\citep{DBLP:conf/nips/SchusterFG0B0TM22} introduces confidence adaptive halting.
More recently, \textsc{LayerSkip}~\citep{DBLP:conf/acl/ElhoushiSLHWL0A24} reuses the original LM head at every layer, whereas~\citet {DBLP:conf/iclr/JiangZZ25} adds separate linear heads; both apply per-layer cross-entropy losses.
Whereas early exit techniques aim to maximise middle layers' performance to match that of the final layer, our study focuses on the excessive jump observed in the final layer of many open-weight models.

\paragraph{Internal trajectory.}
Analyzing the trajectories of hidden state vectors at each Transformer-LM layer and their projections into the vocabulary space is important from both a fundamental perspective, namely, understanding internal mechanisms, and an applied perspective, that is, improving performance and ensuring safety.
By projecting each layer’s hidden states into the vocabulary space, the inference process inside a Transformer-LM can be visualized, and it has been observed that the outputs of the middle layers gradually converge toward the final layer’s predicted word~\citep{belrose2023eliciting, DBLP:conf/acl/ElhoushiSLHWL0A24}.
Instead of relying solely on the final layer at inference time, extrapolating the token probability trends over the last few layers to obtain a more mature predictive distribution has been shown to mitigate hallucinations~\citep{DBLP:conf/coling/DasJSMPY25}.
It has been reported that models with backdoors exhibit unnatural trajectories of hidden state vectors, and that applying a small amount of fine-tuning to correct these trajectories can neutralize the backdoor~\citep{DBLP:journals/corr/abs-2411-12768}.
Taken together, internal trajectories are important targets for both basic and applied research, and their analysis and control can contribute to improving factuality and ensuring safety.
Accordingly, we address the characteristic “jump” observed in the final layer hidden states of pre-trained models through a trajectory regularizing objective.

\section{Final Layer Hidden State Jump}
\label{Final layer displacement jump}

\begin{figure*}[t]
    \centering
    \begin{subfigure}[b]{0.49\linewidth}
       \centering
       \includegraphics[width=\linewidth]{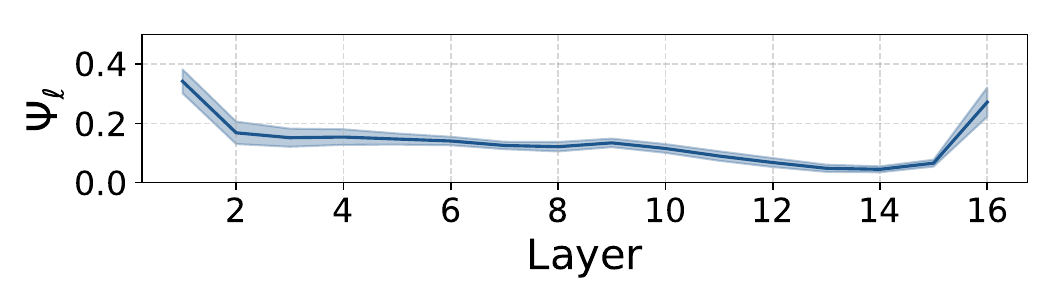}
       \caption{Llama-3.2-1B}
       \label{fig:llama3.2-1B}
   \end{subfigure}
    \begin{subfigure}[b]{0.49\linewidth}
        \centering
        \includegraphics[width=\linewidth]{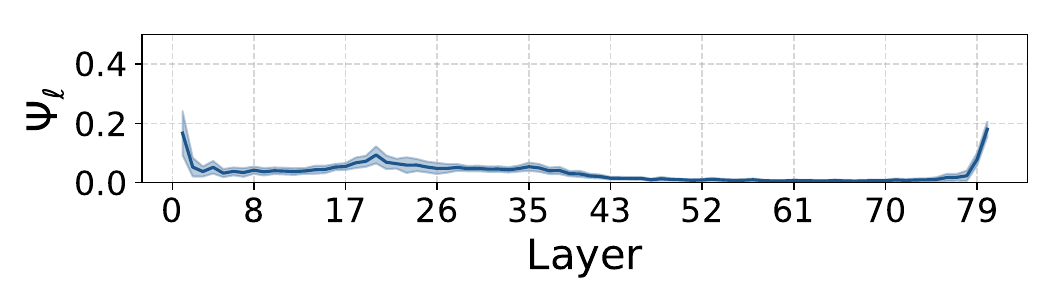}
        \caption{Llama-3.2-70B}
        \label{fig:llama3.2-3B}
    \end{subfigure}
     \hfill
    \begin{subfigure}[b]{0.49\linewidth}
        \centering
        \includegraphics[width=\linewidth]{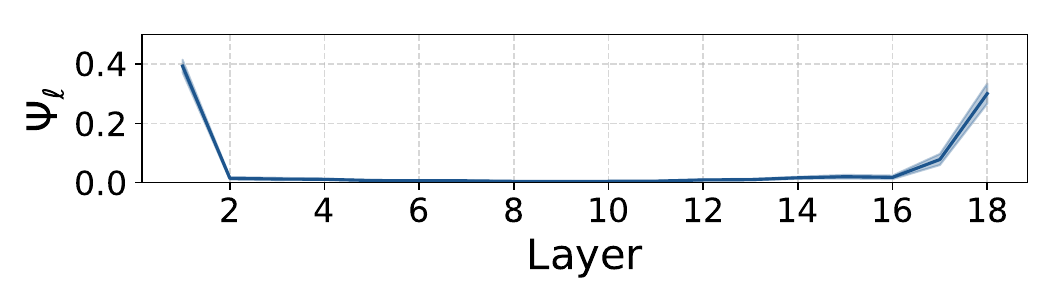}
        \caption{Gemma-2B}
        \label{fig:gemma-2b}
    \end{subfigure}
    \hfill
    \begin{subfigure}[b]{0.49\linewidth}
        \centering
        \includegraphics[width=\linewidth]{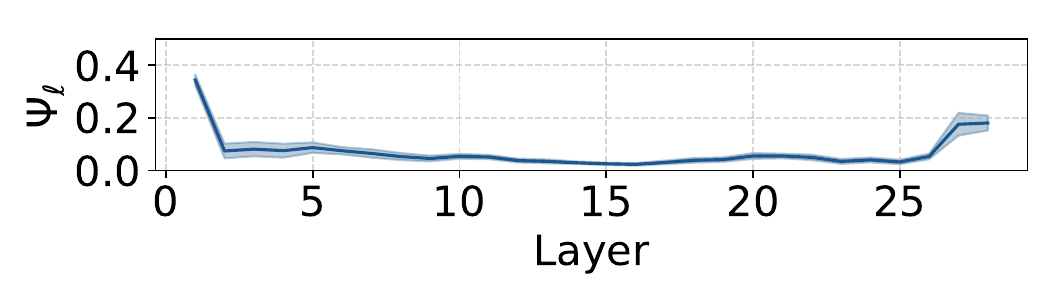}
        \caption{Gemma-7B}
    \end{subfigure}
    \begin{subfigure}[b]{0.49\linewidth}
        \centering
        \includegraphics[width=\linewidth]{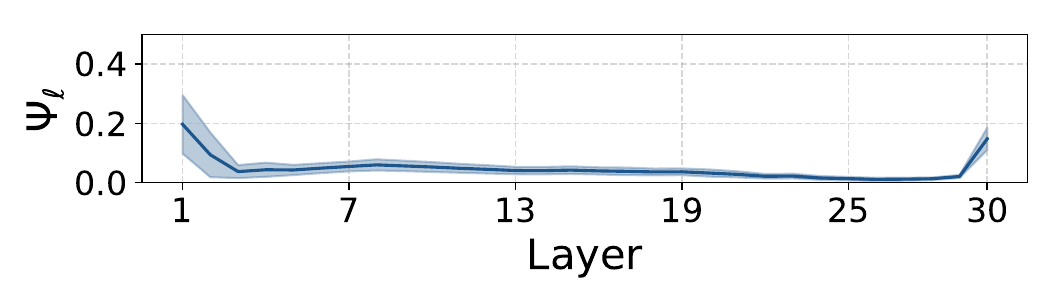}
        \caption{DeepSeek-7B}
    \end{subfigure}
    \hfill
    \begin{subfigure}[b]{0.49\linewidth}
        \centering 
        \includegraphics[width=\linewidth]{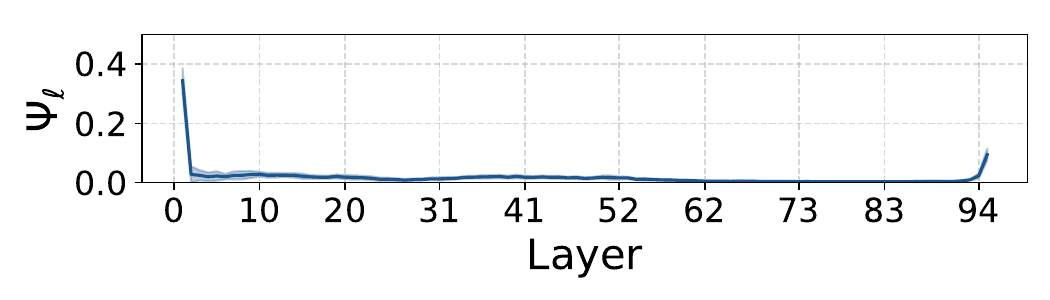}
        \caption{DeepSeek-67B}
        \label{fig:deepseek-67B}
    \end{subfigure}
    \begin{subfigure}[b]{0.49\linewidth}
        \centering
        \includegraphics[width=\linewidth]{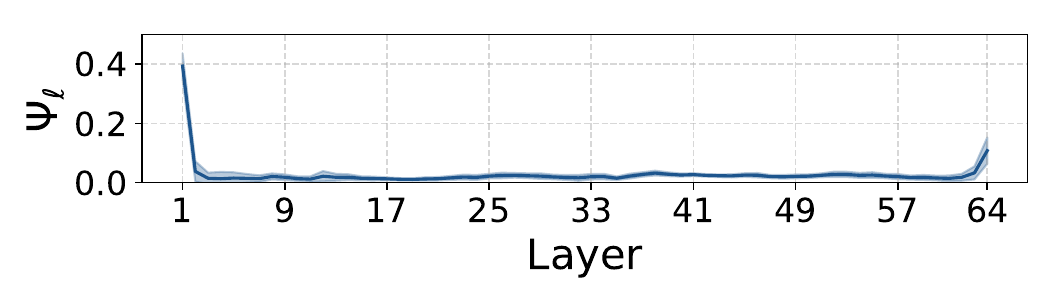}
        \caption{Qwen3-32B}
        \label{fig:qwen3_32b}
    \end{subfigure}
    \begin{subfigure}[b]{0.49\linewidth}
        \centering
        \includegraphics[width=\linewidth]{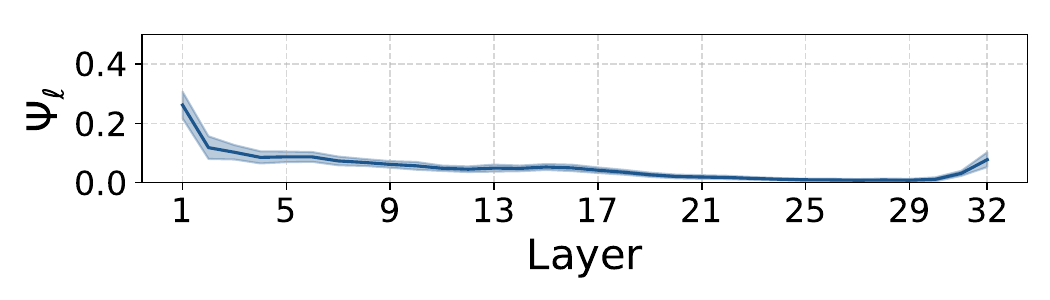}
        \caption{layerskip-llama3.2-8B}
        \label{fig:layerskip-8b}
    \end{subfigure}
    \caption{Layer-wise hidden state displacement $\Psi_\ell$ for next-word prediction on 100 samples from the LAMBADA dataset. Across all model architectures, the displacement at the final layer tends to be larger than that of the middle layers.}
    \label{fig:latest_model_hidden_trajectory}
\end{figure*}

In the following, we introduce an analytical framework to elucidate the inter-layer displacement characteristics that are consistently observed in open-weight large language models.
First, we introduce the displacement $\Psi_\ell$, a metric that quantifies the magnitude of hidden state variation in each layer, jump rate \(\zeta_{\ell}\), a metric that compares displacement magnitudes between the middle and final layers.
Using the $\Psi_\ell$ and \(\zeta_{\ell}\), this study empirically demonstrates a property common to open-weight models, namely, ``jump'', in which the displacement of the final layer markedly exceeds that of the middle layers.

\subsection{Definition of displacement and jump rate}
\label{Difinition_of_Displacement_and_Jump_rate}
We formalize the layer-wise transformation of hidden state vectors in Transformer-LMs.
Hereafter, let $\vh$ denote a $D$-dimensional vector, that is, $\vh \in \mathbb{R}^{D}$.
Let $\vh_{\ell-1}$ be the input hidden state vector of layer $\ell$, and simultaneously the output hidden state vector to layer $\ell-1$, where $L$ denote the total number of layers and the relation $1 \le \ell \le L$ holds.
We can represent the relationship between the input and output hidden state vectors as $\vh_{\ell} = \mathtt{TL}_{\ell}(\vh_{\ell-1})$, where $\mathtt{TL}_{\ell}(\cdot)$ denotes the $\ell$-th Transformer layer, which primarily consists of an attention mechanism and a feed-forward network, along with layer normalizations.
Moreover, for $\ell=0$, $\vh_0$ represents the word embedding.

We geometrically interpret hidden state transitions.
Transformer-LMs generate text via an autoregressive next‐token prediction process. 
Viewed geometrically, each token prediction traces a trajectory in the embedding space that starts from the input layer embedding, $\vh_0$, and terminates near the output layer embedding.
Building on this trajectory perspective, we characterize each step of the path by the angular change between consecutive hidden state vectors, i.e., ($\vh_0, \vh_1, \dots, \vh_L$).
High-dimensional geometry defies direct visual intuition.
To obtain a tractable scalar summary, we therefore introduce the displacement metric, which we construct in accordance with prior work~\cite{DBLP:journals/corr/abs-2407-15516,DBLP:journals/corr/abs-2412-13795,DBLP:conf/acl/SimoulinC21,DBLP:conf/eacl/GodeyCS24} by adopting cosine similarity based measures.
Let $S_C(\va, \vb)$ be the cosine similarity between two arbitrary vectors $\va$ and $\vb$, expressed as:
    \begin{equation}
          S_C(\va, \vb) = \frac{\va^\top \vb}{\|\va\|_2\,\|\vb\|_2},
    \end{equation}
where $\|\cdot\|_2$ represents the Euclidean norm of a given vector.

\begin{dfn}[Displacement between the input and output hidden state vectors]
    We define the displacement between the input and output hidden state vectors at layer $\ell$, denoted as $\Psi_\ell$, as follows:
    \begin{equation}
          \label{eq:D_l}
          \Psi_\ell = 1 - \frac12 {\big(1 + S_C(\vh_{\ell-1}, \vh_{\ell})\big)}
          .
    \end{equation}
By construction, \(\Psi_\ell \in [0, 1]\)  
and a larger \(\Psi_\ell\) indicates a greater change in the angular distance between \(\vh_{\ell-1}\) and \(\vh_{\ell}\).
\end{dfn}

Next, we quantify the jump rate \(\zeta_{\ell}\), which measures the cumulative positive increase in displacement from layer \(\ell-1\) up to the final layer, scaled by \(100\).
\begin{dfn}[Jump rate]
    For a model with \(L\) layers where $2 \leq \ell \leq L$, we define the jump rate \(\zeta_{\ell}\) as  
    \begin{equation}
      \label{eq:zeta}
      \zeta_{\ell} = {\sum}_{k=\ell}^L \max (0, \Psi_{k}-\Psi_{k-1}) \times 100
    \end{equation}
\end{dfn}
For example, $\zeta_{\ell}$ will take the minimum value of 0, i.e., $\zeta_{\ell} = 0$ if the displacement at layers $k$ and $k-1$, where $k \in \{\ell, \ldots, L\}$, satisfy the non-increasing condition (i.e., $\Psi_{k-1} \geq \Psi_{k}$).
Moreover, while the range of $\Psi_\ell$ is $[0,1]$, and most $\Psi_\ell$ values in the middle layers are smaller than 0.1, $\zeta_{\ell} \geq 10 $ are considered to be significantly larger values.

\begin{figure*}[t]
    \centering
    \begin{subfigure}[b]{0.49\linewidth}
        \centering
        \includegraphics[width=\linewidth]{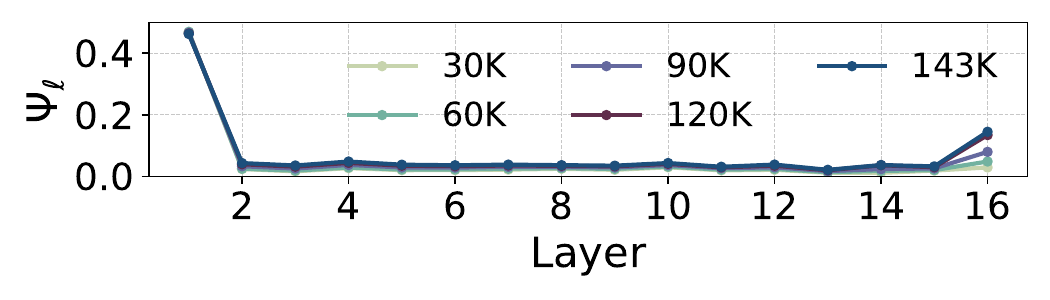}
        \caption{Pythia 1B}
        \label{fig:pythia_1b_pretrain_displacement}
    \end{subfigure}
    \hfill
    \begin{subfigure}[b]{0.49\linewidth}
        \centering
        \includegraphics[width=\linewidth]{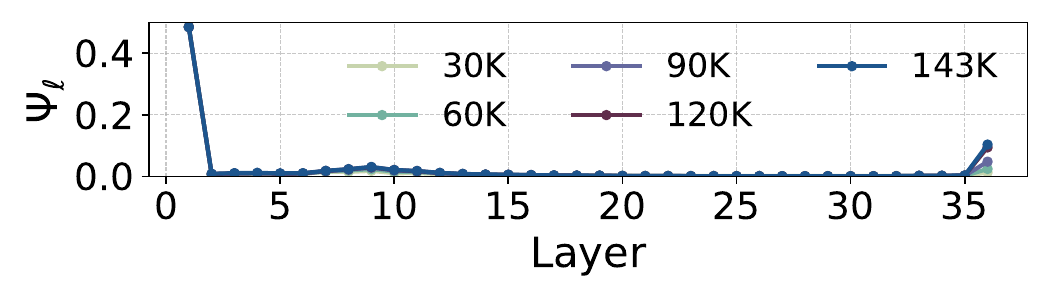}
        \caption{Pythia 12B}
        \label{fig:pythia_12b_pretrain_displacement}
    \end{subfigure}
    \begin{subfigure}[b]{0.49\linewidth}
        \centering
        \includegraphics[width=\linewidth]{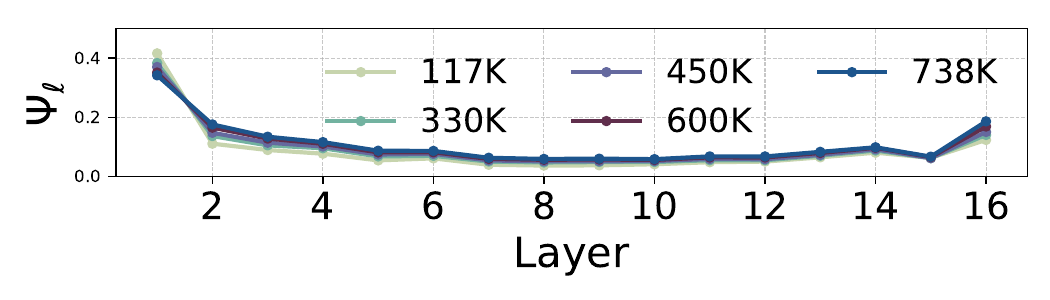}
        \caption{OLMo 1B}
        \label{fig:olmo_1b_pretrain_displacement}
    \end{subfigure}
    \hfill
    \begin{subfigure}[b]{0.49\linewidth}
        \centering
        \includegraphics[width=\linewidth]{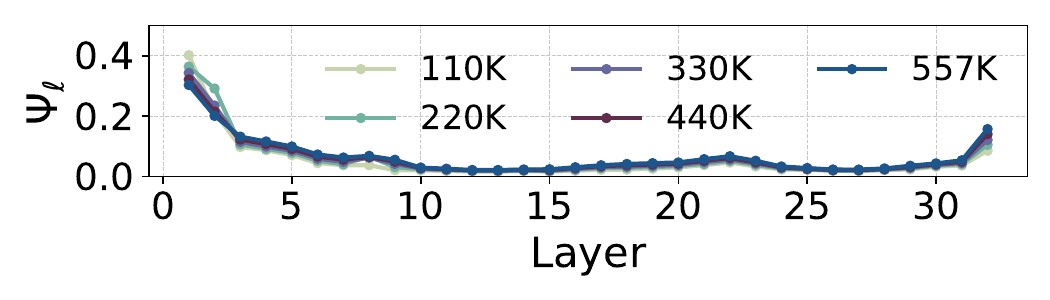}
        \caption{OLMo 7B}
        \label{fig:olmo_12b_pretrain_displacement}
    \end{subfigure}
    \caption{Analysis of checkpoint-wise hidden state displacement using fine-grained Pythia and OLMo pre-training checkpoints, revealing that as training progresses, the final layer exhibits large ``jump'' displacements.}
    \label{fig:pythia_ckpt_trajectory}
\end{figure*}

\subsection{Analysis of open-weight models}
\label{Analysis of open-weight models}

\begin{table}[t]
    \centering
    \small
    \begin{tabular}{cccc}
         \toprule
         Model        & \(\zeta_{L}\) & \(\zeta_{L-1}\) & \(\zeta_{L-2}\) \\
         \midrule
         Llama-3.2-1B & 20.6 & 22.7 & 22.7 \\
         Llama-3.2-70B & 10.1 & 15.7 & 16.2 \\
         Gemma-2B     & 22.2 & 28.2 & 28.2 \\
         Gemma-7B     & 0.48 & 12.7 & 14.8 \\
         DeepSeek-7B  & 13.6 & 13.5 & 12.7 \\
         DeepSeek-67B  & 7.16 & 8.48 & 8.87 \\
         Qwen3-32B & 7.50 & 9.00 & 9.36 \\
         layerskip-llama3.2-8B & 4.62 & 6.64 & 6.94 \\
         \bottomrule
    \end{tabular}
    \caption{Jump rates $\zeta_{L}$, $\zeta_{L-1}$, and $\zeta_{L-2}$ of the six open-weight models. All models exhibit a pronounced hidden state jump at the final layer.}
    \label{tab:open_weight_jump_rate}
\end{table}

We measured the displacement \(\Psi_\ell\) across multiple open-weight models of various sizes to show that these models exhibit a pronounced hidden state jump near the final layer.
Specifically, we evaluated the Llama-3.2-1B, Llama-3.2-70B~\citep{DBLP:journals/corr/abs-2407-21783}, Gemma-2B, Gemma-7B~\citep{DBLP:journals/corr/abs-2403-08295}, DeepSeek-7B, DeepSeek-67B~\cite{DBLP:journals/corr/abs-2401-02954}, Qwen3-32B~\cite{yang2025qwen3technicalreport} and Layerskip-Llama3.2-8B~\citep{DBLP:conf/acl/ElhoushiSLHWL0A24}.
Layerskip is a multi-exit model based on the Llama architecture that adds additional classifiers to each Transformer layer and jointly optimizes all layers with the same cross-entropy loss as the final layer to enable early exit inference.

\cref{fig:latest_model_hidden_trajectory} shows the displacement \(\Psi_\ell\) for each model, computed as the mean per-token displacement during inference on the LAMBADA dataset~\citep{DBLP:conf/acl/PapernoKLPBPBBF16} and~\cref{tab:open_weight_jump_rate} shows jump rate $\zeta_{L}$, $\zeta_{L-1}$, and $\zeta_{L-2}$ for each model.
Across all models, the final layer exhibits significantly greater displacement than the middle layers.
Additionally, deeper models tend to have jumps from one or two layers before the final layer as well.
This observation suggests that large displacement in the final layer persists not only in conventional models but even in Layerskip, which is trained to equip its middle layers with vocabulary prediction performance on par with the final layer.
A similar phenomenon is also observed on the Wikitext~\citep{DBLP:conf/iclr/MerityX0S17} and ARC-easy~\citep{DBLP:journals/corr/abs-1803-05457} datasets, indicating that this behavior is not specific to a particular dataset.\footnote{Detailed results are shown in \cref{appendix:displacemenet_on_other_dataset}.}
\subsection{Conditions for Jump Formation}
\label{Formation condition of the jump}

\begin{table}[t]
    \centering
    \small
    \begin{tabular}{ccccccc}
         \toprule
         Model  & Size & 20\% & 40\% & 60\% & 80\% & 100\%  \\
         \midrule
         \multirow{2}{*}{Pythia}  & 1B & 1.03 & 2.79 & 5.37 & 10.4 & 11.2\\
         & 12B & 1.10 & 2.23 & 4.55 & 9.11 & 10.0\\
         \midrule
         \multirow{2}{*}{OLMo} & 1B & 6.06 & 7.84 & 8.84 & 10.3 & 11.9\\
         & 7B & 4.94 & 6.40 & 7.63 & 9.14 & 10.3\\
         \bottomrule
    \end{tabular}
    \caption{Jump rate values $\zeta_{L}$ for the Pythia and OLMo models, whose pre-training checkpoints are publicly available, measured at 20\%, 40\%, 60\%, 80\%, and 100\% (the final checkpoint) of the total training steps. For each model, we confirmed that the jump rate increases monotonically as training progresses.}
    \label{tab:open_weight_trajectory_via_checkpoint}
\end{table}

We analyzed how displacement changes throughout pre-training using open-weight models for which intermediate checkpoints are available.
We employed the Pythia~\citep{DBLP:conf/icml/BidermanSABOHKP23} and OLMo~\citep{DBLP:conf/acl/GroeneveldBWBKT24} to analyze intermediate checkpoints.
The Pythia 1B and 12B models were trained on 300B tokens for 143,000 steps.
We evaluate them every 30,000 steps.
The OLMo 1B and 7B models were trained on 2T tokens for 738,000 steps and 2.46T tokens for 557,000 steps, respectively.
We evaluate the 1B and 7B models every 150,000 and 110,000 steps\footnote{The final checkpoints for all models are not exactly at every 30,000, 150,000, or 110,000 steps. See~\cref{fig:pythia_ckpt_trajectory}.}, respectively.

\cref{fig:pythia_ckpt_trajectory} shows the displacement values at each checkpoint and \cref{tab:open_weight_trajectory_via_checkpoint} shows these model jump rates.
The jump rate grows monotonically with the number of updates, indicating that larger step counts yield larger \(\zeta_{L}\).

\subsection{Effect of Training Steps on Jump Behavior}
\label{Effect of Training Steps on Jump Behavior}
\begin{figure}[t]
    \centering
    \includegraphics[width=\linewidth]{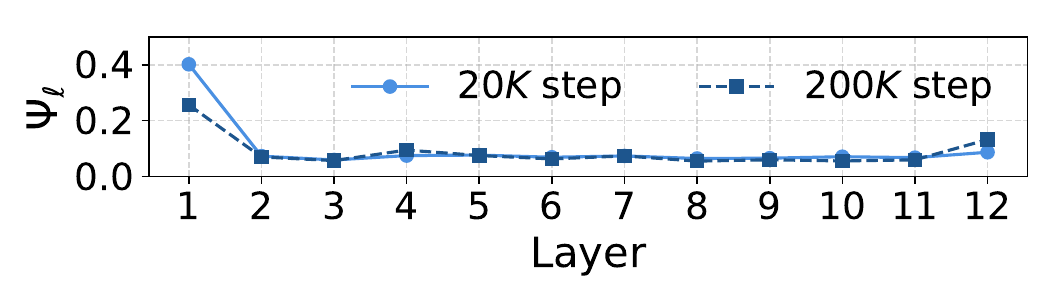}
    \caption{Hidden state displacement for the 170M Llama-based model pre-trained on 100B tokens with two different update steps. (light blue):  20K steps, (dark blue) 200K steps. Jump rates were $\zeta_{L}=1.93$ for 20K steps and $\zeta_{L}=7.24$ for 200K steps. Despite being trained on the same number of tokens, models with more update steps exhibit larger jumps in hidden state displacement.}
    \label{fig:step_token_relationship}
\end{figure}

We investigated how the number of training steps influences the jump behavior in scratch pre-trained models to identify baseline training conditions that can reproduce the jump phenomenon observed in open-weight models.
We examined two 170M Llama-based models.\footnote{The bacic experimental setup can be fonund in~\cref{Pretraining from Scratch}.}
These models were pretrained on the same corpus of 100B tokens but with distinct step counts by varying the batch size.

\cref{fig:step_token_relationship} shows the displacement for models trained for 20K and 200K steps.
The resulting jump rates are \(\zeta_L = 1.93\) for 20K steps and \(\zeta_L = 7.24\) for 200K steps.
These results show that, even with an identical token budget, a greater number of update steps produces a larger jump and hence the properties of displacement more closely resemble those of open-weight models.

\section{Proposed Method}
\label{Proposed Method}

\begin{figure}[t]
    \centering
    \includegraphics[width=\linewidth]{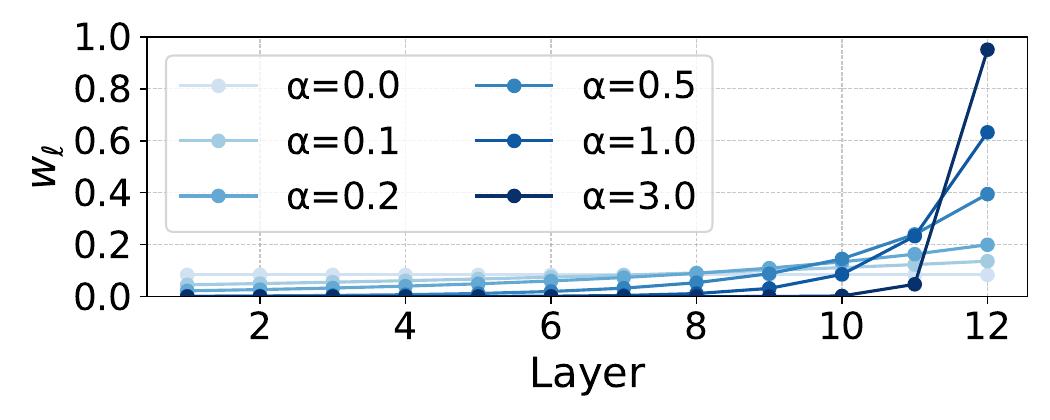}
    \caption{
        Weight coefficients \(w_{\ell}\) for different values of the hyperparameter \(\alpha\) in \cref{eq:L_cos}.
        When $\alpha=0.0$, the weight coefficients are uniform, when $\alpha=0.1$, they are almost linear, and for larger $\alpha$, they increase exponentially.
    }
    \label{fig:alpha-weight}
\end{figure}

As discussed in~\cref{Analysis of open-weight models}, the increased hidden state displacement at the final layer reflects abrupt changes in the angular distance.
Motivated by this observation, this paper focuses on explicitly modeling and controlling this behavior through the design of an appropriate training objective.

Previous studies have shown that 
(1) the angular distance between consecutive hidden states in the middle layers is typically small~\citep{DBLP:conf/aaai/SunPNJ25, DBLP:journals/corr/abs-2407-15516}, and 
(2) layers that exhibit small angular distances can often be pruned without reducing the performance of the model~\citep{DBLP:journals/corr/abs-2403-03853}.
(3) modifying the architecture so as to increase the angular distance change between the input and output hidden states in middle layers can improve model performance~\cite{DBLP:journals/corr/abs-2502-05795, DBLP:journals/corr/abs-2412-13795}.
These findings suggest that small angular distances are closely associated with redundancy and underutilization in the middle layers.

Building on these findings, we hypothesize that the phenomenon of jump may indicate that the network relies too heavily on the final layer while underutilizing the middle layers.
To test this hypothesis, we add a penalty on the final layer displacement to the loss function and experimentally evaluate the performance of models that suppress this jump.  

Let $\mathcal{V}$ be the vocabulary, and let $\lvert\mathcal{V}\rvert$ denotes the number of tokens in the vocabulary.
In general, pre-training minimizes the cross-entropy loss, $\mathcal{L}_{\mathrm{CE}}$, that is:
\begin{align}
    \label{eq:baseline_loss}
    \mathcal{L}_{\mathrm{CE}}
    &= -{\sum}_{i=1}^{\lvert\mathcal{V}\rvert} y_i \log p_i,
\end{align}
where $y_i$ is the $i$-th element of the one-hot vector $\vy\in\{0,1\}^{\lvert\mathcal{V}\rvert}$ that represents the ground-truth token id, and $p_i$ is the $i$-th element of the vector $\vp = \mathtt{softmax}\big(\mW\mathtt{RMS}(\vh_L)\big)$.
Here, \(\vh_L\) denotes the final layer hidden state vector, and the operator $\mathtt{RMS}(\cdot)$ refers to RMSNorm~\citep{DBLP:conf/nips/ZhangS19a}, which rescales the hidden state vectors. 
Finally, the matrix $\mW\in\mathbb{R}^{|\mathcal{V}|\times D}$ projects this normalized vector to the unnormalized vocabulary (logits) vector, and the softmax function, $\mathtt{softmax}(\cdot)$, then converts the logits vector into a valid probability distribution, $\vp$.
Obviously, the cross-entropy loss, $\mathcal{L}_{\mathrm{CE}}$, considers only output performance and does not, in any way, optimize for internal states.
Here, in addition to $\mathcal{L}_{\mathrm{CE}}$, we introduce a loss function $\mathcal{L}_{\mathrm{disp}}$ that suppresses the displacement in the final layer:
\begin{align}
    \mathcal{L}_{\mathrm{disp}} &= {\sum}_{\ell=1}^{L} w_\ell \Psi_\ell, \label{eq:L_cos} 
\end{align}
where $w_\ell$ is the $\ell$-th element of $\vw = \mathtt{softmax}(\alpha \vl)$, and $\vl$ is the $L$-dimantional vector, where $\vl=(1,2,\dots, L)$.
Note that, intuitively, adding a regularization term over $\Psi_L$ would eliminate the pronounced displacement in the final layer, however confining the regularizer to the final layer simply shifts this displacement to the penultimate layer, which would impede our ability to investigate the relationship between the final layer displacement and model performance.\footnote{Details are discussed in~\cref{Regularizing only the final layer displacement}.}
Therefore, we employed $\Psi_\ell$ for all layers in our regularizer, but weighted layers closer to the final layer more heavily.
\cref{fig:alpha-weight} shows examples of the $w_\ell$ distribution over the layer index $\ell$ for each $\alpha$.
As $\alpha$ increases, $w_L$ approaches 1, focusing the strongest penalty on the final layer displacement $\Psi_L$, and yields uniform weights when $\alpha=0$.
Finally, the loss function of the proposed Jump-Suppressing Regularizer (JREG) is defined as follows:
\begin{align}
    \label{eq:proposed_loss}
    \mathcal{L}_{\mathrm{JREG}} &= \mathcal{L}_{\mathrm{CE}} + \lambda \mathcal{L}_{\mathrm{disp}},
\end{align}
where the hyperparameter $\lambda$ is employed to adjust the relative importance of the two loss functions.

\begin{table*}[t]
    \centering
    \small
    \tabcolsep 3pt
    \begin{tabular}{cc c|ccccccccc|c}
        \toprule
         Size&Method & $\alpha$ & ARC-e & BoolQ & HellaSwag & LAMBADA & PIQA & RACE & SocialIQA & SciQ & SWAG & \textbf{avg} \\
         \midrule
         170M &
         Baseline & -
         & 54.9 & 57.5 & 32.1 & 32.0 & 64.0 & 29.1 & 38.4 & 80.2 & 39.7 & 47.5\\
         \cmidrule(r){2-13}
         && 0.0 & 55.7 & 58.3 & 32.4 & \textbf{32.5} & \textbf{65.9} & 29.6 & 38.6 & 81.3 & \textbf{40.2} & 48.3\\
         && 0.1 & 56.1 & 60.5 & 32.2 & 29.8 & 65.6 & 29.6 & 39.2 & 80.9 & 39.9 & 48.2\\
         && 0.3 & 56.0 & 59.2 & 32.2 & 31.7 & 65.0 & 29.1 & 38.1 & \textbf{82.2} & 40.0 & 48.0\\
         && 0.5 & 57.0 & 57.1 & \textbf{32.5} & 30.7 & 65.0 & 29.3 & 38.6 & \textbf{82.2} & 40.0 & 48.0\\
         &&  1.0 & \textbf{57.2} & 60.0 & 32.1 & 31.7 & 65.2 & \textbf{29.9} & 38.8 & 81.1 & \textbf{40.2} & \textbf{48.5}\\
         &\multirow{-6}{*}{\shortstack{JREG (ours)}} &  3.0 & \textbf{57.2} & 60.0 & 32.4 & 31.6 & 65.1 & 28.9 & \textbf{40.0} & 81.7 & 39.9 & \textbf{48.5}\\
         \midrule
         1B
         &Baseline & - & 68.5 & \textbf{61.4} & 42.9 & 46.6 & \textbf{72.4} & 32.9 & 41.0 & 89.5 & 47.3 & 55.8 \\
         \cmidrule(r){2-13}
         & & 1.0 & 69.4 & \textbf{61.4} & \textbf{43.1} & \textbf{47.2} & 71.8 & \textbf{35.3} & \textbf{42.1} & 88.6 & \textbf{47.6} & \textbf{56.2} \\
         &\multirow{-2}{*}{JREG (ours)} & 3.0 & \textbf{70.6} & 59.3 & 42.6 & 45.6 & 71.8 & 34.9 & 41.4 & \textbf{91.2} & 47.3 & 56.1 \\
         \midrule
         3.4B
         &Baseline & - & 75.6 & 59.4 & 49.7 & 55.9 & \textbf{76.5} & 36.3 & \textbf{43.9} & \textbf{93.4} & 51.3 & 60.2 \\
         \cmidrule(r){2-13}
         &JREG (ours) & 1.0 & \textbf{77.9} & \textbf{61.6} & \textbf{50.3} & \textbf{57.2} & 75.2 & \textbf{37.8} & 42.5 & 93.0 & \textbf{51.5} & \textbf{60.8} \\
         \bottomrule
    \end{tabular}
    \caption{
    Results of downstream task performance. $\alpha$ indicates the hyperparameter of JREG.
    Each row shows the score on each benchmark dataset, with its average in the rightmost column.
    The upper table shows the results of models trained with $\mathcal{L}_{\mathrm{CE}}$ (baseline) and $\mathcal{L}_{\mathrm{JREG}}$ (ours) using a 170M model, while the lower table shows those trained with the 1B and 3.4B model.
    }
    \label{tab:eval_scratch}
\end{table*}
\begin{table}[t]
  \centering
  \small
  \begin{tabular}{cccrrr}
    \toprule
    \multicolumn{2}{c}{Model} & $\alpha$ & $\zeta_L$ & $\zeta_{L-1}$ & $\zeta_{L-2}$ \\
    \midrule
    & Baseline & - & 7.24 & 7.63 & 7.63 \\
    & & 0.0 & 0.84 & 1.28 & 1.76 \\
    & & 0.1 & 0.48 & 0.72 & 1.05 \\
    &  & 0.3 & \textbf{0.00} & 0.05 & 0.15 \\
    &  & 0.5 & \textbf{0.00} & \textbf{0.00} & \textbf{0.00} \\
    &  & 1.0 & \textbf{0.00} & \textbf{0.00} & \textbf{0.00}\\
      \multirow{-6}{*}{170M} 
    & \multirow{-6}{*}{JREG (ours)} 
    & 3.0 
    & \textbf{0.00} & \textbf{0.00} & 1.55\\
    \midrule
      & Baseline & - & 5.42 & 5.66 & 5.66 \\
      &
      & 0.5
      & \textbf{0.00} & \textbf{0.00} & \textbf{0.00} \\
      & & 1.0 
      & \textbf{0.00} & \textbf{0.00} & \textbf{0.00} \\
      \multirow{-4}{*}{1B} & \multirow{-3}{*}{JREG (ours)} 
      & 3.0 
      & \textbf{0.00} & \textbf{0.00} & \textbf{0.00} \\
    \midrule
      & Baseline & - & 5.21 & 6.15 & 6.25\\
      \multirow{-2}{*}{3.4B} & JREG (ours) & 1.0 & \textbf{0.00} & \textbf{0.00} & \textbf{0.00} \\
    \bottomrule
  \end{tabular}
  \caption{Comparison of jump rates ($\zeta_L, \zeta_{L-1}, \zeta_{L-2}$) as defined in~\cref{eq:zeta} for baseline models and models trained with JREG. Results are shown for different hyperparameter $\alpha$ settings on 170M, 1B and 3.4B models. JREG effectively suppresses jump rates near the final layers.}
  \label{tab:eval_jump_rate}
\end{table}

\section{Experiments}
\label{experiments}
In the following, we outline the experimental settings for applying the JREG method during pre-training. These experiments were performed to quantitatively evaluate the proposed method’s impact on suppressing final layer displacement, the overall model performance, and further assess the performance after supervised fine-tuning (SFT)
\footnote{
The pre-training overhead of JREG is discussed in ~\cref{sec:impact-on-training-time-and-memory}.
In terms of training time, JREG is comparable to the baseline and is practically acceptable.
}.

\subsection{Pre-training from scratch}
\label{Pretraining from Scratch}
In this study, all experiments employed Llama-based models, and three variants differing in terms of the parameter count and depth were selected.
The 170M model comprises 12 layers with a model dimension of 768, the 1B model comprises 16 layers with a model dimension of 2048, and the 3.4B model comprises 30 layers with a model dimension of 3072.\footnote{Detailed hyper-parameters for model architectures and training are provided in~\cref{appendix:Model settings} and~\cref{appendix:Training settings}, respectively.}
This difference in depth facilitated the evaluation of how both representation scale and depth influence the effectiveness of the proposed JREG method. 
For pretraining data, we use 100B tokens extracted from the FineWeb-Edu~\citep{DBLP:conf/nips/PenedoKALMRW024}.
We set the total number of parameter update steps to 200K to make the ``jump'' more pronounced in the baseline model.

\subsection{Evaluation metrics}
\label{sec:evaluation}

\paragraph{Downstream task performance.}
For our evaluation, we employed several downstream tasks. 
We measure next-token prediction performance on the LAMBADA in term of the top-1 accuracy~\citep{DBLP:conf/acl/PapernoKLPBPBBF16}.
In addition, we assessed the accuracy of binary-choice QA on BoolQ~\citep{DBLP:conf/naacl/ClarkLCK0T19}
and multiple-choice QA on ARC-easy (ARC-e)~\citep{DBLP:journals/corr/abs-1803-05457}, HellaSwag~\citep{DBLP:conf/acl/ZellersHBFC19}, 
PIQA~\cite{DBLP:conf/aaai/BiskZLGC20}, RACE~\citep{DBLP:conf/emnlp/LaiXLYH17}, Social IQA~\citep{DBLP:conf/emnlp/SapRCBC19}, SciQ~\citep{DBLP:conf/aclnut/WelblLG17},  SWAG~\citep{DBLP:conf/emnlp/ZellersBSC18}.

\paragraph{Jump rate.}
We measure the improvements in the jump rates $\zeta_L$, $\zeta_{L-1}$, and $\zeta_{L-2}$ in~\cref{eq:zeta}.

\subsection{Supervised fine-tuning (SFT)}
\paragraph{Training settings}
We performed SFT on the 3.4B model, updating all parameters during training.
For the SFT data, we employed the Tulu-v1 SFT mixture dataset \citep{DBLP:conf/nips/WangIDHKCWMSBH23} which contains diverse instruction-following examples.  
\footnote{Detailed hyper-parameters for training are provided in~\cref{appendix:SFT settings}}

\paragraph{Evaluation metrics}
We use three benchmark, MT-bench \cite{DBLP:conf/nips/ZhengC00WZL0LXZ23},  Vicuna bench \cite{DBLP:conf/nips/ZhengC00WZL0LXZ23} and WizardLM testset \cite{DBLP:conf/iclr/XuSZG0FTLJ24}.
For evaluation, we used GPT-4 as the judge model.  
\footnote{These experimental settings are based on~\cite{DBLP:conf/nips/ZhouTQYJX0024}.}

\section{Results}
\label{Results}
Unless otherwise stated, we fix $\lambda=1.0$.\footnote{This value of $\lambda=1.0$ was selected via a preliminary sweep reported in~\cref{appendix:Hyperparameter Tuning for JREG}} for all experiments. 
We vary $\alpha$ as described in each subsection.

\subsection{Pre-training on 170M model}
\label{Pretraining on 170M model}
For the 170M model, we conducted a sweep over the JREG hyperparameter $\alpha$, testing values $\alpha \in \{0.0, 0.1, 0.3, 0.5, 1.0, 3.0\}$, to evaluate the effect of the weighting parameter.

As shown in~\cref{tab:eval_scratch}, the proposed JREG method consistently outperformed the baseline in average downstream task performance. The performance tended to peak or stabilize with  $\alpha \in\{1.0, 3.0\}$.
In addition, statistical validation confirmed that the observed performance improvements were significant\footnote{Details are discussed in~\cref{Pre-trained Model Evaluation}}
and not dependent on the specific training dataset\footnote{Details are discussed in~\cref{appendix:Result on Other Datasets}}.

For the jump rates $\zeta_L$, $\zeta_{L-1}$, and $\zeta_{L-2}$,~\cref{tab:eval_jump_rate} indicates that increasing $\alpha$ led to a substantial reduction. 
This suppression of the jump rates, which is attributed to JREG's stronger penalty on deeper layer displacement (\cref{fig:alpha-weight}), correlates with improved downstream performance.
These findings suggest that suppressing displacements around the final layers improves the model's overall capability.

\subsection{Pre-training on larger model}
\label{Pretraining on 500M and 1B model}
To evaluate how the proposed JREG method scales, we applied it to larger 1B and 3.4B models, and performed the same evaluations as in the 170M model.
First, we applied the proposed JREG method to 1B models with top performing settings found on the 170M model, $\alpha \in \{1.0, 3.0\}$, and evaluate its effectiveness with jump rate $\zeta_{L}$, downstream task performance.

\cref{tab:eval_scratch} shows downstream task performance for a 1B model after pre-training. 
For all settings of $\alpha$, the average downstream task performance exceeds the baseline.
However, while the 170M model achieved identical performance for $\alpha \in \{1.0, 3.0\}$, the 1B model attained its highest performance at $\alpha = 1.0$.
We attribute this difference to the increased number of layers in the larger model, which shifts the optimal hyperparameter $\alpha$ relative to the 170M model.

\cref{tab:eval_jump_rate} compares the jump rate \(\zeta_L\) of the 1B models between the baseline and JREG methods. 
The jump rates \(\zeta_L\) are 6.42 with the 1B baseline models, whereas in JREG, they are 0.0.
Thus, similar to the 170M model, these results confirm that the proposed method suppresses the final layer hidden state jump.

Furthermore, we extended our experiment to the 3.4B model, comparing the baseline and the JREG method with $\alpha=1.0$, the best setting from 1B. 
Similar to the 1B model,
JREG improves downstream performance and suppresses the jump rate over the baseline.

\subsection{Supervised fine-tuning on 3.4B model}
\label{Supervised fine-tuning on 3.4B model}
\input{table/displacement}
\input{table/acc_sft_table}
\input{table/jump_rate_table_llama_sft}
To further assess the downstream task performance of our pre-trained models, we performed SFT.

\cref{tab:acc_sft_table} shows the results on three benchmarks for the 3.4B baseline and JREG ($\alpha=1.0$) models after SFT.
The proposed JREG method outperformed the baseline on all benchmarks, with average scores of 4.28 and 4.67 for the baseline and JREG method, respectively.
These improvements are statistically significant\footnote{Details are discussed in~\cref{SFT Model Evaluation}}. 
~\cref{tab:jump_rate_table_llama_sft} compares the jump rates $\zeta_{L}$, $\zeta_{L-1}$ and $\zeta_{L-2}$ of the 3.4B models between the baseline and JREG method after SFT.
The baseline model exhibits $\zeta_{L} = 5.14$, which indicates that the jump behavior at the final layer persisted even after SFT.
In contrast, the models pre-trained with JREG exhibit $\zeta_{L}$, $\zeta_{L-1}$, and $\zeta_{L-2}$ equal to $0.0$, which indicates that the jump behavior was eliminated entirely.
Overall, these results suggest that the impact of the JREG method during pre-training is preserved after SFT, resulting in models that do not exhibit jump behavior and improved downstream performance.

\section{Analysis and Discussion}
\label{Analysis and Discussion}
In the following, we quantitatively evaluate the degree of redundancy reduction in the middle layers achieved by JREG.
We define the improvement in redundancy at layer $\ell$, denoted as $\Delta_{\ell}$, as the difference in displacement between the JREG and baseline models.
\begin{equation}
  \label{eq:redundancy}
  \Delta_{\ell}
  = \Psi_{\ell}^{\mathrm{JREG}}
  - \Psi_{\ell}^{\mathrm{Baseline}},
\end{equation}
Here, $\Delta_{\ell} > 0$ indicates that the proposed JREG method yields a larger displacement than the baseline, thereby signifying an improvement in redundancy at that layer. 
We compute $\Delta_{\ell}$ for all layers $\ell$ to assess the degree of redundancy improvement across the network.

\Cref{tab:displacement} shows the displacement $\Delta_{\ell}$ between the JREG and the baseline on the 3.4B model
\footnote{170M and 1B displacement are discussed in the~\cref{Displacement across Different Weight Settings}}.
The JREG models exhibit a tendency for smaller displacement near the final layers compared to the baseline ($\Delta_{\ell} < 0$), while the displacement in the middle layers tends to be larger ($\Delta_{\ell} > 0$).
Specifically, from layers 1 through 26, excluding the 8th layer of the 26 layers, JREG shows greater displacement.
This suggests that, while JREG reduced the overcontribution of the layers near the final layer, it increased the effectiveness of the middle layers, thereby enhancing their contribution relative to the baseline.

\section{Conclusion}
This study investigated the jump behavior, defined as the pronounced large change in angular distance between the input and output hidden state vectors, occurring in or around the final layer(s) in Transformer-LMs, and examined its impact on model capacity. 
We first revealed that this phenomenon is ubiquitous across recent Transformer-LMs, regardless of open-weight pre-trained models and models trained from scratch.
We then confirmed that our proposal of JREG, which penalises large hidden state displacements near the final layer during pre-training, consistently (i) reduced the hidden state jumps in the final layer, (ii) increased the relative contribution of middle layers, and (iii) improved downstream task performance without altering the model architecture, in our experiments conducted on three model sizes of Llama-based architecture.
These findings suggest that mitigating hidden state jumps offers a simple yet effective means to unlock the latent capacity of middle layers in Transformer-LMs.

\clearpage
\section*{Limitations}
Despite the positive findings, several limitations of this study should be addressed in future research.  
First, our experiments were conducted exclusively on Llama-based architectures.  
While Llama represents a prominent family of models in the field, it remains unclear whether the proposed method JREG would be equally effective for other architectural designs, such as Transformer variants with different normalization schemes, attention mechanisms, or entirely different model architectures.
Second, JREG is depth-sensitive.
Therefore, as the number of layers changes, the optimal hyperparameter values also vary, requiring re-tuning to achieve the best performance.

\section*{Ethical Considerations}
In this study, we exclusively used publicly available datasets for pre-training, fine-tuning, and evaluation.
In addition, we developed the language models entirely from scratch, avoiding the use of publicly available models. 
For the publicly available datasets and existing pre-trained models analyzed in this study (including those under terms such as the~\href{https://www.llama.com/llama3/license/}{META LLAMA 3 COMMUNITY LICENSE}), we strictly adhered to all applicable licenses.
Given that the proposed JREG method is a framework for pre-training language models, the risk of ethical concerns is expected to be minimal.

During code development and writing, we used AI assistants, including language models.
All the generated code snippets and texts are checked and modified by the authors for scientific integrity and accuracy.

\section*{Acknowledgments}
This work was supported by the ``R\&D Hub Aimed at Ensuring Transparency and Reliability of Generative AI Models'' project of the Ministry of Education, Culture, Sports, Science and Technology, and JST Moonshot R\&D Grant Number JPMJMS2011-35 (fundamental research), and JST BOOST Grant Number JPMJBS2421.
The computation was carried out using the computer resource offered
under the category of General Projects by Research Institute for
Information Technology, Kyushu University and ABCI 3.0 provided by AIST and AIST Solutions with support from “ABCI 3.0 Development Acceleration Use”.

\bibliography{custom}

\clearpage
\appendix

\section{Details of experimental settings}
\label{appendix:Details of experimental settings}
\subsection{Model settings}
\label{appendix:Model settings}

\begin{table}[h]
    \centering
    \small
    \begin{tabular}{c|ccc}
        \toprule
         & 170M & 1B & 3.4B \\
         \midrule
         Layers & 12 & 16 & 30 \\
         Model Dim & 768 & 2048 & 3072 \\
         FFN Dim & 2048 & 5376 & 8192 \\
         Attention Heads & 12 & 16 & 24 \\
         Key / Value Heads & 12 & 16 & 24\\
         Activation & \multicolumn{3}{c}{SwiGLU}\\
         Vocabulary Size & \multicolumn{3}{c}{32000}\\
         \bottomrule
    \end{tabular}
    \caption{Model settings.}
    \label{tab:model_settings}
\end{table}

\subsection{Pre-training settings}
\label{appendix:Training settings}

\begin{table}[ht]
    \centering
    \small
    \begin{tabular}{c|ccc}
        \toprule
         Configuration & 170M & 1B & 3.4B \\
        \midrule
         lr & \(9\times10^{-4}\) & \(7\times10^{-4}\) & \(5\times10^{-4}\) \\
         local batch & 128 & 64 & 32 \\
         global batch & \multicolumn{3}{c}{512} \\
         sequence len & \multicolumn{3}{c}{1024} \\
         weight decay & \multicolumn{3}{c}{0.1} \\
         epsilon & \multicolumn{3}{c}{\(1\times10^{-8}\)} \\
         Optimizer & \multicolumn{3}{c}{AdamW ($\beta_1 = 0.9$, $\beta_2=0.95$)} \\
         clip & \multicolumn{3}{c}{1.0} \\
         scheduler & \multicolumn{3}{c}{cosine} \\
         warmup & \multicolumn{3}{c}{1000} \\
         lr\_min\_ratio & \multicolumn{3}{c}{0.1} \\
         cycle\_length & \multicolumn{3}{c}{1.0} \\
        \bottomrule
    \end{tabular}
    \caption{Pre-training settings.}
    \label{tab:training_settings}
\end{table}

~\cref{tab:training_settings} details the hyperparameter configurations employed during training. The model was trained on a corpus of 100 billion tokens, and its performance was validated on a dataset of 11 million tokens.
For our implementation, we used the Meta Lingua library~\citep{meta_lingua} as the codebase for our pre-training experiments. The total computational budget varied by model size:
\begin{itemize}
   \item 170M model: 16 hours on 4$\times$ NVIDIA H100 (94GB) GPUs
   \item 1B model: 55 hours on 8$\times$ NVIDIA H200 (141GB) GPUs
   \item 3.4B model: 159 hours on 8$\times$ NVIDIA H200 (141GB) GPUs
\end{itemize}

\subsection{Supervised fine-tuning (SFT) settings}
\label{appendix:SFT settings}
The maximum learning rate is set to $2\times10^{-5}$ with a linear scheduler, and the optimizer is AdamW ($\beta_1=0.9$, $\beta_2=0.95$).  
We use a global batch size of 512 and train for a total of 957 steps.
The total computational budget is $2$ hours on $8 \times$ NVIDIA H200 GPUs.
For training, we employ huggingface transformers~\citep{DBLP:conf/emnlp/WolfDSCDMCRLFDS20},
for evaluation, we employ FastChat \citep{DBLP:conf/nips/ZhengC00WZL0LXZ23}.

\subsection{Evaluation settings for Pre-trained Models}
\label{appendix:Evaluation settings}
For evaluating performance on downstream tasks described in Section~\ref{sec:evaluation}, we utilized the \texttt{eval\_harness} framework~\citep{gao-2024-eval-harness}, which provides standardized implementations for a wide range of NLP benchmark tasks. All evaluations were conducted following the default hyperparameter configurations specified in the \texttt{eval\_harness} framework, ensuring consistency and comparability with prior work.

\section{Displacement on other dataset}
\label{appendix:displacemenet_on_other_dataset}
In order to examine the displacement properties of open-weight models on datasets with distributions different from LAMBADA, we conducted experiments analogous to those in~\cref{Analysis of open-weight models} on the test splits of Wikitext~\citep{DBLP:conf/iclr/MerityX0S17} and ARC-Easy~\citep{DBLP:journals/corr/abs-1803-05457}.
\subsection{Wikitext}

\begin{figure}[t]
    \centering
    \begin{subfigure}[b]{\linewidth}
        \centering
        \includegraphics[width=\linewidth]{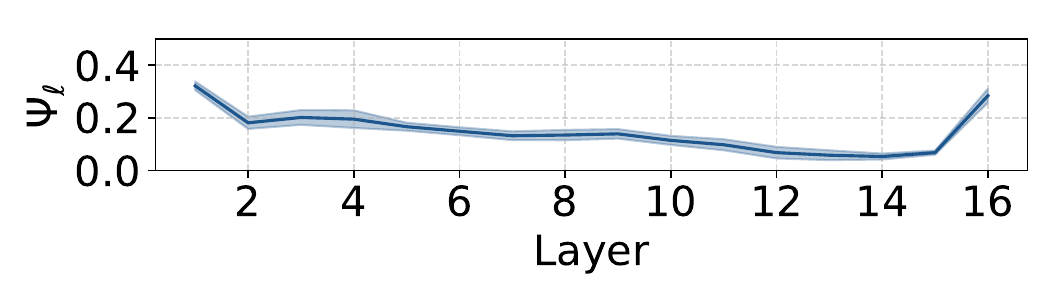}
        \caption{Llama3.2-1B}
        \label{fig:Llama3.2_1b_wikitext}
    \end{subfigure}
    \begin{subfigure}[b]{\linewidth}
        \centering
        \includegraphics[width=\linewidth]{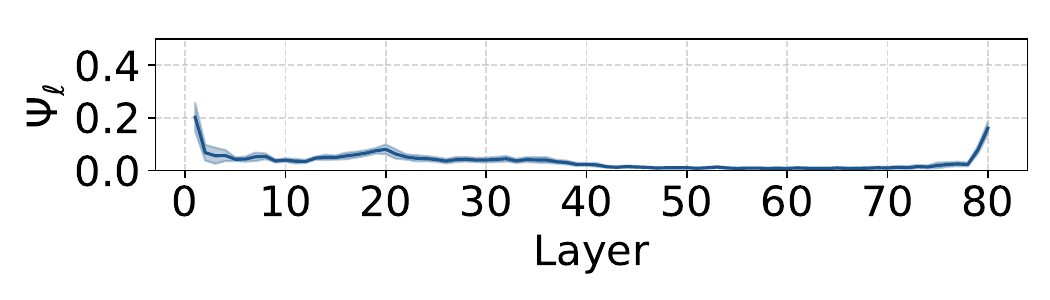}
        \caption{Llama3.2-70B}
        \label{fig:Llama3.2_70b_wikitext}
    \end{subfigure}
    \begin{subfigure}[b]{\linewidth}
        \centering
        \includegraphics[width=\linewidth]{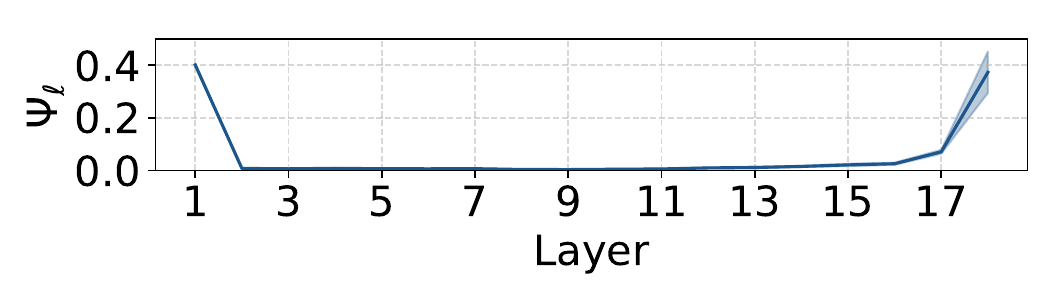}
        \caption{Gemma-2B}
        \label{fig:gemma7b_wikitext}
    \end{subfigure}
    \begin{subfigure}[b]{\linewidth}
        \centering
        \includegraphics[width=\linewidth]{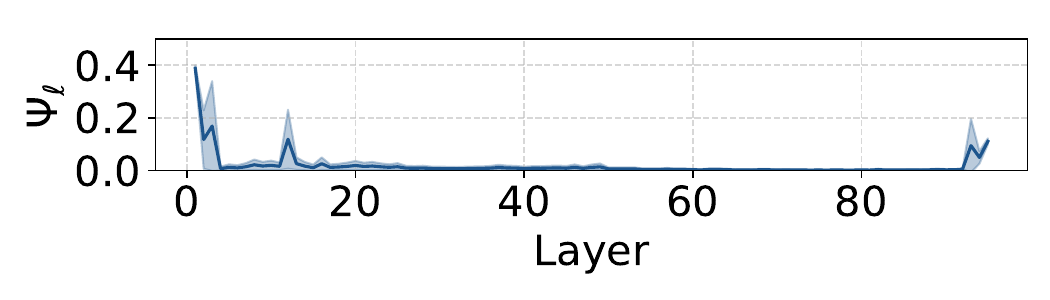}
        \caption{DeepSeek-67B}
        \label{fig:deepseek67b_wikitext}
    \end{subfigure}
    \begin{subfigure}[b]{\linewidth}
        \centering
        \includegraphics[width=\linewidth]{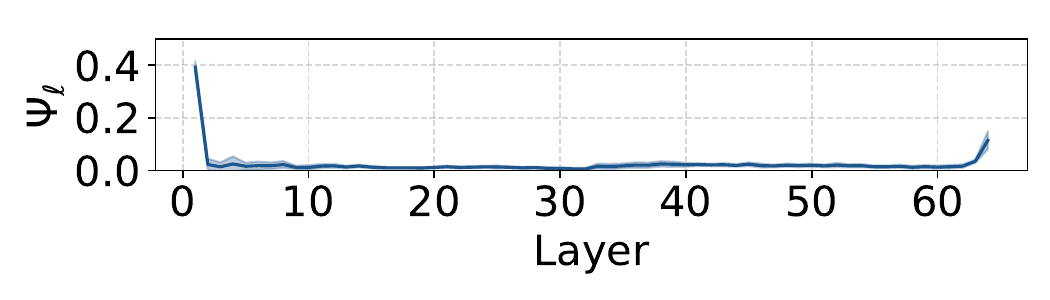}
        \caption{Qwen3-32B}
        \label{fig:qwen3_32b_wikitext}
    \end{subfigure}
    \begin{subfigure}[b]{\linewidth}
        \centering
        \includegraphics[width=\linewidth]{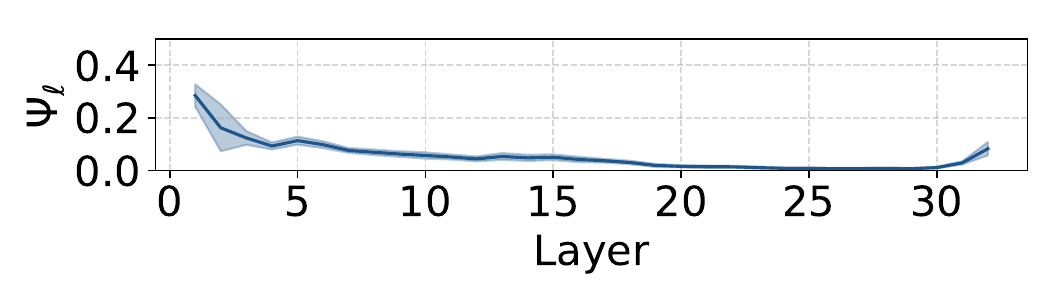}
        \caption{layerskip-llama3.2-8B}
        \label{fig:layerskip-llama3.2-8B_wikitext}
    \end{subfigure}
    \caption{Displacement on Wikitext.}
    \label{appendix:trajectory_wikitext}
\end{figure}

~\cref{appendix:trajectory_wikitext} shows the results of computing displacement using Wikitext.
\subsection{ARC-Easy}

\begin{figure}[t]
    \centering
    \begin{subfigure}[b]{\linewidth}
        \centering
        \includegraphics[width=\linewidth]{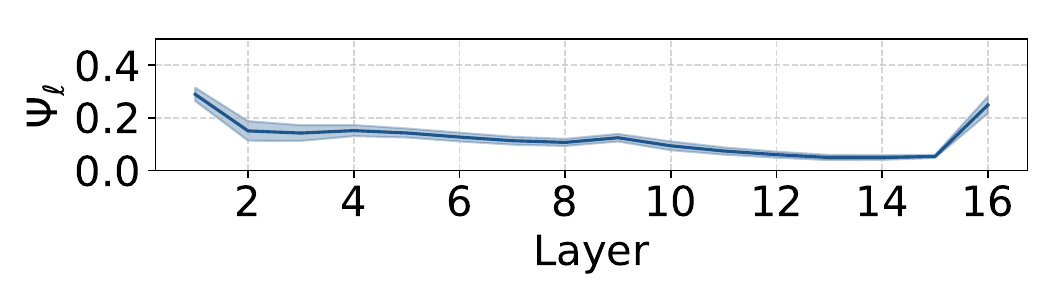}
        \caption{Llama3.2-1B}
        \label{fig:Llama3.2_1b_arc}
    \end{subfigure}
    \begin{subfigure}[b]{\linewidth}
        \centering
        \includegraphics[width=\linewidth]{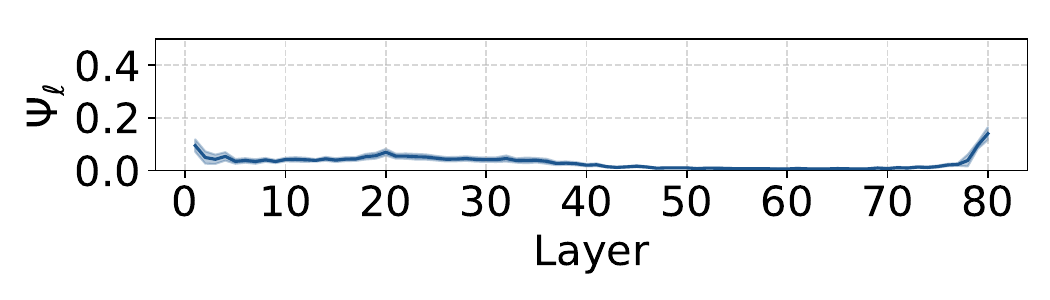}
        \caption{Llama3.2-70B}
        \label{fig:Llama3.2_70b_arc}
    \end{subfigure}
    \begin{subfigure}[b]{\linewidth}
        \centering
        \includegraphics[width=\linewidth]{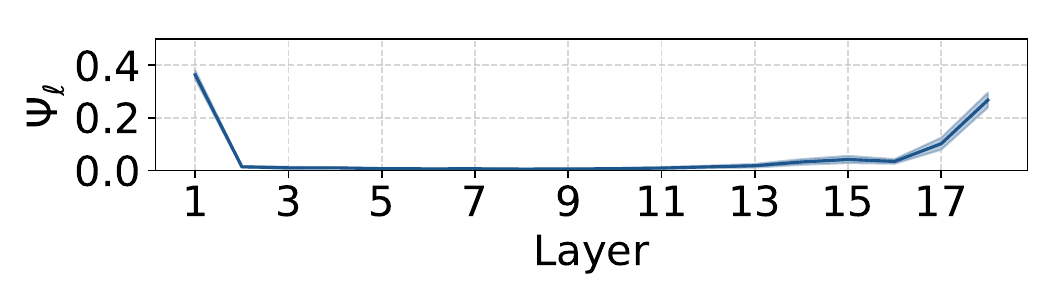}
        \caption{Gemma-2B}
        \label{fig:gemma2b}
    \end{subfigure}
    \begin{subfigure}[b]{\linewidth}
        \centering
        \includegraphics[width=\linewidth]{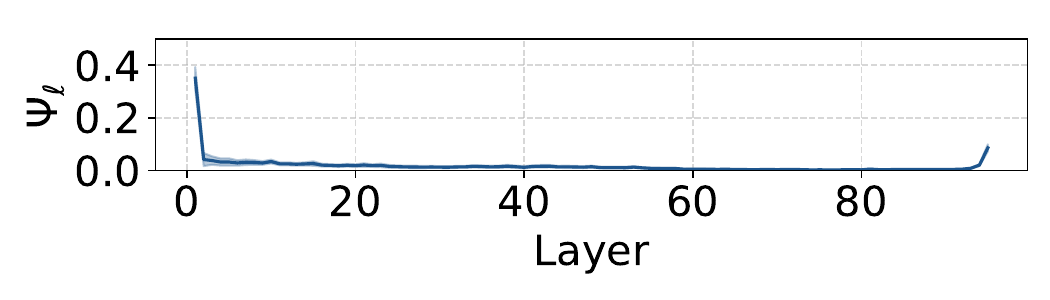}
        \caption{DeepSeek-67B}
        \label{fig:deepseek_67b_arc}
    \end{subfigure}
    \begin{subfigure}[b]{\linewidth}
        \centering
        \includegraphics[width=\linewidth]{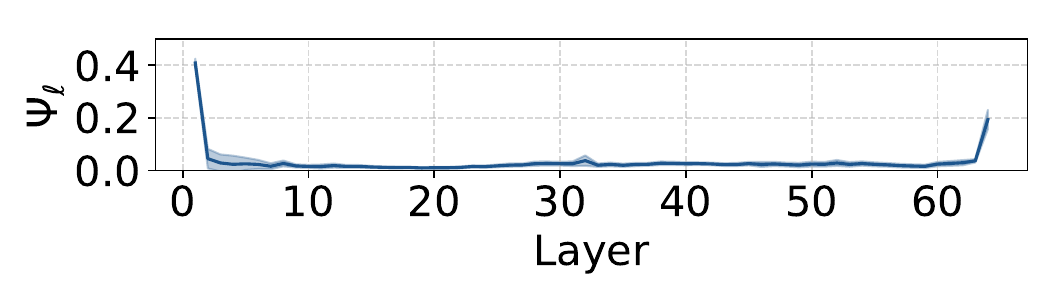}
        \caption{Qwen3-32B}
        \label{fig:qwen3_32b_arc}
    \end{subfigure}
    \begin{subfigure}[b]{\linewidth}
        \centering
        \includegraphics[width=\linewidth]{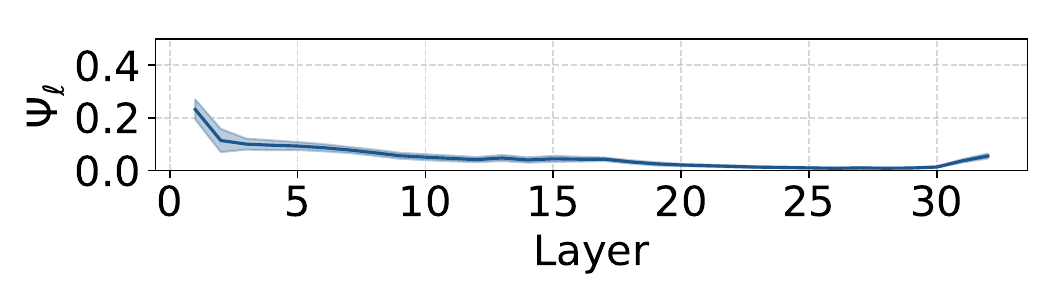}
        \caption{layerskip-llama3.2-8B}
        \label{fig:layerskip-llama3.2-8B_arc}
    \end{subfigure}
    \caption{Displacement on ARC-easy.}
    \label{appendix:trajectory_arc}
\end{figure}

~\cref{appendix:trajectory_arc} shows the results of computing displacement using ARC-Easy.

\section{Non-Final Hidden State Jump Models}

\begin{figure}[t]
    \centering
    \begin{subfigure}[b]{\linewidth}
        \includegraphics[width=\linewidth]{figure/gemma_7b.pdf}
        \caption{Gemma-7B}
    \end{subfigure}
    \begin{subfigure}[b]{\linewidth}
        \includegraphics[width=\linewidth]{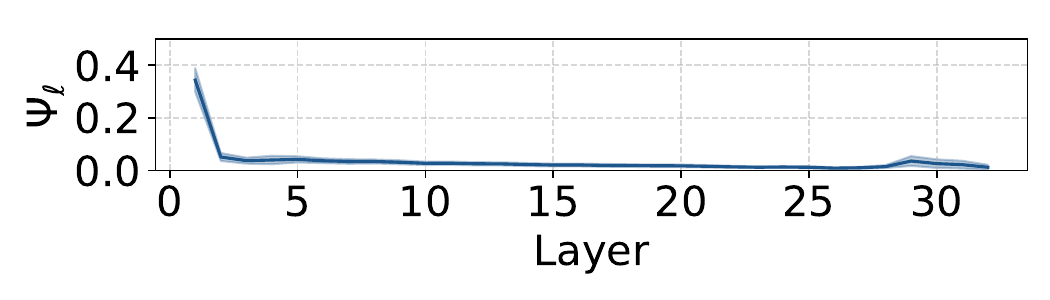}
        \caption{Phi2}
    \end{subfigure}
    \caption{Layer-wise hidden state displacement $\Psi_{\ell}$.}
    \label{fig:enter-label}
\end{figure}

In this section, we examine two models, Gemma-7B and Phi2~\citep{javaheripi2023phi2}, that exhibit hidden states jumps in layers other than the final one (see \cref{fig:enter-label}). In Gemma-7B, the largest jump occurs at the penultimate layer, while Phi2 shows a subtler jump at the third-to-last layer. We attribute these behaviors to each model’s atypical training strategy. Gemma-7B gradually increases the proportion of high-quality, internally filtered data as training nears completion, whereas Phi2 initializes from the weights of a smaller, pretrained model. Thus, although these alternative strategies eliminate the pronounced final layer hidden state jump, they do not prevent all near final layer hidden state jumps.

\section{Regularizing Only The Final Layer Displacement}
\label{Regularizing only the final layer displacement}

\begin{figure}[ht]
    \centering
    \includegraphics[width=\linewidth]{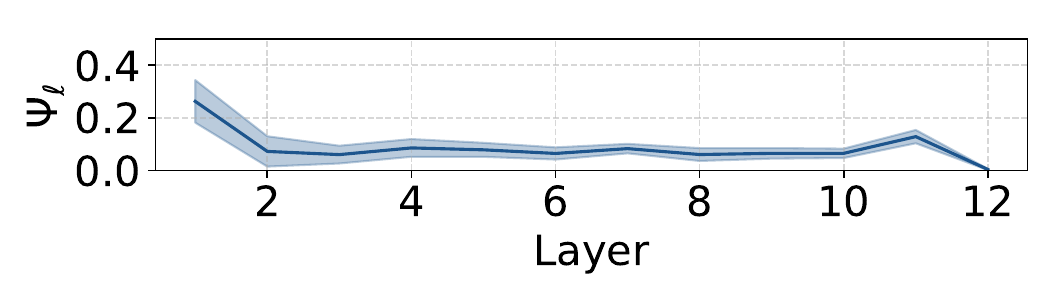}
    \caption{Hidden state displacement trajectory when applying regularization only to the final layer.}\label{fig:Llama170M_last_cos_displacement}
\end{figure}

While our main experiments use a weighted loss function that applies penalties across all layers with exponentially increasing weights (\cref{eq:L_cos}-\cref{eq:proposed_loss}), we also explored a simplified variant that focuses exclusively on the final layer displacement. In this variant, we replace the $\mathcal{L}_{\mathrm{disp}}$ term in \cref{eq:proposed_loss} with:
\begin{align}
\mathcal{L}_{\mathrm{disp}} = \Psi_{L}
\label{eq:only_last}
\end{align}
This approach directly penalizes only the final layer displacement, ignoring middle layers.
\cref{eq:only_last} are equivalent to $\alpha \to \infty$ in $w_{\ell}$ (\cref{eq:L_cos}).
As shown in \cref{fig:Llama170M_last_cos_displacement}, while this approach successfully suppresses the displacement at the final layer ($\Psi_{L}$), the displacement ``jump'' shifts to the penultimate layer, resulting in an increased $\Psi_{L-1}$ value.
Our hypothesis is that overreliance on a specific layer (the final layer in open-weight models) gives rise to redundancy in the middle layers. 
However, in \cref{eq:only_last}, the point of overreliance merely shifts from the final layer to the penultimate one, making it an inappropriate setting for testing this hypothesis.

\section{Impact on Training Time and Memory}
\label{sec:impact-on-training-time-and-memory}
We analyze the change in training complexity introduced by JREG from the perspectives of training time and memory consumption during training.
\cref{tab:training_time} shows a comparison of training time between the baseline and JREG ($\lambda=1.0$, $\alpha=1.0$) across different model sizes.
As shown in the table, no significant change in training time is observed.

Next, \cref{tab:memory_increase_amount} shows the increase ratio of memory consumption per step relative to the baseline.
Memory consumption increases with larger model sizes.
This is because, when computing \cref{eq:L_cos}, the hidden states of each layer must be temporarily stored, and models with more layers require storing a greater number of hidden states.

In summary, while the introduction of JREG leads to an increase in memory consumption during training, it does not incur a substantial change in training time.
\input{table/table_training_time}
\input{table/table_memory_increase_amount}

\section{Hyperparameter Tuning for JREG}
\label{appendix:Hyperparameter Tuning for JREG}
\begin{table*}[t]
    \centering
    \small
    \begin{tabular}{c c|ccccccccc|c}
        \toprule
        $\lambda$ & $\alpha$ & ARC-e & BoolQ & HellaSwag & LAMBADA & PIQA & RACE & SocialIQA & SciQ & SWAG & \textbf{avg}\\
        \midrule
         & 0.0 & 55.7 & 58.3 & 32.4 & 32.5 & 65.9 & 29.6 & 38.6 & 81.3 & 40.2 & 48.3 \\
         & 0.1 & 56.1 & 60.5 & 32.2 & 29.8 & 65.6 & 29.6 & 39.2 & 80.9 & 39.9 & 48.2\\
         & 0.5 & 57.0 & 57.1 & 32.5 & 30.7 & 65.0 & 29.3 & 38.6 & 82.2 & 40.0 & 48.0\\
        \multirow{-4}{*}{1.0} & 1.0 & 57.2 & 60.0 & 32.1 & 31.7 & 65.2 & 29.9 & 38.8 & 81.1 & 40.2 & \textbf{48.5} \\
        \midrule
         & 0.0 & 54.7 & 60.9 & 32.2 & 31.7 & 65.1 & 29.1 & 37.9 & 81.3 & 40.1 & 48.1\\
         & 0.1 & 56.6 & 61.1 & 32.0 & 31.9 & 65.7 & 28.2 & 38.3 & 80.9 & 40.0 & 48.3 \\
         & 0.5 & 55.7 & 56.8 & 31.8 & 31.0 & 65.6 & 29.3 & 37.8 & 82.7 & 39.7 & 47.8\\
        \multirow{-4}{*}{2.0} & 1.0 & 57.1 & 56.4 & 32.3 & 32.8 & 65.9 & 30.1 & 39.5 & 82.5 & 40.1 & \textbf{48.5} \\
        \midrule
         & 0.0 & 56.4 & 58.7 & 31.8 & 32.2 & 64.8 & 30.1 & 38.2 & 81.5 & 39.9 & 48.2 \\
         & 0.1 & 56.8 & 58.8 & 31.9 & 30.8 & 64.9 & 29.2 & 38.3 & 81.3 & 39.8 & 48.0 \\
         & 0.5 & 57.3 & 57.0 & 32.1 & 30.7 & 65.5 & 30.2 & 39.2 & 82.1 & 39.8 & 48.2 \\
        \multirow{-4}{*}{3.0} & 1.0 & 57.2 & 58.9 & 32.0 & 31.0 & 65.3 & 29.1 & 38.8 & 81.7 & 40.0 & 48.2 \\
        \bottomrule
    \end{tabular}
    \caption{Downstream task performance of the 170M model evaluated across 12 combinations of JREG hyperparameters $\lambda \in \{1.0, 2.0, 3.0\}$ and $\alpha \in \{0.0, 0.1, 0.5, 1.0\}$.}
    \label{tab:appendix_other_lambda}
\end{table*}

The JREG loss function involves two hyperparameters, $\lambda$ and $\alpha$ (see Eq.~\cref{eq:L_cos},~\cref{eq:proposed_loss}).  
While Section~\cref{Results} of the main paper reported results only for $\lambda=1.0$, this appendix evaluates 12 combinations of $\lambda \in \{1.0, 2.0, 3.0\}$ and $\alpha \in \{0.0, 0.1, 0.5, 1.0\}$ using a 170M model to identify the optimal $\lambda$.  
The results are summarized in~\cref{tab:appendix_other_lambda}.

The highest average performance was achieved with $(\lambda, \alpha)=(1.0, 1.0)$ and $(2.0, 1.0)$, and when averaging over $\alpha$ for each $\lambda$, the best overall performance occurs at $\lambda=1.0$.
Consequently, the main paper fixes $\lambda=1.0$ and conducts additional experiments with $\alpha$.  

\section{Statistical Evaluation of Performance under Random Seed Perturbations}
\label{appendix:Statistical Evaluation of Performance under Random‐Seed Perturbations}

~\cref{Results} shows the performance differences on the downstream task, and we further investigated their statistical significance by repeating the experiments with different random seeds.

\subsection{Pre-trained Model Evaluation}
\label{Pre-trained Model Evaluation}
\input{table/statistical_check_on_170m_pretrain_model}

In this subsection, we focus on the 170M pre-trained model, corresponding to the experiments presented in~\cref{Pretraining on 170M model}.
The model architecture is shown in~\cref{tab:model_settings}, and the training configuration was identical to that in~\cref{tab:training_settings}, except for the random seed, which was varied across 10 settings.
We used the JREG hyperparameters $\lambda=1.0$ and $\alpha=1.0$, which yielded the highest downstream task performance for the 170M models (\cref{tab:eval_scratch}).

\cref{tab:appendix_statistical_significance_on_fineweb} shows the downstream task performance of the models trained with each seed.
To test whether the mean difference ($\mu_{\mathrm{JREG}}-\mu$) is positive, we conducted a one-sample, one-side $t$-test.
The mean performance difference was 0.41, and the p-value was 0.034, indicating statistical significance at the 5\% level. 
The lower bound of the 95\% one-sided confidence interval was 0.047.
Therefore, although the model’s performance improvement is modest, it was shown to be statistically significant.

\subsection{SFT Model Evaluation}
\label{SFT Model Evaluation}
\input{table/statistical_check_on_4b_sft_model}
In this subsection, we focus on the 3.4B model after SFT, corresponding to the experiments presented in ~\cref{Supervised fine-tuning on 3.4B model}.
SFT configuration was identical to~\cref{appendix:SFT settings} except for the random seed, which was varied across 5 settings.

~\cref{tab:statistical_check_sft} shows the downstream task performance of the models trained with each seed.
The statistical testing procedure is the same as in~\cref{Pre-trained Model Evaluation}.
The mean performance difference was 0.46, and the p-value was $9.89\times10^{-6}$, indicating statistical significance at the 5\% level. 
The lower bound of the 95\% one-sided confidence interval was 0.42.
Therefore, JREG consistently yields statistically significant improvements in the 3.4B model after SFT.

\section{Transition of Jump Rate During Training}
In this section, we analyze the Jump rate ($\zeta_{L}$, $\zeta_{L-1}$, $\zeta_{L-2}$) and displacement $\Psi_{\ell}$ during training
at checkpoints corresponding to 25\%, 50\%, 75\%, and 100\% of training, i.e., at 50K, 100K, 150K, and 200K steps.

\cref{tab:appendix_jump_rate_table} shows the values of $\zeta_{L}$, $\zeta_{L-1}$, and $\zeta_{L-2}$computed at each checkpoint,
while \cref{appendix:170m_ckpt_wise_displacement}, \cref{appendix:1b_ckpt_wise_displacement}, and \cref{appendix:3.4b_ckpt_wise_displacement} show the displacement $\Psi_{\ell}$ for the 170M, 1B, and 3.4B models.
Similar to the characteristics observed in the open-weight models analyzed in \ref{Final layer displacement jump}, the baseline exhibits a monotonically increasing Jump rate as training progresses across checkpoints.
In contrast, for models with JREG applied (with $\alpha \geq 0.5$ for the 170M model), we observe $\zeta_{L} = 0.00$,
indicating that the training dynamics are altered.

\input{table/appendix_jump_rate_table}
\input{latex/figure/appendix_170m_ckpt_wise_displacement}
\input{latex/figure/appendix_1b_ckpt_wise_displacement}
\input{latex/figure/appendix_3.4B_ckpt_wise_displacement}

\section{Displacement across Different Weight Settings}
\label{Displacement across Different Weight Settings}
\input{latex/figure/appendix_displacement}
In this section, we evaluate the degree of redundancy reduction in the middle layers on 170M and 1B models across different weight settings using $\Delta_{\ell}$ metrics ~\cref{eq:redundancy}.

\begin{figure*}[t]
    \centering
    \begin{subfigure}[b]{0.48\linewidth}
        \centering
        \includegraphics[width=\linewidth]{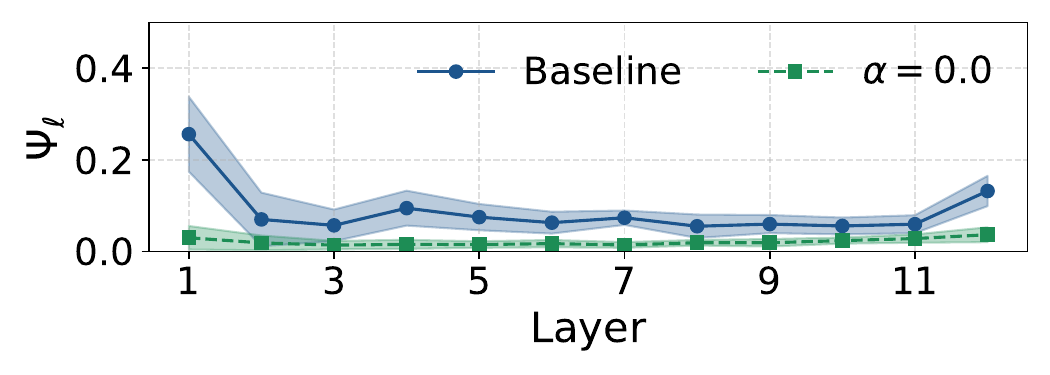}
        \caption{170M model ($\alpha=0.0$)}
    \end{subfigure}
    \begin{subfigure}[b]{0.48\linewidth}
        \centering
        \includegraphics[width=\linewidth]{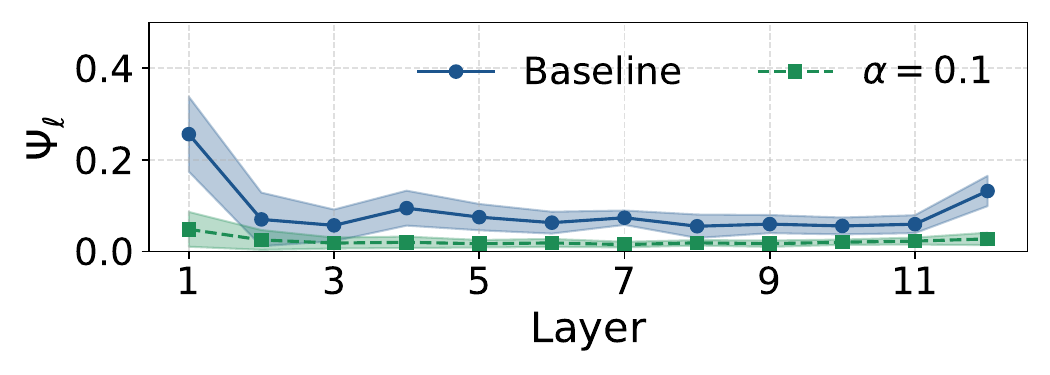}
        \caption{170M model ($\alpha=0.1$)}
    \end{subfigure}
    \begin{subfigure}[b]{0.48\linewidth}
        \centering
        \includegraphics[width=\linewidth]{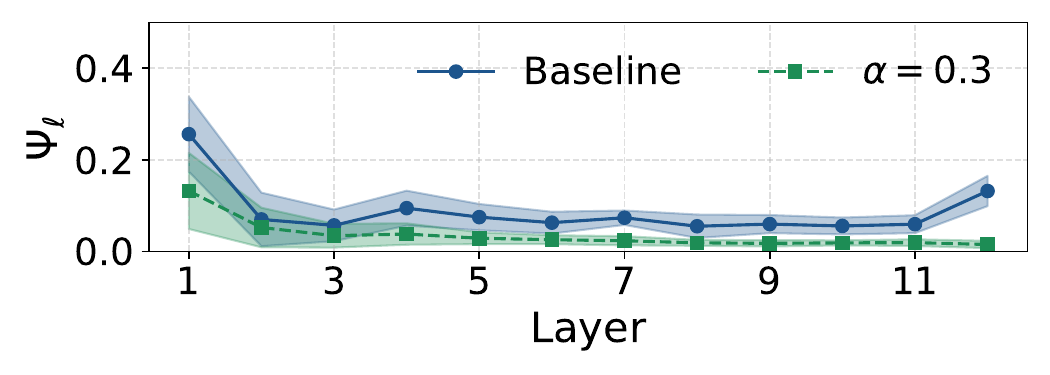}
        \caption{170M model ($\alpha=0.3$)}
    \end{subfigure}
    \begin{subfigure}[b]{0.48\linewidth}
        \centering
        \includegraphics[width=\linewidth]{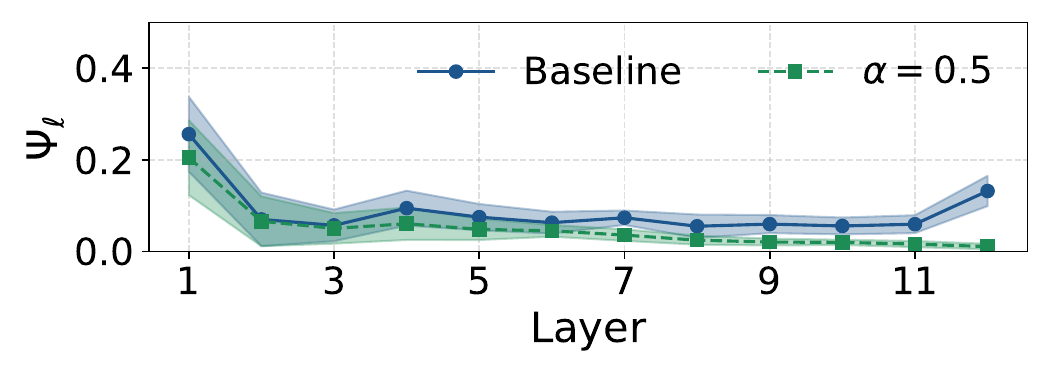}
        \caption{170M model ($\alpha=0.5$)}
    \end{subfigure}
    \begin{subfigure}[b]{0.48\linewidth}
        \centering
        \includegraphics[width=\linewidth]{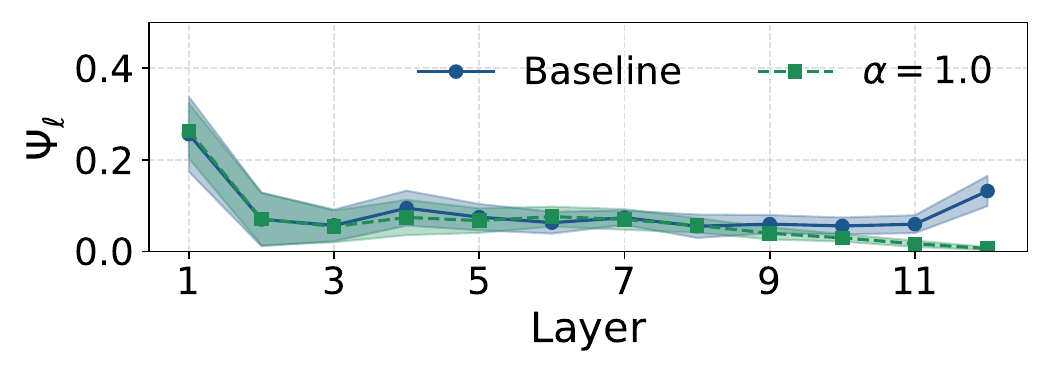}
        \caption{170M model ($\alpha=1.0$)}
    \end{subfigure}
    \begin{subfigure}[b]{0.48\linewidth}
        \centering
        \includegraphics[width=\linewidth]{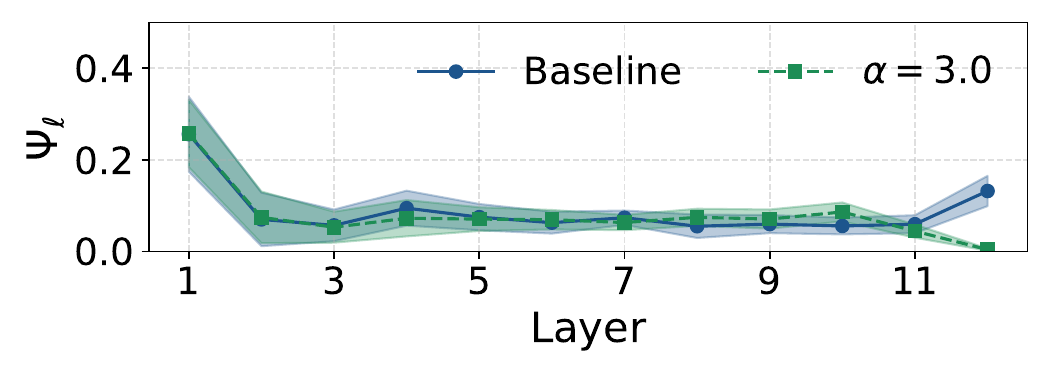}
        \caption{1B model ($\alpha=1.0$)}
    \end{subfigure}
    \begin{subfigure}[b]{0.48\linewidth}
        \centering
        \includegraphics[width=\linewidth]{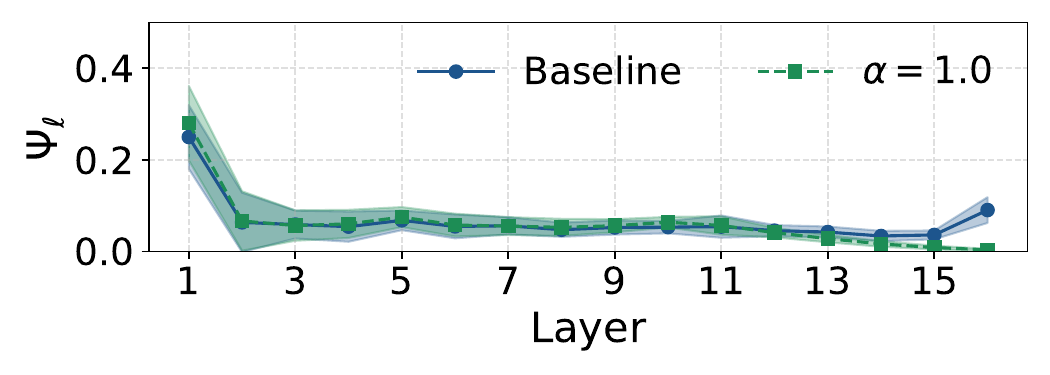}
        \caption{1B model ($\alpha=1.0$)}
        \label{fig:Llama170M_pretrain_displacement_alpha_1.0}
    \end{subfigure}
    \begin{subfigure}[b]{0.48\linewidth}
        \centering
        \includegraphics[width=\linewidth]{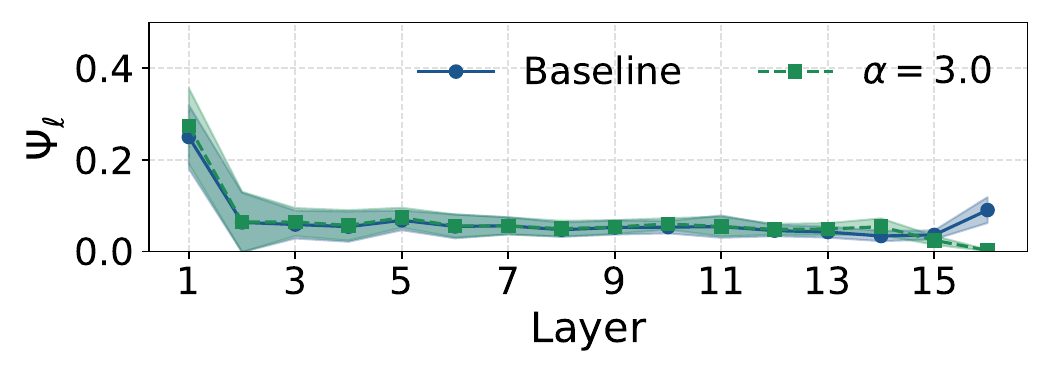}
        \caption{1B model ($\alpha=3.0$)}
        \label{fig:Llama170M_pretrain_displacement_alpha_3.0}
    \end{subfigure}
    \hfill
    \begin{subfigure}[b]{0.48\linewidth}
        \centering
        \includegraphics[width=\linewidth]{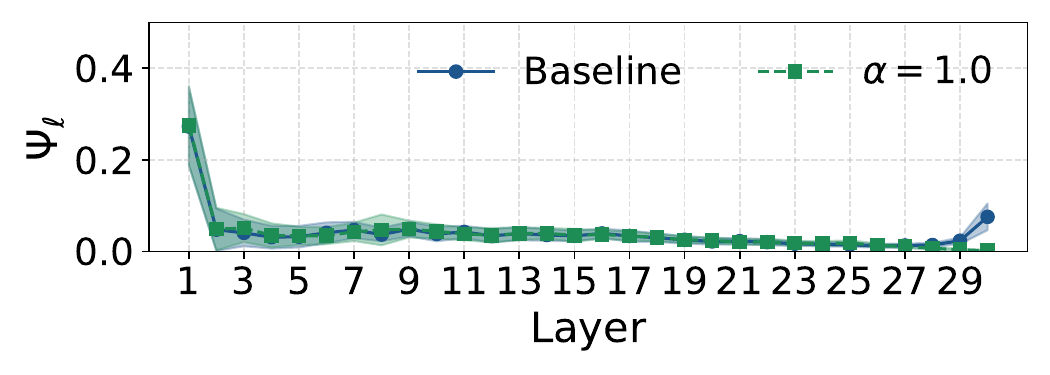}
        \caption{3.4B Model}
        \label{fig:llama1B_pretrain_displacement}
    \end{subfigure}
    \caption{Layer-wise displacement $\Psi_{\ell}$. 
    Blue line and green line represent the results for baseline and JREG, respectively.}
    \label{fig:llama_pre_trajetory}
\end{figure*}

\Cref{fig:llama_pre_trajetory} shows $\Psi_{\ell}$ in graph and~\cref{tab:displacement_small_model} shows $\Psi_{\ell}$ and $\Delta_{\ell}$ for all $\alpha$ settings on 170M and 1B models.
In the 1B model, both $\alpha=1.0$ and $\alpha=3.0$ settings, the displacement in the middle layers tends to be larger ($\Delta_{\ell} > 0$), a similar trend is observed in the 3.4B model (\cref{Analysis and Discussion}).
For $\alpha=1.0$, excluding layer 3 and 7, from layers 1 through 11 exhibit greater displacement under JREG, 
For $\alpha=3.0$, excluding layer 7, from layers 1 through 14 exhibit greater displacement under JREG.
However, in the 170M model, the effect of JREG is less pronounced, and the displacement increase is observed in fewer layers.
Since the 170M model has only 12 layers, fewer than the 1B (16 layers) and 3.4B models (30 layers), the proportion of layers where $\Delta_{\ell} > 0$ is smaller compared to those larger models.

\section{Result on Other Datasets}
\label{appendix:Result on Other Datasets}

\input{table/acc_llama_pretrain_with_web_organizer}
\begin{table*}[t]
    \centering
    \small
    \begin{tabular}{c c|ccccccccc|c}
        \toprule
         seed & Method & ARC-e & BoolQ & HellaSwag & LAMBADA & PIQA & RACE & SocialIQA & SciQ & SWAG & \textbf{avg} \\
         \midrule
          & Baseline 
         & 43.8 & 60.1 & 31.4 & 37.4 & 66.9 & 28.7 & 38.6 & 76.9 & 41.4 & 46.7\\
         \multirow{-2}{*}{123} & JREG & 43.6 & 61.0 & 31.1 & 37.8 & 65.8 & 75.8 & 41.4 & 79.3 & 41.8 & \textbf{47.5} \\
         \midrule
         & Baseline 
         & 43.5 & 55.4 & 31.3 & 37.9 & 66.1 & 28.6 & 37.7 & 75.8 & 41.4 & 46.4\\
         \multirow{-2}{*}{777} & \gc{JREG} & \gc{42.7} & \gc{61.0} & \gc{31.4} & \gc{37.6} & \gc{64.6} & \gc{30} & \gc{37.9} & \gc{75.7} & \gc{41.8} & \gc{\textbf{47.0}} \\
         \midrule
         & Baseline 
         & 43 & 53.7 & 31.5 & 37.6 & 64.7 & 28.3 & 37.2 & 78.0 & 41.6 & 46.2 \\
         \multirow{-2}{*}{888} & \gc{JREG} & \gc{44.3} & \gc{57.7} & \gc{31.3} & \gc{38.3} & \gc{65.6} & \gc{27.9} & \gc{37.9} & \gc{77.5} & \gc{41.7} & \gc{\textbf{46.9}} \\
         \midrule
         & Baseline 
         & 43.4 & 55.3 & 31.2 & 36.3 & 65.5 & 28.5 & 37.2 & 80.4 & 41.6 & 46.6 \\
         \multirow{-2}{*}{2025} & \gc{JREG} & \gc{43.4} & \gc{59.2} & \gc{31.6} & \gc{37.1} & \gc{65.0} & \gc{29.1} & \gc{38.4} & \gc{76.3} & \gc{41.6} & \gc{\textbf{46.9}} \\
         \midrule
         & Baseline 
         & 44.7 & 56.7 & 31.6 & 37.5 & 66.6 & 27.9 & 37.0 & 78.8 & 41.2 & 46.9 \\
         \multirow{-2}{*}{10000} & \gc{JREG} & \gc{44.9} & \gc{59.4} & \gc{31.3} & \gc{36.6} & \gc{65.1} & \gc{28.7} & \gc{38.6} & \gc{78.3} & \gc{41.3} & \gc{\textbf{47.1}} \\
         \bottomrule
    \end{tabular}
    \caption{Results of evaluating downstream task performance for a 170M parameter model pre‑trained on the WebOrganizer corpus and trained with five different random seeds.}
    \label{tab:appendix_result_in_weborganizer}
\end{table*}

\input{latex/figure/appendix_170m_web_displacement_table}
\input{table/appendix_web_jump_rate_table}
\input{latex/figure/appendix_web_170m_ckpt_wise_displacement}

To examine the dataset dependency of JREG, we train models using WebOrganizer~\cite{DBLP:journals/corr/abs-2502-10341}.
The base settings of the experiment were identical to those in~\cref{experiments} while the training was conducted on the FineWeb-Edu dataset.

\Cref{tab:acc_llama_pretrain_with_web_orgainzier} shows the downstream task performance.
The average performance gap between the baseline and JREG was smaller when using the WebOrganizer dataset.
When trained on the FineWeb-Edu dataset, the performance gap reached up to 1.0 for $\alpha=1.0$ and $\alpha=3.0$.
In contrast, when trained on the WebOrganizer dataset, the performance gap were 0.6 for $\alpha=1.0$ and 0.3 for $\alpha=3.0$.
As in~\cref{Pre-trained Model Evaluation}, we conducted a statistical test on the performance differences to examine their significance for $\alpha=1.0$.
The testing procedure followed the same protocol as described in~\cref{Pre-trained Model Evaluation}.
The mean performance difference was 0.41, and the p-value was 0.0052, indicating statistical significance at the 5\% level. 
The lower bound of the 95\% one-sided confidence interval was 0.271.
Therefore, the performance improvement achieved by applying JREG is statistically significant regardless of the choice of training dataset.

~\Cref{tab:170m_web_displacement} shows displacement $\Psi_{\ell}$ and $\Delta_{\ell}$,~\cref{tab:appendix_jump_rate_table_web} shows checkpoint-wise jump rates $\zeta_{L}$, $\zeta_{L-1}$, $\zeta_{L-2}$, and~\cref{appendix:170m_ckpt_wise_displacement} shows checkpoint-wise displacement.
Similar to the results using the FineWeb-Edu dataset, $\zeta_L$ remains at 0.00 for $\alpha \geq 0.3$, and the proportion of layers with $\Delta_{\ell} > 0$ increases as $\alpha$ increases.

\section{Layer Skip Evaluation}
\begin{figure*}[t]
    \centering
    \begin{subfigure}[b]{0.48\linewidth}
        \centering
        \includegraphics[width=\linewidth]{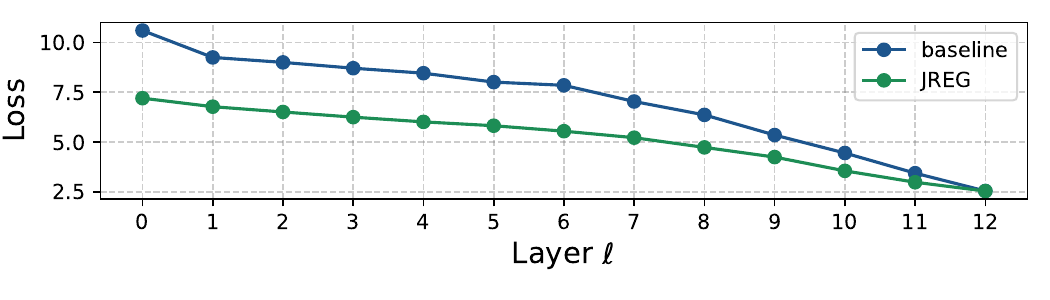}
        \caption{170M model ($\alpha=0.0$)}
    \end{subfigure}
    \begin{subfigure}[b]{0.48\linewidth}
        \centering
        \includegraphics[width=\linewidth]{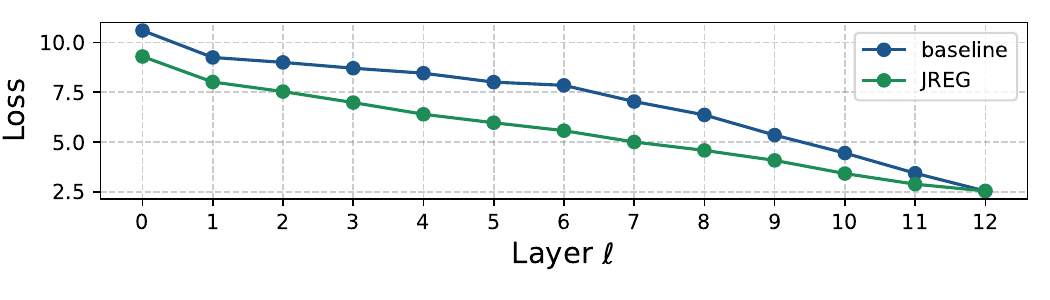}
        \caption{170M model ($\alpha=0.1$)}
    \end{subfigure}
    \begin{subfigure}[b]{0.48\linewidth}
        \centering
        \includegraphics[width=\linewidth]{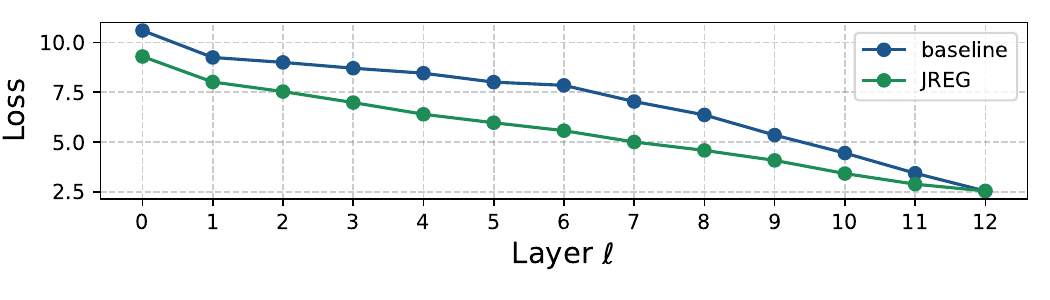}
        \caption{170M model ($\alpha=0.3$)}
    \end{subfigure}
    \begin{subfigure}[b]{0.48\linewidth}
        \centering
        \includegraphics[width=\linewidth]{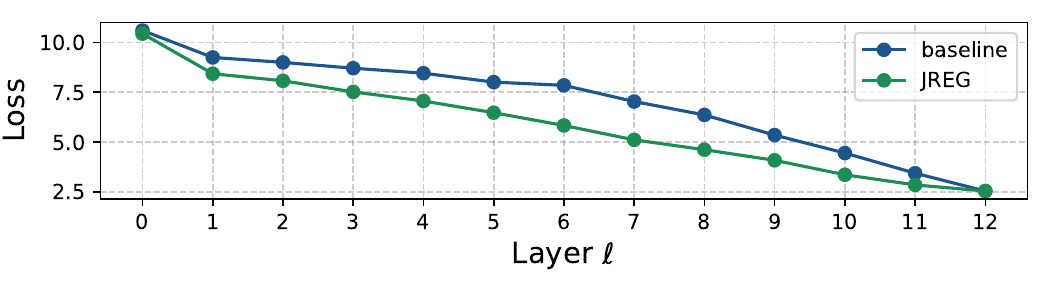}
        \caption{170M model ($\alpha=0.5$)}
    \end{subfigure}
    \begin{subfigure}[b]{0.48\linewidth}
        \centering
        \includegraphics[width=\linewidth]{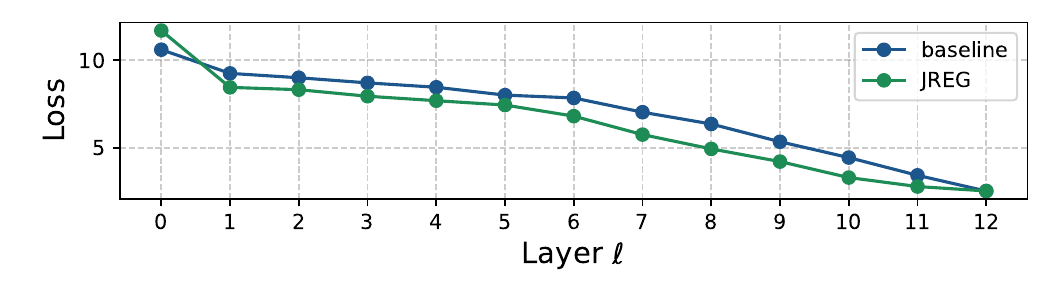}
        \caption{170M model ($\alpha=1.0$)}
    \end{subfigure}
    \begin{subfigure}[b]{0.48\linewidth}
        \centering
        \includegraphics[width=\linewidth]{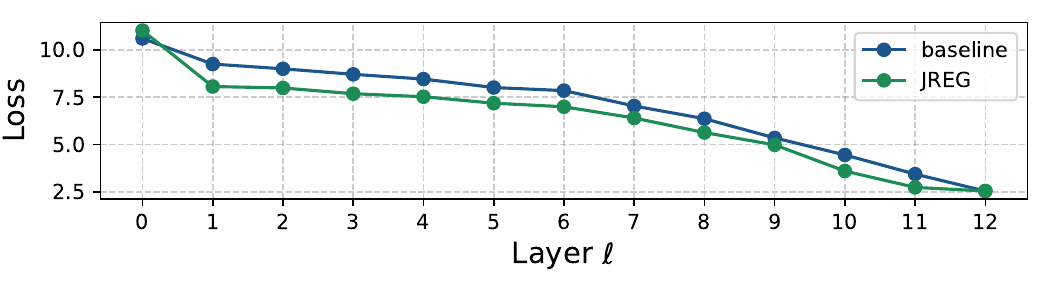}
        \caption{170M model ($\alpha=3.0$)}
    \end{subfigure}
    \begin{subfigure}[b]{0.48\linewidth}
        \centering
        \includegraphics[width=\linewidth]{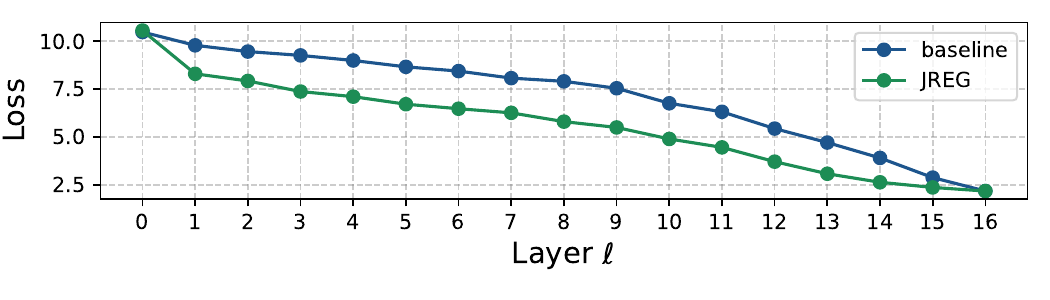}
        \caption{1B model ($\alpha=1.0$)}
    \end{subfigure}
    \begin{subfigure}[b]{0.48\linewidth}
        \centering
        \includegraphics[width=\linewidth]{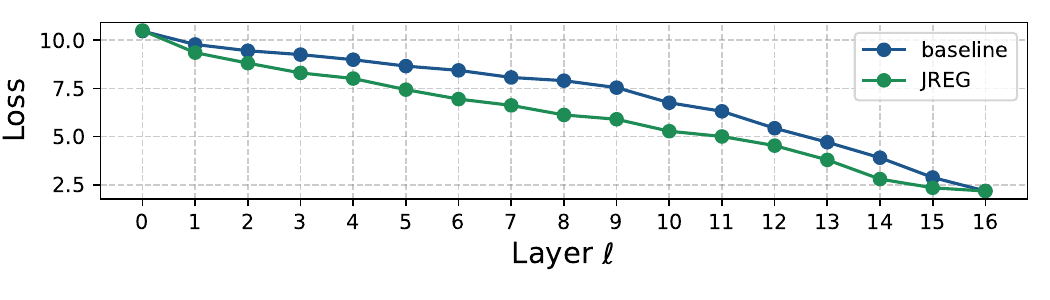}
        \caption{1B model ($\alpha=3.0$)}
    \end{subfigure}
    \begin{subfigure}[b]{0.48\linewidth}
        \centering
        \includegraphics[width=\linewidth]{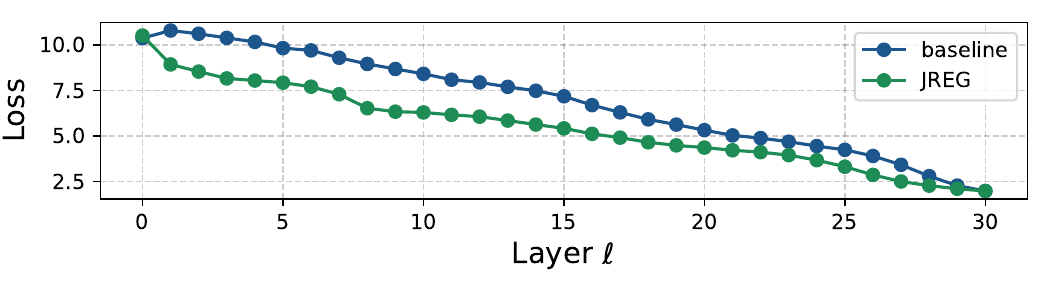}
        \caption{3.4B model ($\alpha=1.0$)}
    \end{subfigure}
    \caption{Validation loss at each layer when inference is stopped at $\vh_{\ell}$. Lower values indicate better performance.}
    \label{fig:early-exit}
\end{figure*}

In \cref{tab:displacement}, we measured the reduction of layer redundancy by the displacement difference $\Delta_{\ell}=\Psi_{\ell}^{\mathrm{JREG}}-\Psi_{\ell}^{\mathrm{Baseline}}$.
In this section, however, we assess redundancy from a different perspective by comparing the extent of performance degradation under layer skip between the baseline and the JREG-trained models.
Layer skip is a technique that accelerates inference by terminating the forward pass at a middle layer and using that layer’s output as the model prediction, instead of propagating to the final layer.
Because JREG is designed to reduce redundancy in the middle layers, we expect it to achieve higher performance than the baseline when inference is stopped at a middle layer.

To test this, we perform layer skip evaluation.
Using the same 170M, 1B 3.4B parameter Llama-based architecture as in, we compare baseline and JREG models across different $\alpha$ settings.
At inference time, instead of taking the final hidden state $\vh_L$, we output the hidden state $\vh_{\ell}$ ($0 \leq \ell \leq L$) from a middle layer and terminate decoding early.
Performance is measured by validation loss on the FineWeb-Edu split.

\cref{fig:early-exit} shows the validation loss when inference is stopped at each layer $\ell$.
Across all middle layers ($0 < \ell < L$), JREG achieves lower validation loss than the baseline. 
This indicates that JREG enhances the quality of middle-layer representations and utilizes parameters more effectively.

\end{document}

%% file: table/displacement.tex
\begin{table*}[t]
    \centering
    \begin{subtable}[t]{\textwidth}
        \centering
        \small
        \tabcolsep 3pt
        \begin{tabular}{crrrrrrrrrrrrrrr}
            \toprule
            Layer index & 1 & 2 & 3 & 4 & 5 & 6 & 7 & 8 & 9 & 10 & 11 & 12 & 13 & 14 & 15 \\
            \midrule
            Baseline    & 27.30 & 4.78 & 4.08 & 3.14 & 3.25 & 4.10 & 4.70 & 3.71 & 4.91 & 3.85 & 4.28 & 3.43 & 3.93 & 3.62 & 3.42 \\
            JREG (ours) & 27.58 & 4.89 & 5.13 & 3.54 & 3.41 & 3.48 & 4.31 & 4.73 & 4.92 & 4.39 & 3.88 & 3.57 & 3.92 & 3.97 & 3.56 \\
            $\Delta_{\ell}$ & \textbf{0.28} & \textbf{0.11} & \textbf{1.05} & \textbf{0.40} & \textbf{0.16} & -0.62 & -0.39 & \textbf{1.02} & \textbf{0.01} & \textbf{0.54} & -0.40 & \textbf{0.14} & -0.01 & \textbf{0.35} & \textbf{0.14} \\
            \bottomrule
        \end{tabular}
        \caption{3.4B Model (Layers 1–15)}
        \label{tab:llama34b_firsthalf}
    \end{subtable}

    \begin{subtable}[t]{\textwidth}
        \centering
        \small
        \tabcolsep 3pt
        \begin{tabular}{crrrrrrrrrrrrrrr}
            \toprule
            Layer index & 16 & 17 & 18 & 19 & 20 & 21 & 22 & 23 & 24 & 25 & 26 & 27 & 28 & 29 & 30 \\
            \midrule
            Baseline    & 3.96 & 3.4 & 3.01 & 2.57 & 2.22 & 2.27 & 2.05 & 1.71 & 1.60 & 1.48 & 1.25 & 1.33 & 1.43 & 2.37 & 7.58 \\
            JREG (ours) & 3.74 & 3.39 & 3.12 & 2.57 & 2.47 & 2.07 & 2.13 & 1.86 & 1.76 & 1.86 & 1.37 & 1.11 & 0.71 & 0.40 & 0.18 \\
            $\Delta_{\ell}$ & -0.22 & -0.01 & \textbf{0.11} & 0.00 & \textbf{0.25} & -0.20 & \textbf{0.08} & \textbf{0.15} & \textbf{0.16} & \textbf{0.38} & \textbf{0.12} & -0.22 & -0.72 & -1.97 & -7.40 \\
            \bottomrule
        \end{tabular}
        \caption{3.4B Model (Layers 16–30)}
        \label{tab:llama34b_secondhalf}
    \end{subtable}
    
    \caption{
        Displacement $\Psi_{\ell}$ for the baseline and JREG ($\alpha=1.0$) on the 3.4B model, and the displacement difference $\Delta_{\ell}$ ($\times 10^{-2}$). 
        $\Delta_{\ell} > 0$ indicates that JREG reduces redundancy at layer $\ell$.
    }
    \label{tab:displacement}
\end{table*}

%% file: table/acc_sft_table.tex
\begin{table}[t]
    \centering
    \small
    \tabcolsep 3pt
    \begin{tabular}{cccc|c}
        \toprule
              & Vicuna & WizardLM & MT &   \\
        Model & Bench & testset & Bench & avg  \\
        \midrule
         Baseline & 5.89 & 4.10 & 2.84 & 4.28 \\
         JREG ($\alpha=1.0$) & \textbf{6.36} & \textbf{4.23} & \textbf{3.40} & \textbf{4.67} \\
        \bottomrule
    \end{tabular}
    \caption{Results of different benchmark performances on 3.4B baseline and JREG ($\alpha=1.0$). Each row shows the score on each benchmark dataset, with its average in the rightmost column.}
    \label{tab:acc_sft_table}
\end{table}

%% file: table/jump_rate_table_llama_sft.tex
\begin{table}[t]
    \centering
    \small
    \begin{tabular}{cccc}
        \toprule
         Model & $\zeta_{L}$ & $\zeta_{L-1}$ & $\zeta_{L-2}$ \\
        \midrule
         Baseline & 5.14 & 6.11 & 6.20 \\
         JREG ($\alpha=1.0$) & \textbf{0.00} & \textbf{0.00} & \textbf{0.00} \\
        \bottomrule
    \end{tabular}
    \caption{Comparison of jump rates between 3.4B baseline and JREG with $\alpha=1.0$ model after SFT. Consistent with the pre-training results, the JREG maintains a jump rate of 0.0 after SFT.}
    \label{tab:jump_rate_table_llama_sft}
\end{table}

%% file: table/table_training_time.tex
\begin{table}[t]
    \centering
    \begin{tabular}{ccc}
        \toprule
        \multicolumn{2}{c}{Model} & Training time (hours)  \\
        \midrule
        & Baseline & 16 \\
        \multirow{-2}{*}{170M} & JREG & 16  \\
        \midrule
        & Baseline & 55 \\
        \multirow{-2}{*}{1B} & JREG & 53 \\
        \midrule
        & Baseline & 159 \\
        \multirow{-2}{*}{3.4B} & JREG & 159 \\
        \bottomrule
    \end{tabular}
    \caption{Comparison of training time between the baseline and JREG ($\lambda=1.0, \alpha=1.0$)}
    \label{tab:training_time}
\end{table}

%% file: table/table_memory_increase_amount.tex
\begin{table}[t]
    \centering
    \begin{tabular}{cc}
        \toprule
        Model Size & Memory Increase Amount  \\
        \midrule
        170M & 8.5\% \\
        1B & 15.5\% \\
        3.4B & 35.6\%\\
        \bottomrule
    \end{tabular}
    \caption{Increase in memory consumption with JREG ($\lambda=1.0, \alpha=1.0$) compared to the baseline}
    \label{tab:memory_increase_amount}
\end{table}

%% file: table/statistical_check_on_170m_pretrain_model.tex
\begin{table*}[t]
    \centering
    \small
    \begin{tabular}{c cccccccccc|c}
        \toprule
         seed & Method & ARC-e & BoolQ & HellaSwag & LAMBADA & PIQA & RACE & SocialIQA & SciQ & SWAG & \textbf{avg} \\
         \midrule
         & Baseline & 56.5 & 61.4 & 31.9 & 32.5 & 65.1 & 28.9 & 38.0 & 80.6 & 40.1 & \textbf{48.3} \\
         \multirow{-2}{*}{42} & \gc{JREG} & \gc{57.0} & \gc{56.2} & \gc{31.9} & \gc{32.6} & \gc{65.5} & \gc{28.6} & \gc{38.6} & \gc{80.9} & \gc{40.0} & \gc{47.9} \\
         \midrule
         & Baseline & 55.6 & 59.1 & 31.9 & 32.9 & 64.0 & 31.6 & 39.3 & 82.1 & 40 & \textbf{48.5}  \\
         \multirow{-2}{*}{123} & \gc{JREG} & \gc{56.4} & \gc{56.0} & \gc{32.2} & \gc{31.3} & \gc{65.1} & \gc{29.8} & \gc{37.5} & \gc{82.1} & \gc{40.0} & \gc{47.8} \\
         \midrule
         & Baseline & 54.9 & 57.5 & 32.1 & 32.0 & 64.0 & 29.1 & 38.4 & 80.2 & 39.7 & 47.5 \\
         \multirow{-2}{*}{777} & \gc{JREG} & \gc{57.2} & \gc{60.0} & \gc{32.1} & \gc{31.7} & \gc{65.2} & \gc{29.9} & \gc{38.8} & \gc{81.1} & \gc{40.2} & \gc{\textbf{48.5}} \\
         \midrule
         & Baseline & 55.4 & 51.1 & 32.2 & 33.0 & 65.7 & 28.6 & 39.0 & 81.2 & 39.8 & 47.3 \\
         \multirow{-2}{*}{888} & \gc{JREG} & \gc{57.3} & \gc{60.8} & \gc{32.1} & \gc{32.1} & \gc{64.4} & \gc{28.0} & \gc{39.4} & \gc{81.6} & \gc{40.1} & \gc{\textbf{48.4}} \\
         \midrule
         & Baseline & 55.4 & 57.9 & 32.3 & 32.1 & 65.0 & 28.5 & 39.7 & 80.9 & 40.0 & 48.0 \\
         \multirow{-2}{*}{2025} & \gc{JREG} & \gc{56.5} & \gc{59.7} & \gc{32.0} & \gc{32.4} & \gc{65.3} & \gc{30.1} & \gc{37.5} & \gc{82.1} & \gc{40.0} & \gc{\textbf{48.4}} \\
         \midrule
         & Baseline & 56.6 & 51.6 & 31.9 & 32.2 & 65.1 & 28.9 & 38.1 & 81.9 & 40.0 & 47.4 \\
         \multirow{-2}{*}{10000} & \gc{JREG} & \gc{55.9} & \gc{61.7} & \gc{31.9} & \gc{31.7} & \gc{64.5} & \gc{29.2} & \gc{39.2} & \gc{82.5} & \gc{39.7} & \gc{\textbf{48.5}} \\
         \midrule
         & Baseline & 55.9 & 60.4 & 32.2 & 30.0 & 65.2 & 29.9 & 38.7 & 79.8 & 39.9 & 48.0 \\
         \multirow{-2}{*}{65537} & \gc{JREG} & \gc{57.2} & \gc{61.3} & \gc{32.2} & \gc{31.3} & \gc{65.5} & \gc{30.0} & \gc{38.1} & \gc{82.4} & \gc{40.1} & \gc{\textbf{48.7}} \\
         \midrule
         & Baseline & 56.8 & 55.0 & 32.4 & 32.4 & 64.2 & 30.9 & 38.6 & 81.8 & 40.0 & 48.0 \\
         \multirow{-2}{*}{141421} & \gc{JREG} & \gc{58.2} & \gc{59.4} & \gc{32.3} & \gc{31.7} & \gc{64.3} & \gc{30.0} & \gc{38.9} & \gc{82.1} & \gc{39.8} & \gc{\textbf{48.5}} \\
         \midrule
         & Baseline & 55.2 & 58.7 & 32.4 & 32.4 & 64.9 & 29.5 & 38.4 & 80.9 & 40.3 & 48.1 \\
         \multirow{-2}{*}{271828} & \gc{JREG} & \gc{55.7} & \gc{61.8} & \gc{32.1} & \gc{32.2} & \gc{65.6} & \gc{29.5} & \gc{37.7} & \gc{82.4} & \gc{40.1} & \gc{\textbf{48.6}} \\
         \midrule
         & Baseline & 54.4 & 56.7 & 32.0 & 32.1 & 65.8 & 28.3 & 39.3 & 81.4 & 40.0 & \textbf{47.8} \\
         \multirow{-2}{*}{314159} & \gc{JREG} & \gc{55.9} & \gc{53.9} & \gc{32.1} & \gc{33.1} & \gc{64.1} & \gc{29.7} & \gc{38.4} & \gc{82.3} & \gc{39.9} & \gc{47.7} \\
         \bottomrule
    \end{tabular}
    \caption{Results of evaluating downstream task performance for a 170M parameter model pre‑trained on the Fineweb-edu corpus and trained with ten different random seeds.}
    \label{tab:appendix_statistical_significance_on_fineweb}
\end{table*}

%% file: table/statistical_check_on_4b_sft_model.tex
\begin{table*}[t]
    \centering
    \begin{tabular}{cccccc}
        \toprule
         seed & Method & Vicuna-Bench & WizardLM testset & MT-bench & avg \\
        \midrule
         \multirow{2}{*}{123} & Baseline & 5.80 & 4.00 & 2.91 & 4.24 \\
         & JREG & \textbf{6.39} & \textbf{4.32} & \textbf{3.42} & \textbf{4.71} \\
        \midrule
         \multirow{2}{*}{777} & Baseline & 5.89 & 4.10 & 2.84 & 4.28 \\
          & JREG & \textbf{6.36} & \textbf{4.24} & \textbf{3.40} & \textbf{4.67} \\
        \midrule
         \multirow{2}{*}{888} & Baseline & 5.81 & 3.96 & 2.82 & 4.20 \\
         & JREG & \textbf{6.34} & \textbf{4.35} & \textbf{3.39} & \textbf{4.70} \\
        \midrule
         \multirow{2}{*}{2025} & Baseline & 5.89 & 4.03 & 2.88 & 4.27 \\
         & JREG & \textbf{6.36} & \textbf{4.36} & \textbf{3.42} & \textbf{4.71} \\
        \midrule
         \multirow{2}{*}{10000} & Baseline & 5.78 & 3.94 & 2.92 & 4.21 \\
         & JREG & \textbf{6.31} & \textbf{4.37} & \textbf{3.42} & \textbf{4.70} \\
        \bottomrule
    \end{tabular}
    \caption{Results of evaluating different benchmarks for the 3.4B-parameter model after SFT across five random seeds.}
    \label{tab:statistical_check_sft}
\end{table*}

%% file: table/appendix_jump_rate_table.tex
\begin{table*}[t]
    \centering
    \input{table/jump_rate_table_llama_0}
    \input{table/jump_rate_table_llama_1}
    \input{table/jump_rate_table_llama_2}

    \caption{Checkpoint-wise jump rate of models pre-trained on Fineweb-edu}
    \label{tab:appendix_jump_rate_table}
\end{table*}

%% file: table/jump_rate_table_llama_0.tex
\begin{subtable}[t]{\textwidth}
  \centering
  \small
  \begin{tabular}{cccrrrr}
    \toprule
    \multicolumn{2}{c}{Model} & $\alpha$ & 50k & 100k & 150k & 200k \\
    \midrule
    & Baseline & - & 4.13 & 5.07 & 5.91 & 7.24 \\
    & & 0.0 & 0.48 & 0.38 & 0.55 & 0.84 \\
    & & 0.1 & 0.10 & 0.25 & 0.29 & 0.48  \\
    & & 0.3 & \textbf{0.00} & \textbf{0.00} & \textbf{0.00} & \textbf{0.00} \\
    & & 0.5 & \textbf{0.00} & \textbf{0.00} & \textbf{0.00} & \textbf{0.00} \\
    & & 1.0 & \textbf{0.00} & \textbf{0.00} & \textbf{0.00} & \textbf{0.00} \\
      \multirow{-6}{*}{170M} 
    & \multirow{-6}{*}{JREG} 
    & 3.0 
    & \textbf{0.00} & \textbf{0.00} & \textbf{0.00} & \textbf{0.00} \\
    \midrule
    & Baseline & - & 2.07 & 2.96 & 4.13 & 5.42 \\
      & & 1.0 
      & \textbf{0.00} & \textbf{0.00} & \textbf{0.00} & \textbf{0.00} \\
      \multirow{-3}{*}{1B} & \multirow{-2}{*}{JREG} 
      & 3.0 
      & \textbf{0.00} & \textbf{0.00} & \textbf{0.00} & \textbf{0.00} \\
    \midrule
      & Baseline & - & 1.62 & 2.57 & 3.98 & 5.21 \\
      \multirow{-2}{*}{3.4B} & JREG & 1.0 & \textbf{0.00} & \textbf{0.00} & \textbf{0.00} & \textbf{0.00} \\
    \bottomrule
  \end{tabular}
  \caption{$\zeta_{L}$}
  \label{tab:jump_rate_table_llama_0}
\end{subtable}

%% file: table/jump_rate_table_llama_1.tex
\begin{subtable}[t]{\textwidth}
  \centering
  \small
  \begin{tabular}{cccrrrr}
    \toprule
    \multicolumn{2}{c}{Model} & $\alpha$ & 50k & 100k & 150k & 200k \\
    \midrule
    & Baseline & - & 4.25 & 5.64 & 6.34 & 7.63 \\
    & & 0.0 & 0.56 & 0.56 & 0.87 & 1.28 \\
    & & 0.1 & 0.11 & 0.33 & 0.43 & 0.72 \\
    & & 0.3 & \textbf{0.00} & 0.03 & 0.00 & 0.05 \\
    & & 0.5 & \textbf{0.00} & \textbf{0.00} & \textbf{0.00} & \textbf{0.00} \\
    & & 1.0 & \textbf{0.00} & \textbf{0.00} & \textbf{0.00} & \textbf{0.00} \\
      \multirow{-6}{*}{170M} 
    & \multirow{-6}{*}{JREG} 
    & 3.0 
    & \textbf{0.00} & \textbf{0.00} & \textbf{0.00} & \textbf{0.00} \\
    \midrule
    & Baseline & - & 2.16 & 3.16 & 4.27 & 5.66 \\
      & & 1.0 
      & \textbf{0.00} & \textbf{0.00} & \textbf{0.00} & \textbf{0.00} \\
      \multirow{-3}{*}{1B} & \multirow{-2}{*}{JREG} 
      & 3.0 
      & \textbf{0.00} & \textbf{0.00} & \textbf{0.00} & \textbf{0.00} \\
    \midrule
      & Baseline & - & 2.05 & 3.12 & 4.74 & 6.15 \\
      \multirow{-2}{*}{3.4B} & JREG & 1.0 & \textbf{0.00} & \textbf{0.00} & \textbf{0.00} & \textbf{0.00} \\
    \bottomrule
  \end{tabular}
  \caption{$\zeta_{L-1}$}
  \label{tab:jump_rate_table_llama_1}
\end{subtable}

%% file: table/jump_rate_table_llama_2.tex
\begin{subtable}[t]{\textwidth}
  \centering
  \small
  \begin{tabular}{cccrrrr}
    \toprule
    \multicolumn{2}{c}{Model} & $\alpha$ & 50k & 100k & 150k & 200k \\
    \midrule
    & Baseline & - & 4.25 & 5.64 & 6.34 & 7.63 \\
    & & 0.0 & 1.13 & 1.05 & 1.30 & 1.76 \\
    & & 0.1 & 0.53 & 0.67 & 0.72 & 1.05 \\
    & & 0.3 & 0.08 & 0.09 & 0.07 & 0.15 \\
    & & 0.5 & \textbf{0.00} & \textbf{0.00} & \textbf{0.00} & \textbf{0.00} \\
    & & 1.0 & \textbf{0.00} & \textbf{0.00} & \textbf{0.00} & \textbf{0.00} \\
      \multirow{-6}{*}{170M} 
    & \multirow{-6}{*}{JREG} 
    & 3.0 
    & 1.48 & 1.55 & 1.41 & 1.55 \\
    \midrule
    & Baseline & - & 2.16 & 3.16 & 4.27 & 5.66 \\
      & & 1.0 
      & \textbf{0.00} & \textbf{0.00} & \textbf{0.00} & \textbf{0.00} \\
      \multirow{-3}{*}{1B} & \multirow{-2}{*}{JREG} 
      & 3.0 
      & 0.22 & 0.31 & 0.43 & 0.50 \\
    \midrule
      & Baseline & - & 2.11 & 3.22 & 4.85 & 6.25 \\
      \multirow{-2}{*}{3.4B} & JREG & 1.0 & \textbf{0.00} & \textbf{0.00} & \textbf{0.00} & \textbf{0.00} \\
    \bottomrule
  \end{tabular}
  \caption{$\zeta_{L-2}$}
  \label{tab:jump_rate_table_llama_2}
\end{subtable}

%% file: latex/figure/appendix_170m_ckpt_wise_displacement.tex
\begin{figure}[t]
    \centering
    \begin{subfigure}[b]{\linewidth}
        \centering
        \includegraphics[width=\linewidth]{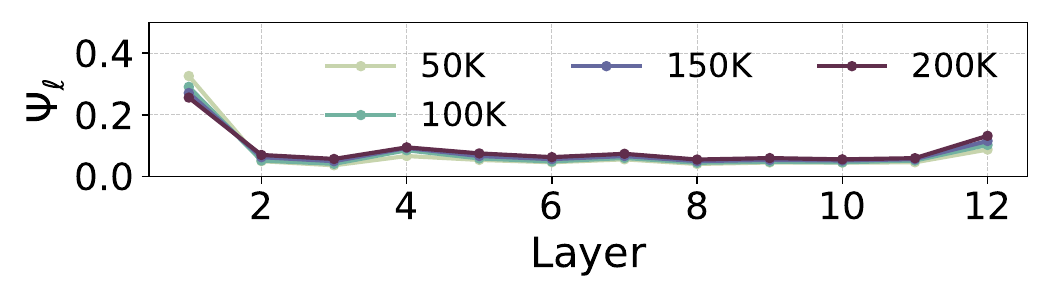}
        \caption{Baseline}
        \label{fig:170m_baseline_displacement}
    \end{subfigure}
    \begin{subfigure}[b]{\linewidth}
        \centering
        \includegraphics[width=\linewidth]{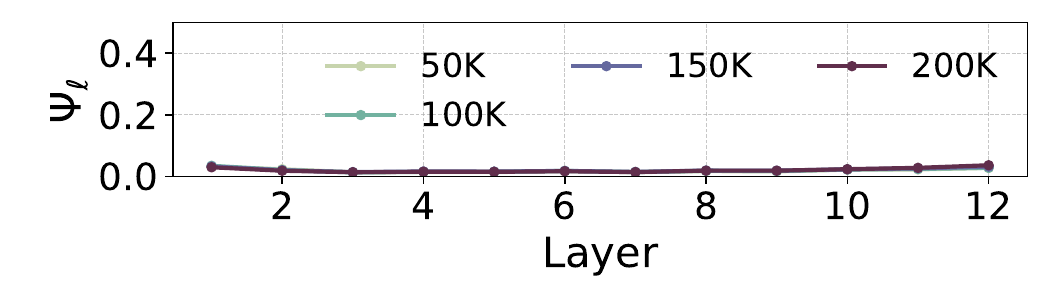}
        \caption{JREG ($\alpha=0.0$)}
        \label{fig:170m_alpha_0.0_displacement}
    \end{subfigure}
    \begin{subfigure}[b]{\linewidth}
        \centering
        \includegraphics[width=\linewidth]{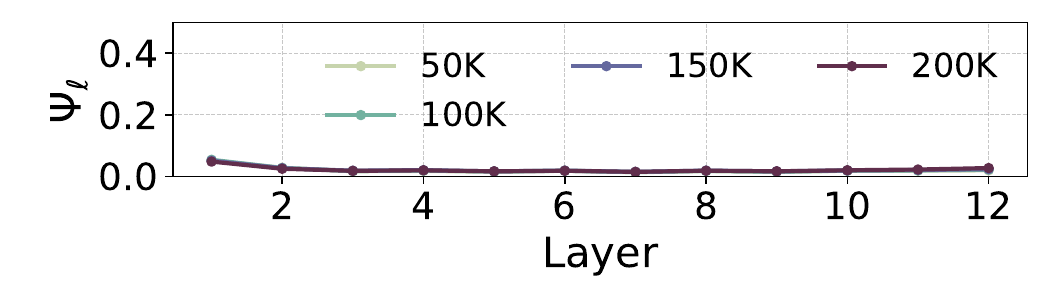}
        \caption{JREG ($\alpha=0.1$)}
        \label{fig:170m_alpha_0.1_displacement}
    \end{subfigure}
    \begin{subfigure}[b]{\linewidth}
        \centering
        \includegraphics[width=\linewidth]{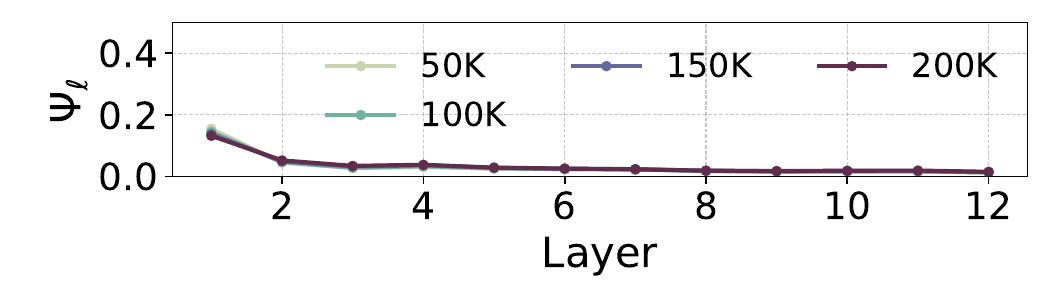}
        \caption{JREG ($\alpha=0.3$)}
        \label{fig:170m_alpha_0.3_displacement}
    \end{subfigure}
    \begin{subfigure}[b]{\linewidth}
        \centering
        \includegraphics[width=\linewidth]{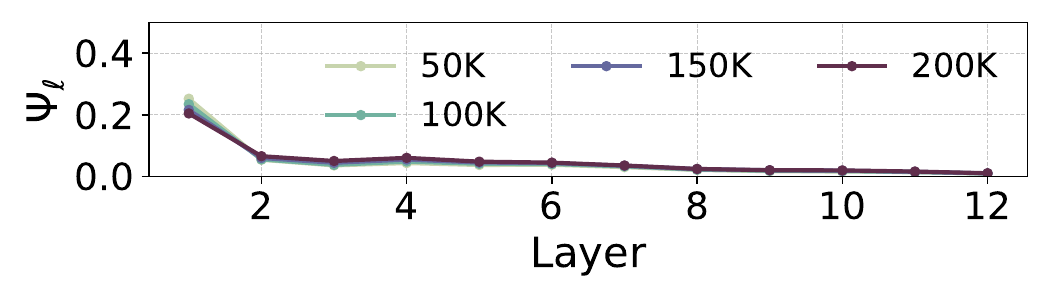}
        \caption{JREG ($\alpha=0.5$)}
        \label{fig:170m_alpha_0.5_displacement}
    \end{subfigure}
    \begin{subfigure}[b]{\linewidth}
        \centering
        \includegraphics[width=\linewidth]{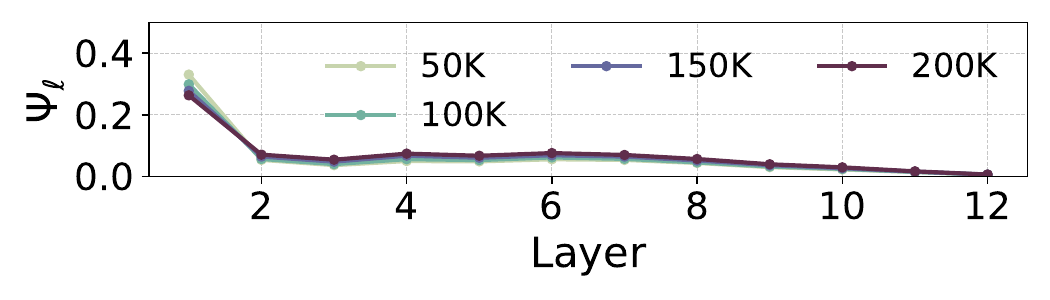}
        \caption{JREG ($\alpha=1.0$)}
        \label{fig:170m_alpha_1.0_displacement}
    \end{subfigure}
    \begin{subfigure}[b]{\linewidth}
        \centering
        \includegraphics[width=\linewidth]{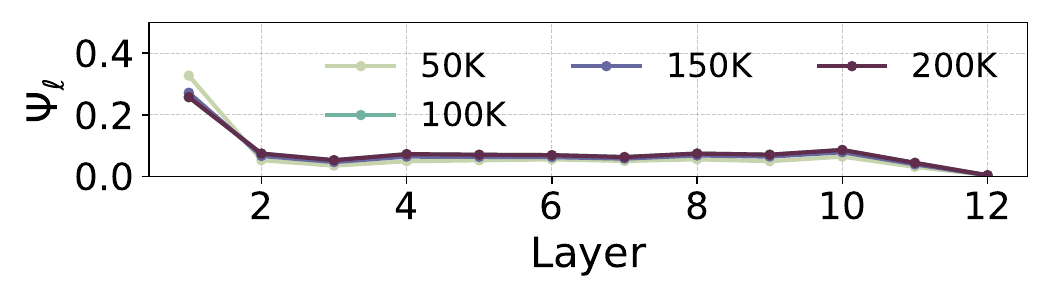}
        \caption{JREG ($\alpha=3.0$)}
        \label{fig:170m_alpha_3.0_displacement}
    \end{subfigure}
    \caption{Checkpoint-wise hidden state displacement on 170M model}
    \label{appendix:170m_ckpt_wise_displacement}
\end{figure}

%% file: latex/figure/appendix_1b_ckpt_wise_displacement.tex
\begin{figure}[t]
    \centering
    \begin{subfigure}[b]{\linewidth}
        \centering
        \includegraphics[width=\linewidth]{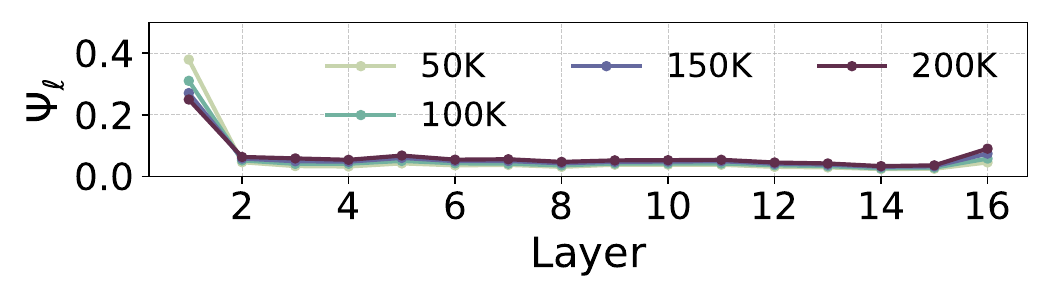}
        \caption{Baseline}
        \label{fig:1b_baseline_displacement}
    \end{subfigure}
    \begin{subfigure}[b]{\linewidth}
        \centering
        \includegraphics[width=\linewidth]{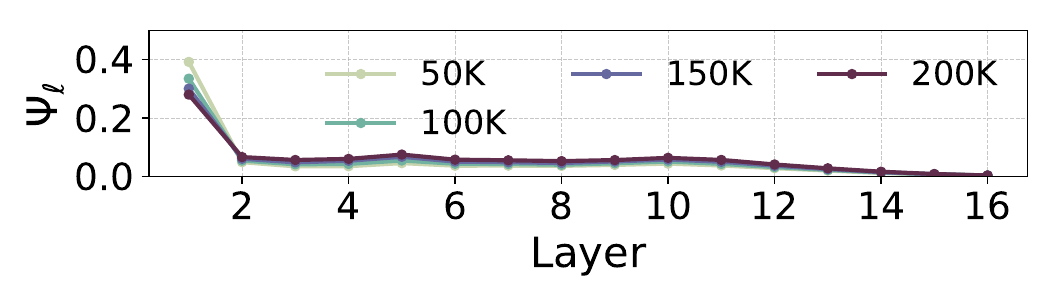}
        \caption{JREG ($\alpha=1.0$)}
        \label{fig:1b_alpha_1.0_displacement}
    \end{subfigure}
    \begin{subfigure}[b]{\linewidth}
        \centering
        \includegraphics[width=\linewidth]{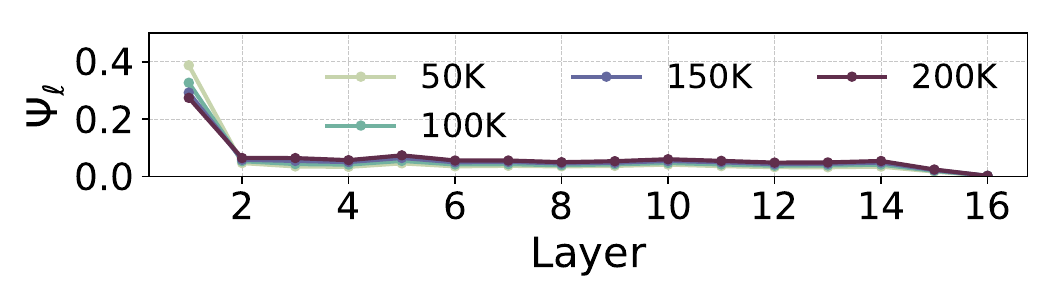}
        \caption{JREG ($\alpha=3.0$)}
        \label{fig:1b_alpha_3.0_displacement}
    \end{subfigure}
    \caption{Checkpoint-wise hidden state displacement on 1B model}
    \label{appendix:1b_ckpt_wise_displacement}
\end{figure}

%% file: latex/figure/appendix_3.4B_ckpt_wise_displacement.tex
\begin{figure}[t]
    \centering
    \begin{subfigure}[b]{\linewidth}
        \centering
        \includegraphics[width=\linewidth]{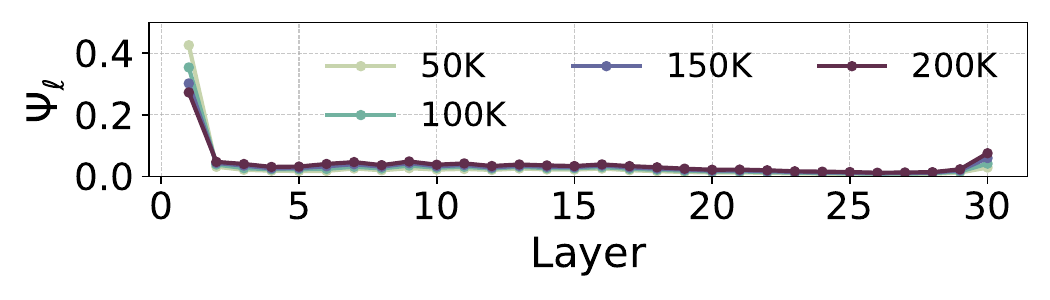}
        \caption{Baseline}
        \label{fig:3.4b_baseline_displacement}
    \end{subfigure}
    \begin{subfigure}[b]{\linewidth}
        \centering
        \includegraphics[width=\linewidth]{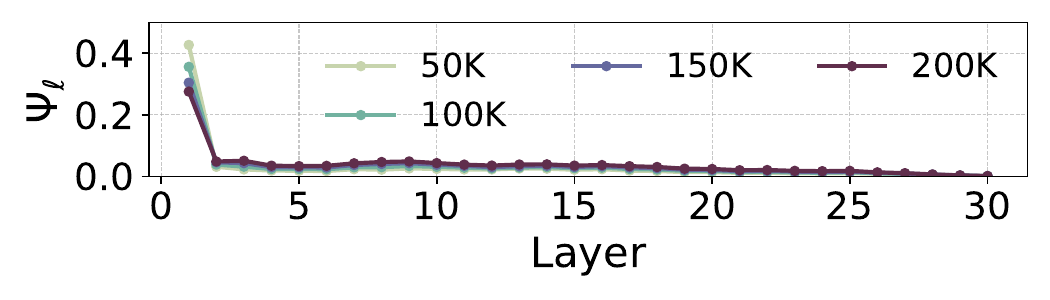}
        \caption{JREG ($\alpha=1.0$)}
        \label{fig:3.4b_alpha_1.0_displacement}
    \end{subfigure}
    \caption{Checkpoint-wise hidden state displacement on 3.4B model}
    \label{appendix:3.4b_ckpt_wise_displacement}
\end{figure}

%% file: latex/figure/appendix_displacement.tex
\begin{table*}[t]
    \centering
    \input{latex/figure/appendix_170m_displacement_table}
    \input{table/appendix_1b_displacement}
    \caption{Displacement $\Psi_{\ell}$ for the baseline and JREG, and the displacement difference $\Delta_{\ell}$ ($\times 10^{-2}$) on the 170M and 1B models.}
    \label{tab:displacement_small_model}
\end{table*}

%% file: latex/figure/appendix_170m_displacement_table.tex
\begin{subtable}[t]{\textwidth}
    \centering
    \small
    \begin{tabular}{crrrrrrrrrrrr}
        \toprule
          Layer index& 1 & 2 & 3 & 4 & 5 & 6 & 7 & 8 & 9 & 10 & 11 & 12\\
        \midrule
         Baseline & 25.63 & 7.02 & 5.72 & 9.48 & 7.52 & 6.31 & 7.39 & 5.53 & 6.00 & 5.59 & 5.98 & 13.22\\
         \midrule
         JREG ($\alpha=0.0$) & 3.02 & 1.89 & 1.41 & 1.58 & 1.54 & 1.73 & 1.45 & 1.93 & 1.92 & 2.4 & 2.84 & 3.68 \\
         $\Delta_{\ell}$ & -22.61 & -5.13 & -4.31 & -7.9 & -5.98 & -4.58 & -5.94 & -3.6 & -4.08 & -3.19 & -3.14 & -9.54\\
         \midrule
         JREG ($\alpha=0.1$) & 4.86 & 2.54 & 1.86 & 2.03 & 1.71 & 1.89 & 1.53 & 1.91 & 1.72 & 2.05 & 2.29 & 2.77 \\
         $\Delta_{\ell}$ & -20.77 & -4.48 & -3.86 & -7.45 & -5.81 & -4.42 & -5.86 & -3.62 & -4.28 & -3.54 & -3.69 & -10.45 \\
         \midrule
         JREG ($\alpha=0.3$) & 13.24 & 5.25 & 3.47 & 3.83 & 2.91 & 2.59 & 2.36 & 1.91 & 1.80 & 1.90 & 1.95 & 1.55 \\
         $\Delta_{\ell}$ & -12.39 & -1.77 & -2.25 & -5.65 & -4.61 & -3.72 & -5.03 & -3.62 & -4.20 & -3.69 & -4.03 & -11.67 \\
         \midrule
         JREG ($\alpha=0.5$) & 20.49 & 6.62 & 5.06 & 6.08 & 4.85 & 4.53 & 3.60 & 2.49 & 2.08 & 2.01 & 1.64 & 1.09 \\
         $\Delta_{\ell}$ & -5.14 & -0.40 & -0.66 & -3.40 & -2.67 & -1.78 & -3.79 & -3.04 & -3.92 & -3.58 & -4.34 & -12.13\\
         \midrule
         JREG ($\alpha=1.0$) & 26.36 & 7.12 & 5.48 & 7.45 & 6.76 & 7.63 & 7.02 & 5.71 & 4.00 & 3.00 & 1.70 & 0.67 \\
         $\Delta_{\ell}$ & 0.73 & \textbf{0.10} & -0.24 & -2.03 & -0.76 & \textbf{1.32} & -0.37 & \textbf{0.18} & -2.00 & -2.59 & -4.28 & -12.55 \\
         \midrule
         JREG ($\alpha=3.0$) & 25.77 & 7.47 & 5.32 & 7.28 & 7.09 & 6.96 & 6.34 & 7.48 & 7.10 & 8.65 & 4.46 & 0.43 \\
         $\Delta_{\ell}$ & \textbf{0.14} & \textbf{0.45} & -0.40 & -2.20 & -0.43 & \textbf{0.65} & -1.05 & \textbf{1.95} & \textbf{1.10} & \textbf{3.06} & -1.52 & -12.79\\
        \bottomrule
    \end{tabular}
    \caption{170M Model}
    \label{tab:170m_displacement}
\end{subtable}

%% file: table/appendix_1b_displacement.tex
\begin{subtable}[t]{\textwidth}
    \centering
    \small
    \tabcolsep 3pt
    \begin{tabular}{crrrrrrrrrrrrrrrr}
        \toprule
         Layer index& 1 & 2 & 3 & 4 & 5 & 6 & 7 & 8 & 9 & 10 & 11 & 12 & 13 & 14 & 15 & 16 \\
        \midrule
         Baseline & 25.00 & 6.37 & 5.90 & 5.44 & 6.82 & 5.49 & 5.63 & 4.77 & 5.26 & 5.33 & 5.45 & 4.57 & 4.26 & 3.40 & 3.64 & 9.06 \\
        \midrule
         JREG ($\alpha=1.0$) & 28.04 & 6.68 & 5.68 & 6.05 & 7.53 & 5.79 & 5.57 & 5.29 & 5.65 & 6.41 & 5.70 & 4.14 & 2.83 & 1.67 & 0.88 & 0.39 \\
         $\Delta_{\ell}$ & \textbf{3.04} & \textbf{0.31} & -0.22 & \textbf{0.61} &  \textbf{0.71} &  \textbf{0.30} & -0.06 & \textbf{0.52} & \textbf{0.39} & \textbf{1.08} & \textbf{0.25} & -0.43 & -1.43 & -1.73 & -2.76 & -8.67 \\
        \midrule
         JREG ($\alpha=3.0$) & 27.42 & 6.46 & 6.44 & 5.72 & 7.39 & 5.60 & 5.60 & 5.01 & 5.36 & 6.04 & 5.47 & 4.86 & 4.93 & 5.43 & 2.46 & 0.26 \\
         $\Delta_{\ell}$ & \textbf{2.42} & \textbf{0.09} & \textbf{0.54} & \textbf{0.28} & \textbf{0.57} & \textbf{0.11} & -0.03 & \textbf{0.24} & \textbf{0.10} & \textbf{0.71} & \textbf{0.02} & \textbf{0.29} & \textbf{0.67} &  \textbf{2.03} & -1.18 & -8.8 \\
        \bottomrule
    \end{tabular}
    \caption{1B Model}
    \label{tab:llama1b_trajectory_diff}
\end{subtable}

%% file: table/acc_llama_pretrain_with_web_organizer.tex
\begin{table*}[t]
    \centering
    \small
    \begin{tabular}{c c|ccccccccc|c}
        \toprule
         Method & $\alpha$ & ARC-e & BoolQ & HellaSwag & LAMBADA & PIQA & RACE & SocialIQA & SciQ & SWAG & \textbf{avg} \\
         \midrule
         \multicolumn{12}{c}{\textbf{170M Param. Model (Llama architecture)}} \\
         Baseline & - & 43.5 & 55.4 & 31.3 & 37.9 & 66.1 & 28.6 & 37.7 & 75.8 & 41.4 & 46.4 \\
         \midrule
         & 0.0 & 44.2 & 56.9 & 31.4 & 35.8 & 65.5 & 28.5 & 37.8 & 74.6 & 41.3 & 46.2 \\
         & 0.1 & 43.7 & 58.2 & 31.2 & 36.2 & 65.6 & 28.9 & 38.0 & 78.0 & 41.4 & 46.8 \\
         & 0.3 & 43.7 & 59.3 & 31.5 & 36.1 & 66.7 & 29.4 & 39.4 & 78.8 & 41.5 & 47.4 \\
         & 0.5 & 44.2 & 56.4 & 31.0 & 35.6 & 64.9 & 28.1 & 37.9 & 77.2 & 41.4 & 46.3 \\
         & 1.0 & 42.7 & 61.0 & 31.4 & 37.6 & 64.6 & 30.0 & 37.9 & 75.7 & 41.8 & 47.0 \\
         \multirow{-6}{*}{\shortstack{JREG (ours)}} & 3.0 & 42.4 & 55.3 & 31.7 & 36.8 & 65.6 & 29.5 & 38.3 & 79.1 & 41.5 & 46.7 \\
         \bottomrule
    \end{tabular}
    \caption{
        Results of downstream task performance pre-trained on Weborganizer.
        $\alpha$ indicates the hyperparameter of JREG.
        Each row shows the score on each benchmark dataset,
        with its average in the rightmost column.
    }
    \label{tab:acc_llama_pretrain_with_web_orgainzier}
\end{table*}

%% file: latex/figure/appendix_170m_web_displacement_table.tex
\begin{table*}[t]
    \centering
    \small
    \begin{tabular}{crrrrrrrrrrrr}
        \toprule
          Layer index& 1 & 2 & 3 & 4 & 5 & 6 & 7 & 8 & 9 & 10 & 11 & 12\\
        \midrule
         Baseline & 27.19 & 6.72 & 6.43 & 6.90 & 6.28 & 8.59 & 5.24 & 5.76 & 6.14 & 4.96 & 5.09 & 12.41\\
         \midrule
         JREG ($\alpha=0.0$) & 2.84 & 1.7 & 1.64 & 1.58 & 1.4 & 1.51 & 2.04 & 1.66 & 1.94 & 1.94 & 2.58 & 3.79\\
         $\Delta_{\ell}$ & -24.35 & -5.02 & -4.79 & -5.32 & -4.88 & -7.08 & -3.20 & -4.10 & -4.20 & -3.02 & -2.51 & -8.62\\
         \midrule
         JREG ($\alpha=0.1$) & 4.86 & 2.54 & 1.86 & 2.03 & 1.71 & 1.89 & 1.53 & 1.91 & 1.72 & 2.05 & 2.29 & 2.77 \\
         $\Delta_{\ell}$ & -22.45 & -4.38 & -4.29 & -4.97 & -4.66 & -7.04 & -3.05 & -4.26 & -4.42 & -3.29 & -3.11 & -9.71 \\
         \midrule
         JREG ($\alpha=0.3$) & 13.79 & 4.67 & 4.35 & 3.60 & 2.93 & 2.28 & 2.66 & 1.78 & 1.74 & 1.47 & 1.46 & 1.56\\
         $\Delta_{\ell}$ & -13.4 & -2.05 & -2.08 & -3.30 & -3.35 & -6.31 & -2.58 & -3.98 & -4.40 & -3.49 & -3.63 & -10.85 \\
         \midrule
         JREG ($\alpha=0.5$) & 21.43 & 6.40 & 6.35 & 6.30 & 5.07 & 3.98 & 4.00 & 2.39 & 2.13 & 1.61 & 1.29 & 1.09 \\
         $\Delta_{\ell}$ & -5.76 & -0.32 & -0.08 & -0.60 & -1.21 & -4.61 & -1.24 & -3.37 & -4.01 & -3.35 & -3.80 & -11.32\\
         \midrule
         JREG ($\alpha=1.0$) & 29.3 & 6.84 & 6.35 & 7.85 & 7.14 & 6.85 & 8.52 & 5.16 & 3.99 & 2.32 & 1.37 & 0.67 \\
         $\Delta_{\ell}$ & \textbf{2.11} & \textbf{0.12} & {-0.08} & \textbf{0.95} & \textbf{0.86} & {-1.74} & \textbf{3.28} & {-0.60} & {-2.15} & {-2.64} & {-3.72} & {-11.74} \\
         \midrule
         JREG ($\alpha=3.0$) & 27.14 & 7.21 & 5.99 & 8.01 & 7.23 & 6.12 & 8.01 & 6.16 & 7.54 & 7.68 & 3.47 & 0.46 \\
         $\Delta_{\ell}$ & -0.05 & \textbf{0.49} & -0.44 & \textbf{1.11} & \textbf{0.95} & -2.47 & \textbf{2.77} & \textbf{0.40} & \textbf{1.40} & \textbf{2.72} & -1.62 & -11.95\\
        \bottomrule
    \end{tabular}
    \caption{Displacement $\Psi_{\ell}$ for the baseline and JREG on 170M model trained on Weborganizer}
    \label{tab:170m_web_displacement}
\end{table*}

%% file: table/appendix_web_jump_rate_table.tex
\begin{table*}[t]
    \centering
    \input{table/jump_rate_table_llama_web_organizer_0}
    \input{table/jump_rate_table_llama_web_organizer_1}
    \input{table/jump_rate_table_llama_web_organizer_2}

    \caption{Checkpoint-wise jump rate of models pre-trained on Weborganizer}
    \label{tab:appendix_jump_rate_table_web}
\end{table*}

%% file: table/jump_rate_table_llama_web_organizer_0.tex
\begin{subtable}[t]{\textwidth}
  \centering
  \small
  \begin{tabular}{cccrrrr}
    \toprule
    \multicolumn{2}{c}{Model} & $\alpha$ & 50k & 100k & 150k & 200k \\
    \midrule
    & Baseline & - & 3.96 & 4.74 & 5.76 & 7.32 \\
    & & 0.0 & 0.49 & 0.53 & 0.9 & 1.21 \\
    & & 0.1 & 0.19 & 0.35 & 0.59 & 0.72 \\
    &  & 0.3 & \textbf{0.00} & 0.03 & 0.11 & 0.10 \\
    &  & 0.5 & \textbf{0.00} & \textbf{0.00} & \textbf{0.00} & \textbf{0.00}\\
    &  & 1.0 & \textbf{0.00} & \textbf{0.00} & \textbf{0.00} & \textbf{0.00} \\
      \multirow{-6}{*}{170M} 
    & \multirow{-6}{*}{JREG} 
    & 3.0 
    & \textbf{0.00} & \textbf{0.00} & \textbf{0.00} & \textbf{0.00} \\
    \bottomrule
  \end{tabular}
  \caption{$\zeta_{L}$}
  \label{tab:jump_rate_table_llama_web_organizer_0}
\end{subtable}

%% file: table/jump_rate_table_llama_web_organizer_1.tex
\begin{subtable}[t]{\textwidth}
  \centering
  \small
  \begin{tabular}{cccrrrr}
    \toprule
    \multicolumn{2}{c}{Model} & $\alpha$ & 50k & 100k & 150k & 200k \\
    \midrule
    & Baseline & - & 4.01 & 4.88 & 5.88 & 7.45 \\
    & & 0.0 & 0.77 & 0.91 & 1.41 & 1.85 \\
    & & 0.1 & 0.31 & 0.50 & 0.81 & 1.03 \\
    &  & 0.3 & \textbf{0.00} & 0.03 & 0.11 & 0.10 \\
    &  & 0.5 & \textbf{0.00} & \textbf{0.00} & \textbf{0.00} & \textbf{0.00} \\
    &  & 1.0 & \textbf{0.00} & \textbf{0.00} & \textbf{0.00} & \textbf{0.00} \\
      \multirow{-6}{*}{170M} 
    & \multirow{-6}{*}{JREG} 
    & 3.0 
    & \textbf{0.00} & \textbf{0.00} & \textbf{0.00} & \textbf{0.00} \\
    \bottomrule
  \end{tabular}
  \caption{$\zeta_{L-1}$}
  \label{tab:jump_rate_table_llama_web_organizer_1}
\end{subtable}

%% file: table/jump_rate_table_llama_web_organizer_2.tex
\begin{subtable}[t]{\textwidth}
  \centering
  \small
  \begin{tabular}{cccrrrr}
    \toprule
    \multicolumn{2}{c}{Model} & $\alpha$ & 50k & 100k & 150k & 200k \\
    \midrule
    & Baseline & - & 4.01 & 4.88 & 5.88 & 7.45 \\
    & & 0.0 & 0.77 & 0.91 & 1.41 & 1.85 \\
    & & 0.1 & 0.31 & 0.50 & 0.81 & 1.03 \\
    &  & 0.3 & \textbf{0.00} & 0.03 & 0.11 & 0.10 \\
    &  & 0.5 & \textbf{0.00} & \textbf{0.00} & \textbf{0.00} & \textbf{0.00} \\
    &  & 1.0 & \textbf{0.00} & \textbf{0.00} & \textbf{0.00} & \textbf{0.00} \\
      \multirow{-6}{*}{170M} 
    & \multirow{-6}{*}{JREG} 
    & 3.0 
    & \textbf{0.00} & 0.25 & 0.13 & 0.14 \\
    \bottomrule
  \end{tabular}
  \caption{$\zeta_{L-2}$}
  \label{tab:jump_rate_table_llama_web_organizer_2}
\end{subtable}

%% file: latex/figure/appendix_web_170m_ckpt_wise_displacement.tex
\begin{figure}[t]
    \centering
    \begin{subfigure}[b]{\linewidth}
        \centering
        \includegraphics[width=\linewidth]{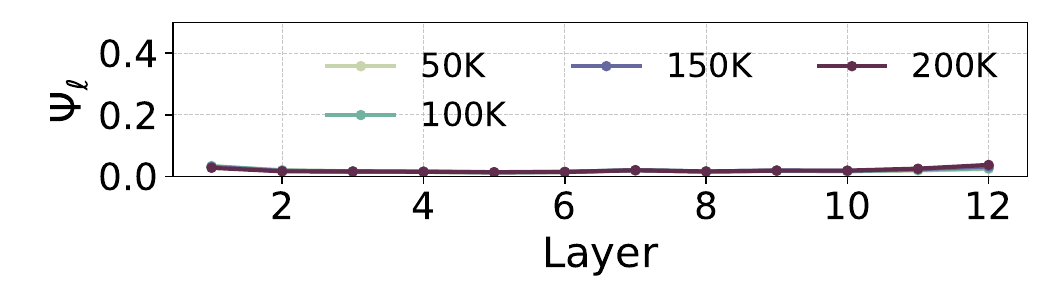}
        \caption{JREG ($\alpha=0.0$)}
        \label{fig:170m_web_alpha_0.0_displacement}
    \end{subfigure}
    \begin{subfigure}[b]{\linewidth}
        \centering
        \includegraphics[width=\linewidth]{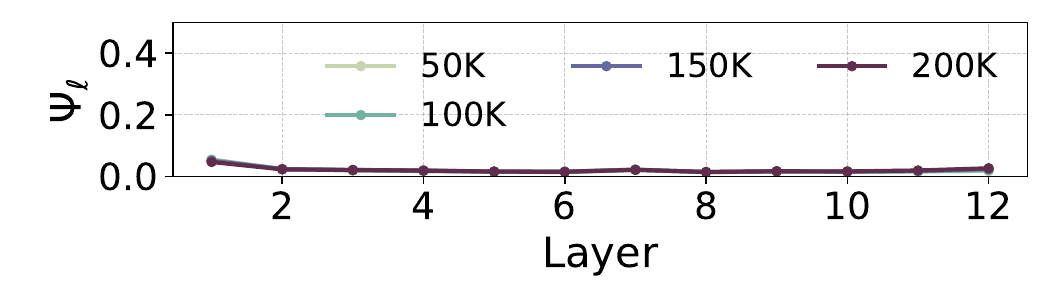}
        \caption{JREG ($\alpha=0.1$)}
        \label{fig:170m_web_alpha_0.1_displacement}
    \end{subfigure}
    \begin{subfigure}[b]{\linewidth}
        \centering
        \includegraphics[width=\linewidth]{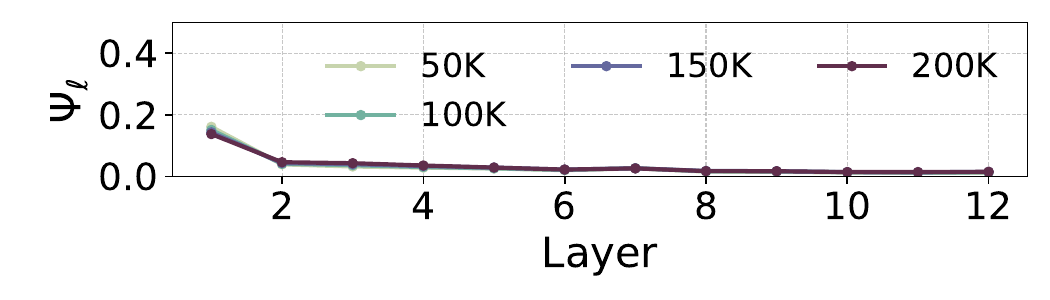}
        \caption{JREG ($\alpha=0.3$)}
        \label{fig:170m_web_alpha_0.3_displacement}
    \end{subfigure}
    \begin{subfigure}[b]{\linewidth}
        \centering
        \includegraphics[width=\linewidth]{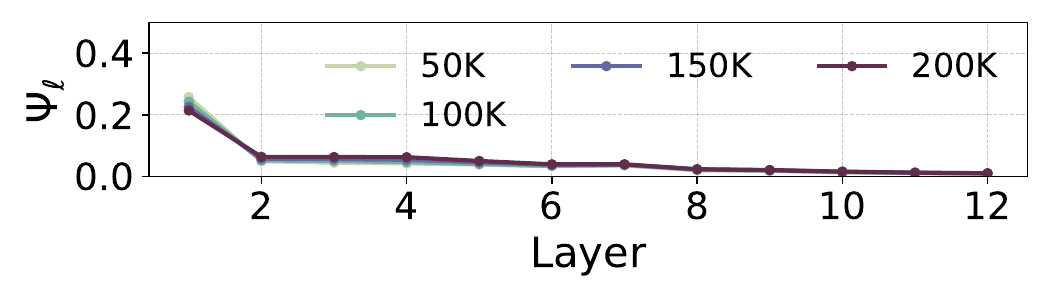}
        \caption{JREG ($\alpha=0.5$)}
        \label{fig:170m_web_alpha_0.5_displacement}
    \end{subfigure}
    \begin{subfigure}[b]{\linewidth}
        \centering
        \includegraphics[width=\linewidth]{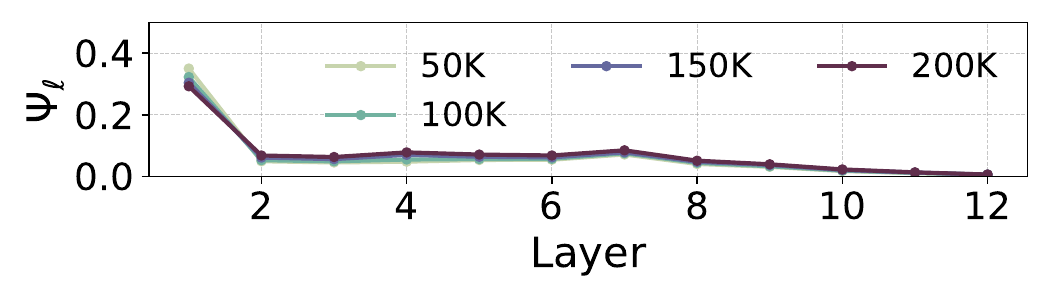}
        \caption{JREG ($\alpha=1.0$)}
        \label{fig:170m_web_alpha_1.0_displacement}
    \end{subfigure}
    \begin{subfigure}[b]{\linewidth}
        \centering
        \includegraphics[width=\linewidth]{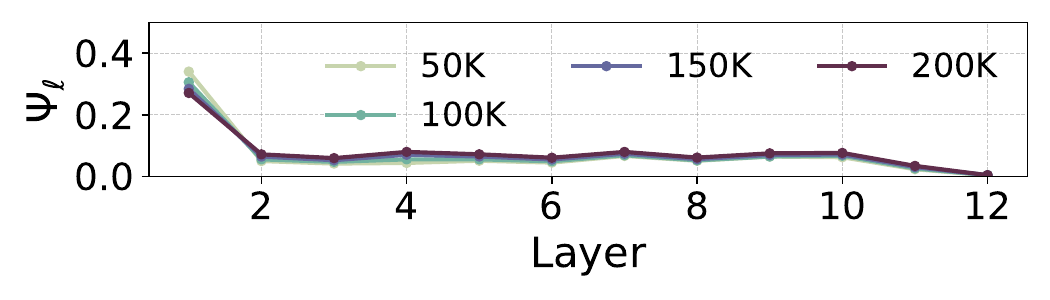}
        \caption{JREG ($\alpha=3.0$)}
        \label{fig:170m_web_alpha_3.0_displacement}
    \end{subfigure}
    \caption{170M checkpoint-wise hidden state displacement pretrained on Weborganizer.}
    \label{appendix:170m_web_ckpt_wise_displacement}
\end{figure}

%% file: custom.bib
@inproceedings{DBLP:conf/icpr/Teerapittayanon16,
  author       = {Surat Teerapittayanon and
                  Bradley McDanel and
                  H. T. Kung},
  title        = {BranchyNet: Fast inference via early exiting from deep neural networks},
  booktitle    = {23rd International Conference on Pattern Recognition, {ICPR} 2016,
                  Canc{\'{u}}n, Mexico, December 4-8, 2016},
  pages        = {2464--2469},
  publisher    = {{IEEE}},
  year         = {2016},
  url          = {https://doi.org/10.1109/ICPR.2016.7900006},
  doi          = {10.1109/ICPR.2016.7900006},
  bibsource    = {dblp computer science bibliography, https://dblp.org}
}

@inproceedings{DBLP:conf/acl/ElhoushiSLHWL0A24,
  author       = {Mostafa Elhoushi and
                  Akshat Shrivastava and
                  Diana Liskovich and
                  Basil Hosmer and
                  Bram Wasti and
                  Liangzhen Lai and
                  Anas Mahmoud and
                  Bilge Acun and
                  Saurabh Agarwal and
                  Ahmed Roman and
                  Ahmed A Aly and
                  Beidi Chen and
                  Carole{-}Jean Wu},
  editor       = {Lun{-}Wei Ku and
                  Andre Martins and
                  Vivek Srikumar},
  title        = {LayerSkip: Enabling Early Exit Inference and Self-Speculative Decoding},
  booktitle    = {Proceedings of the 62nd Annual Meeting of the Association for Computational
                  Linguistics (Volume 1: Long Papers), {ACL} 2024, Bangkok, Thailand,
                  August 11-16, 2024},
  pages        = {12622--12642},
  publisher    = {Association for Computational Linguistics},
  year         = {2024},
  url          = {https://doi.org/10.18653/v1/2024.acl-long.681},
  doi          = {10.18653/V1/2024.ACL-LONG.681},
  bibsource    = {dblp computer science bibliography, https://dblp.org}
}

@inproceedings{DBLP:conf/iclr/MerityX0S17,
  author       = {Stephen Merity and
                  Caiming Xiong and
                  James Bradbury and
                  Richard Socher},
  title        = {Pointer Sentinel Mixture Models},
  booktitle    = {5th International Conference on Learning Representations, {ICLR} 2017,
                  Toulon, France, April 24-26, 2017, Conference Track Proceedings},
  publisher    = {OpenReview.net},
  year         = {2017},
  url          = {https://openreview.net/forum?id=Byj72udxe},
  bibsource    = {dblp computer science bibliography, https://dblp.org}
}

@inproceedings{DBLP:conf/acl/PapernoKLPBPBBF16,
  author       = {Denis Paperno and
                  Germ{\'{a}}n Kruszewski and
                  Angeliki Lazaridou and
                  Quan Ngoc Pham and
                  Raffaella Bernardi and
                  Sandro Pezzelle and
                  Marco Baroni and
                  Gemma Boleda and
                  Raquel Fern{\'{a}}ndez},
  title        = {The {LAMBADA} dataset: Word prediction requiring a broad discourse
                  context},
  booktitle    = {Proceedings of the 54th Annual Meeting of the Association for Computational
                  Linguistics, {ACL} 2016, August 7-12, 2016, Berlin, Germany, Volume
                  1: Long Papers},
  publisher    = {The Association for Computer Linguistics},
  year         = {2016},
  url          = {https://doi.org/10.18653/v1/p16-1144},
  doi          = {10.18653/V1/P16-1144},
  bibsource    = {dblp computer science bibliography, https://dblp.org}
}

@inproceedings{DBLP:conf/naacl/ClarkLCK0T19,
  author       = {Christopher Clark and
                  Kenton Lee and
                  Ming{-}Wei Chang and
                  Tom Kwiatkowski and
                  Michael Collins and
                  Kristina Toutanova},
  editor       = {Jill Burstein and
                  Christy Doran and
                  Thamar Solorio},
  title        = {BoolQ: Exploring the Surprising Difficulty of Natural Yes/No Questions},
  booktitle    = {Proceedings of the 2019 Conference of the North American Chapter of
                  the Association for Computational Linguistics: Human Language Technologies,
                  {NAACL-HLT} 2019, Minneapolis, MN, USA, June 2-7, 2019, Volume 1 (Long
                  and Short Papers)},
  pages        = {2924--2936},
  publisher    = {Association for Computational Linguistics},
  year         = {2019},
  url          = {https://doi.org/10.18653/v1/n19-1300},
  doi          = {10.18653/V1/N19-1300},
  bibsource    = {dblp computer science bibliography, https://dblp.org}
}

@misc{DBLP:journals/corr/abs-1803-05457,
  title={Think you have Solved Question Answering? Try ARC, the AI2 Reasoning Challenge}, 
      author={Peter Clark and Isaac Cowhey and Oren Etzioni and Tushar Khot and Ashish Sabharwal and Carissa Schoenick and Oyvind Tafjord},
      year={2018},
      eprint={1803.05457},
      archivePrefix={arXiv},
      primaryClass={cs.AI},
      url={https://arxiv.org/abs/1803.05457}, 
}

@inproceedings{DBLP:conf/emnlp/LaiXLYH17,
  author       = {Guokun Lai and
                  Qizhe Xie and
                  Hanxiao Liu and
                  Yiming Yang and
                  Eduard H. Hovy},
  editor       = {Martha Palmer and
                  Rebecca Hwa and
                  Sebastian Riedel},
  title        = {{RACE:} Large-scale ReAding Comprehension Dataset From Examinations},
  booktitle    = {Proceedings of the 2017 Conference on Empirical Methods in Natural
                  Language Processing, {EMNLP} 2017, Copenhagen, Denmark, September
                  9-11, 2017},
  pages        = {785--794},
  publisher    = {Association for Computational Linguistics},
  year         = {2017},
  url          = {https://doi.org/10.18653/v1/d17-1082},
  doi          = {10.18653/V1/D17-1082},
  bibsource    = {dblp computer science bibliography, https://dblp.org}
}

@inproceedings{DBLP:conf/emnlp/SapRCBC19,
  author       = {Maarten Sap and
                  Hannah Rashkin and
                  Derek Chen and
                  Ronan Le Bras and
                  Yejin Choi},
  editor       = {Kentaro Inui and
                  Jing Jiang and
                  Vincent Ng and
                  Xiaojun Wan},
  title        = {Social IQa: Commonsense Reasoning about Social Interactions},
  booktitle    = {Proceedings of the 2019 Conference on Empirical Methods in Natural
                  Language Processing and the 9th International Joint Conference on
                  Natural Language Processing, {EMNLP-IJCNLP} 2019, Hong Kong, China,
                  November 3-7, 2019},
  pages        = {4462--4472},
  publisher    = {Association for Computational Linguistics},
  year         = {2019},
  url          = {https://doi.org/10.18653/v1/D19-1454},
  doi          = {10.18653/V1/D19-1454},
  bibsource    = {dblp computer science bibliography, https://dblp.org}
}

@misc{DBLP:journals/corr/abs-2406-19384,
  title={The Remarkable Robustness of LLMs: Stages of Inference?}, 
      author={Vedang Lad and Jin Hwa Lee and Wes Gurnee and Max Tegmark},
      year={2025},
      eprint={2406.19384},
      archivePrefix={arXiv},
      primaryClass={cs.LG},
      url={https://arxiv.org/abs/2406.19384}, 
}

@inproceedings{DBLP:journals/corr/abs-2403-03853,
  title = "{S}hort{GPT}: Layers in Large Language Models are More Redundant Than You Expect",
    author = "Men, Xin  and
      Xu, Mingyu  and
      Zhang, Qingyu  and
      Yuan, Qianhao  and
      Wang, Bingning  and
      Lin, Hongyu  and
      Lu, Yaojie  and
      Han, Xianpei  and
      Chen, Weipeng",
    editor = "Che, Wanxiang  and
      Nabende, Joyce  and
      Shutova, Ekaterina  and
      Pilehvar, Mohammad Taher",
    booktitle = "Findings of the Association for Computational Linguistics: ACL 2025",
    month = jul,
    year = "2025",
    address = "Vienna, Austria",
    publisher = "Association for Computational Linguistics",
    url = "https://aclanthology.org/2025.findings-acl.1035/",
    pages = "20192--20204",
    ISBN = "979-8-89176-256-5",
}

@inproceedings{DBLP:conf/aaai/SunPNJ25,
  author       = {Qi Sun and
                  Marc Pickett and
                  Aakash Kumar Nain and
                  Llion Jones},
  editor       = {Toby Walsh and
                  Julie Shah and
                  Zico Kolter},
  title        = {Transformer Layers as Painters},
  booktitle    = {AAAI-25, Sponsored by the Association for the Advancement of Artificial
                  Intelligence, February 25 - March 4, 2025, Philadelphia, PA, {USA}},
  pages        = {25219--25227},
  publisher    = {{AAAI} Press},
  year         = {2025},
  url          = {https://doi.org/10.1609/aaai.v39i24.34708},
  doi          = {10.1609/AAAI.V39I24.34708},
  bibsource    = {dblp computer science bibliography, https://dblp.org}
}

@inproceedings{DBLP:journals/corr/abs-2412-13795,
  title={Mix-{LN}: Unleashing the Power of Deeper Layers by Combining Pre-{LN} and Post-{LN}},
author={Pengxiang Li and Lu Yin and Shiwei Liu},
booktitle={The Thirteenth International Conference on Learning Representations},
year={2025},
url={https://openreview.net/forum?id=BChpQU64RG}
}

@misc{DBLP:journals/corr/abs-2502-05795,
  title={The Curse of Depth in Large Language Models}, 
      author={Wenfang Sun and Xinyuan Song and Pengxiang Li and Lu Yin and Yefeng Zheng and Shiwei Liu},
      year={2025},
      eprint={2502.05795},
      archivePrefix={arXiv},
      primaryClass={cs.LG},
      url={https://arxiv.org/abs/2502.05795}, 
}

@inproceedings{DBLP:conf/nips/ZhangH20,
  author       = {Minjia Zhang and
                  Yuxiong He},
  editor       = {Hugo Larochelle and
                  Marc'Aurelio Ranzato and
                  Raia Hadsell and
                  Maria{-}Florina Balcan and
                  Hsuan{-}Tien Lin},
  title        = {Accelerating Training of Transformer-Based Language Models with Progressive
                  Layer Dropping},
  booktitle    = {Advances in Neural Information Processing Systems 33: Annual Conference
                  on Neural Information Processing Systems 2020, NeurIPS 2020, December
                  6-12, 2020, virtual},
  year         = {2020},
  url          = {https://proceedings.neurips.cc/paper/2020/hash/a1140a3d0df1c81e24ae954d935e8926-Abstract.html},
  bibsource    = {dblp computer science bibliography, https://dblp.org}
}

@inproceedings{DBLP:conf/iclr/FanGJ20,
  author       = {Angela Fan and
                  Edouard Grave and
                  Armand Joulin},
  title        = {Reducing Transformer Depth on Demand with Structured Dropout},
  booktitle    = {8th International Conference on Learning Representations, {ICLR} 2020,
                  Addis Ababa, Ethiopia, April 26-30, 2020},
  publisher    = {OpenReview.net},
  year         = {2020},
  url          = {https://openreview.net/forum?id=SylO2yStDr},
  bibsource    = {dblp computer science bibliography, https://dblp.org}
}

@inproceedings{DBLP:conf/icml/BidermanSABOHKP23,
  author       = {Stella Biderman and
                  Hailey Schoelkopf and
                  Quentin Gregory Anthony and
                  Herbie Bradley and
                  Kyle O'Brien and
                  Eric Hallahan and
                  Mohammad Aflah Khan and
                  Shivanshu Purohit and
                  USVSN Sai Prashanth and
                  Edward Raff and
                  Aviya Skowron and
                  Lintang Sutawika and
                  Oskar van der Wal},
  editor       = {Andreas Krause and
                  Emma Brunskill and
                  Kyunghyun Cho and
                  Barbara Engelhardt and
                  Sivan Sabato and
                  Jonathan Scarlett},
  title        = {Pythia: {A} Suite for Analyzing Large Language Models Across Training
                  and Scaling},
  booktitle    = {International Conference on Machine Learning, {ICML} 2023, 23-29 July
                  2023, Honolulu, Hawaii, {USA}},
  series       = {Proceedings of Machine Learning Research},
  volume       = {202},
  pages        = {2397--2430},
  publisher    = {{PMLR}},
  year         = {2023},
  url          = {https://proceedings.mlr.press/v202/biderman23a.html},
  bibsource    = {dblp computer science bibliography, https://dblp.org}
}

@misc{DBLP:journals/corr/abs-2407-21783,
  title={The Llama 3 Herd of Models}, 
      author={Aaron Grattafiori and Abhimanyu Dubey and Abhinav Jauhri and Abhinav Pandey and Abhishek Kadian and Ahmad Al-Dahle and Aiesha Letman and Akhil Mathur and Alan Schelten and Alex Vaughan and Amy Yang and Angela Fan and Anirudh Goyal and Anthony Hartshorn and Aobo Yang and Archi Mitra and Archie Sravankumar and Artem Korenev and Arthur Hinsvark and Arun Rao and Aston Zhang and Aurelien Rodriguez and Austen Gregerson and Ava Spataru and Baptiste Roziere and Bethany Biron and Binh Tang and Bobbie Chern and Charlotte Caucheteux and Chaya Nayak and Chloe Bi and Chris Marra and Chris McConnell and Christian Keller and Christophe Touret and Chunyang Wu and Corinne Wong and Cristian Canton Ferrer and Cyrus Nikolaidis and Damien Allonsius and Daniel Song and Danielle Pintz and Danny Livshits and Danny Wyatt and David Esiobu and Dhruv Choudhary and Dhruv Mahajan and Diego Garcia-Olano and Diego Perino and Dieuwke Hupkes and Egor Lakomkin and Ehab AlBadawy and Elina Lobanova and Emily Dinan and Eric Michael Smith and Filip Radenovic and Francisco Guzmán and Frank Zhang and Gabriel Synnaeve and Gabrielle Lee and Georgia Lewis Anderson and Govind Thattai and Graeme Nail and Gregoire Mialon and Guan Pang and Guillem Cucurell and Hailey Nguyen and Hannah Korevaar and Hu Xu and Hugo Touvron and Iliyan Zarov and Imanol Arrieta Ibarra and Isabel Kloumann and Ishan Misra and Ivan Evtimov and Jack Zhang and Jade Copet and Jaewon Lee and Jan Geffert and Jana Vranes and Jason Park and Jay Mahadeokar and Jeet Shah and Jelmer van der Linde and Jennifer Billock and Jenny Hong and Jenya Lee and Jeremy Fu and Jianfeng Chi and Jianyu Huang and Jiawen Liu and Jie Wang and Jiecao Yu and Joanna Bitton and Joe Spisak and Jongsoo Park and Joseph Rocca and Joshua Johnstun and Joshua Saxe and Junteng Jia and Kalyan Vasuden Alwala and Karthik Prasad and Kartikeya Upasani and Kate Plawiak and Ke Li and Kenneth Heafield and Kevin Stone and Khalid El-Arini and Krithika Iyer and Kshitiz Malik and Kuenley Chiu and Kunal Bhalla and Kushal Lakhotia and Lauren Rantala-Yeary and Laurens van der Maaten and Lawrence Chen and Liang Tan and Liz Jenkins and Louis Martin and Lovish Madaan and Lubo Malo and Lukas Blecher and Lukas Landzaat and Luke de Oliveira and Madeline Muzzi and Mahesh Pasupuleti and Mannat Singh and Manohar Paluri and Marcin Kardas and Maria Tsimpoukelli and Mathew Oldham and Mathieu Rita and Maya Pavlova and Melanie Kambadur and Mike Lewis and Min Si and Mitesh Kumar Singh and Mona Hassan and Naman Goyal and Narjes Torabi and Nikolay Bashlykov and Nikolay Bogoychev and Niladri Chatterji and Ning Zhang and Olivier Duchenne and Onur Çelebi and Patrick Alrassy and Pengchuan Zhang and Pengwei Li and Petar Vasic and Peter Weng and Prajjwal Bhargava and Pratik Dubal and Praveen Krishnan and Punit Singh Koura and Puxin Xu and Qing He and Qingxiao Dong and Ragavan Srinivasan and Raj Ganapathy and Ramon Calderer and Ricardo Silveira Cabral and Robert Stojnic and Roberta Raileanu and Rohan Maheswari and Rohit Girdhar and Rohit Patel and Romain Sauvestre and Ronnie Polidoro and Roshan Sumbaly and Ross Taylor and Ruan Silva and Rui Hou and Rui Wang and Saghar Hosseini and Sahana Chennabasappa and Sanjay Singh and Sean Bell and Seohyun Sonia Kim and Sergey Edunov and Shaoliang Nie and Sharan Narang and Sharath Raparthy and Sheng Shen and Shengye Wan and Shruti Bhosale and Shun Zhang and Simon Vandenhende and Soumya Batra and Spencer Whitman and Sten Sootla and Stephane Collot and Suchin Gururangan and Sydney Borodinsky and Tamar Herman and Tara Fowler and Tarek Sheasha and Thomas Georgiou and Thomas Scialom and Tobias Speckbacher and Todor Mihaylov and Tong Xiao and Ujjwal Karn and Vedanuj Goswami and Vibhor Gupta and Vignesh Ramanathan and Viktor Kerkez and Vincent Gonguet and Virginie Do and Vish Vogeti and Vítor Albiero and Vladan Petrovic and Weiwei Chu and Wenhan Xiong and Wenyin Fu and Whitney Meers and Xavier Martinet and Xiaodong Wang and Xiaofang Wang and Xiaoqing Ellen Tan and Xide Xia and Xinfeng Xie and Xuchao Jia and Xuewei Wang and Yaelle Goldschlag and Yashesh Gaur and Yasmine Babaei and Yi Wen and Yiwen Song and Yuchen Zhang and Yue Li and Yuning Mao and Zacharie Delpierre Coudert and Zheng Yan and Zhengxing Chen and Zoe Papakipos and Aaditya Singh and Aayushi Srivastava and Abha Jain and Adam Kelsey and Adam Shajnfeld and Adithya Gangidi and Adolfo Victoria and Ahuva Goldstand and Ajay Menon and Ajay Sharma and Alex Boesenberg and Alexei Baevski and Allie Feinstein and Amanda Kallet and Amit Sangani and Amos Teo and Anam Yunus and Andrei Lupu and Andres Alvarado and Andrew Caples and Andrew Gu and Andrew Ho and Andrew Poulton and Andrew Ryan and Ankit Ramchandani and Annie Dong and Annie Franco and Anuj Goyal and Aparajita Saraf and Arkabandhu Chowdhury and Ashley Gabriel and Ashwin Bharambe and Assaf Eisenman and Azadeh Yazdan and Beau James and Ben Maurer and Benjamin Leonhardi and Bernie Huang and Beth Loyd and Beto De Paola and Bhargavi Paranjape and Bing Liu and Bo Wu and Boyu Ni and Braden Hancock and Bram Wasti and Brandon Spence and Brani Stojkovic and Brian Gamido and Britt Montalvo and Carl Parker and Carly Burton and Catalina Mejia and Ce Liu and Changhan Wang and Changkyu Kim and Chao Zhou and Chester Hu and Ching-Hsiang Chu and Chris Cai and Chris Tindal and Christoph Feichtenhofer and Cynthia Gao and Damon Civin and Dana Beaty and Daniel Kreymer and Daniel Li and David Adkins and David Xu and Davide Testuggine and Delia David and Devi Parikh and Diana Liskovich and Didem Foss and Dingkang Wang and Duc Le and Dustin Holland and Edward Dowling and Eissa Jamil and Elaine Montgomery and Eleonora Presani and Emily Hahn and Emily Wood and Eric-Tuan Le and Erik Brinkman and Esteban Arcaute and Evan Dunbar and Evan Smothers and Fei Sun and Felix Kreuk and Feng Tian and Filippos Kokkinos and Firat Ozgenel and Francesco Caggioni and Frank Kanayet and Frank Seide and Gabriela Medina Florez and Gabriella Schwarz and Gada Badeer and Georgia Swee and Gil Halpern and Grant Herman and Grigory Sizov and Guangyi and Zhang and Guna Lakshminarayanan and Hakan Inan and Hamid Shojanazeri and Han Zou and Hannah Wang and Hanwen Zha and Haroun Habeeb and Harrison Rudolph and Helen Suk and Henry Aspegren and Hunter Goldman and Hongyuan Zhan and Ibrahim Damlaj and Igor Molybog and Igor Tufanov and Ilias Leontiadis and Irina-Elena Veliche and Itai Gat and Jake Weissman and James Geboski and James Kohli and Janice Lam and Japhet Asher and Jean-Baptiste Gaya and Jeff Marcus and Jeff Tang and Jennifer Chan and Jenny Zhen and Jeremy Reizenstein and Jeremy Teboul and Jessica Zhong and Jian Jin and Jingyi Yang and Joe Cummings and Jon Carvill and Jon Shepard and Jonathan McPhie and Jonathan Torres and Josh Ginsburg and Junjie Wang and Kai Wu and Kam Hou U and Karan Saxena and Kartikay Khandelwal and Katayoun Zand and Kathy Matosich and Kaushik Veeraraghavan and Kelly Michelena and Keqian Li and Kiran Jagadeesh and Kun Huang and Kunal Chawla and Kyle Huang and Lailin Chen and Lakshya Garg and Lavender A and Leandro Silva and Lee Bell and Lei Zhang and Liangpeng Guo and Licheng Yu and Liron Moshkovich and Luca Wehrstedt and Madian Khabsa and Manav Avalani and Manish Bhatt and Martynas Mankus and Matan Hasson and Matthew Lennie and Matthias Reso and Maxim Groshev and Maxim Naumov and Maya Lathi and Meghan Keneally and Miao Liu and Michael L. Seltzer and Michal Valko and Michelle Restrepo and Mihir Patel and Mik Vyatskov and Mikayel Samvelyan and Mike Clark and Mike Macey and Mike Wang and Miquel Jubert Hermoso and Mo Metanat and Mohammad Rastegari and Munish Bansal and Nandhini Santhanam and Natascha Parks and Natasha White and Navyata Bawa and Nayan Singhal and Nick Egebo and Nicolas Usunier and Nikhil Mehta and Nikolay Pavlovich Laptev and Ning Dong and Norman Cheng and Oleg Chernoguz and Olivia Hart and Omkar Salpekar and Ozlem Kalinli and Parkin Kent and Parth Parekh and Paul Saab and Pavan Balaji and Pedro Rittner and Philip Bontrager and Pierre Roux and Piotr Dollar and Polina Zvyagina and Prashant Ratanchandani and Pritish Yuvraj and Qian Liang and Rachad Alao and Rachel Rodriguez and Rafi Ayub and Raghotham Murthy and Raghu Nayani and Rahul Mitra and Rangaprabhu Parthasarathy and Raymond Li and Rebekkah Hogan and Robin Battey and Rocky Wang and Russ Howes and Ruty Rinott and Sachin Mehta and Sachin Siby and Sai Jayesh Bondu and Samyak Datta and Sara Chugh and Sara Hunt and Sargun Dhillon and Sasha Sidorov and Satadru Pan and Saurabh Mahajan and Saurabh Verma and Seiji Yamamoto and Sharadh Ramaswamy and Shaun Lindsay and Shaun Lindsay and Sheng Feng and Shenghao Lin and Shengxin Cindy Zha and Shishir Patil and Shiva Shankar and Shuqiang Zhang and Shuqiang Zhang and Sinong Wang and Sneha Agarwal and Soji Sajuyigbe and Soumith Chintala and Stephanie Max and Stephen Chen and Steve Kehoe and Steve Satterfield and Sudarshan Govindaprasad and Sumit Gupta and Summer Deng and Sungmin Cho and Sunny Virk and Suraj Subramanian and Sy Choudhury and Sydney Goldman and Tal Remez and Tamar Glaser and Tamara Best and Thilo Koehler and Thomas Robinson and Tianhe Li and Tianjun Zhang and Tim Matthews and Timothy Chou and Tzook Shaked and Varun Vontimitta and Victoria Ajayi and Victoria Montanez and Vijai Mohan and Vinay Satish Kumar and Vishal Mangla and Vlad Ionescu and Vlad Poenaru and Vlad Tiberiu Mihailescu and Vladimir Ivanov and Wei Li and Wenchen Wang and Wenwen Jiang and Wes Bouaziz and Will Constable and Xiaocheng Tang and Xiaojian Wu and Xiaolan Wang and Xilun Wu and Xinbo Gao and Yaniv Kleinman and Yanjun Chen and Ye Hu and Ye Jia and Ye Qi and Yenda Li and Yilin Zhang and Ying Zhang and Yossi Adi and Youngjin Nam and Yu and Wang and Yu Zhao and Yuchen Hao and Yundi Qian and Yunlu Li and Yuzi He and Zach Rait and Zachary DeVito and Zef Rosnbrick and Zhaoduo Wen and Zhenyu Yang and Zhiwei Zhao and Zhiyu Ma},
      year={2024},
      eprint={2407.21783},
      archivePrefix={arXiv},
      primaryClass={cs.AI},
      url={https://arxiv.org/abs/2407.21783}, 
}

@misc{DBLP:journals/corr/abs-2407-15516,
   title={Attention Is All You Need But You Don't Need All Of It For Inference of Large Language Models}, 
      author={Georgy Tyukin and Gbetondji J-S Dovonon and Jean Kaddour and Pasquale Minervini},
      year={2024},
      eprint={2407.15516},
      archivePrefix={arXiv},
      primaryClass={cs.LG},
      url={https://arxiv.org/abs/2407.15516}, 
}

@inproceedings{DBLP:conf/nips/PenedoKALMRW024,
  author       = {Guilherme Penedo and
                  Hynek Kydl{\'{\i}}cek and
                  Loubna Ben Allal and
                  Anton Lozhkov and
                  Margaret Mitchell and
                  Colin A. Raffel and
                  Leandro von Werra and
                  Thomas Wolf},
  editor       = {Amir Globersons and
                  Lester Mackey and
                  Danielle Belgrave and
                  Angela Fan and
                  Ulrich Paquet and
                  Jakub M. Tomczak and
                  Cheng Zhang},
  title        = {The FineWeb Datasets: Decanting the Web for the Finest Text Data at
                  Scale},
  booktitle    = {Advances in Neural Information Processing Systems 38: Annual Conference
                  on Neural Information Processing Systems 2024, NeurIPS 2024, Vancouver,
                  BC, Canada, December 10 - 15, 2024},
  year         = {2024},
  url          = {http://papers.nips.cc/paper\_files/paper/2024/hash/370df50ccfdf8bde18f8f9c2d9151bda-Abstract-Datasets\_and\_Benchmarks\_Track.html},
  bibsource    = {dblp computer science bibliography, https://dblp.org}
}

@misc{meta_lingua,
  author = {Mathurin Videau and Badr Youbi Idrissi and Daniel Haziza and Luca Wehrstedt and Jade Copet and Olivier Teytaud and David Lopez-Paz},
  title = {{Meta Lingua}: A minimal {PyTorch LLM} training library},
  url = {https://github.com/facebookresearch/lingua},
  year = {2024}
}

@inproceedings{DBLP:conf/acl/SimoulinC21,
  author       = {Antoine Simoulin and
                  Beno{\^{\i}}t Crabb{\'{e}}},
  editor       = {Jad Kabbara and
                  Haitao Lin and
                  Amandalynne Paullada and
                  Jannis Vamvas},
  title        = {How Many Layers and Why? An Analysis of the Model Depth in Transformers},
  booktitle    = {Proceedings of the {ACL-IJCNLP} 2021 Student Research Workshop, {ACL}
                  2021, Online, JUli 5-10, 2021},
  pages        = {221--228},
  publisher    = {Association for Computational Linguistics},
  year         = {2021},
  url          = {https://doi.org/10.18653/v1/2021.acl-srw.23},
  doi          = {10.18653/V1/2021.ACL-SRW.23},
  bibsource    = {dblp computer science bibliography, https://dblp.org}
}

@inproceedings{DBLP:conf/eacl/GodeyCS24,
  author       = {Nathan Godey and
                  {\'{E}}ric Villemonte de la Clergerie and
                  Beno{\^{\i}}t Sagot},
  editor       = {Yvette Graham and
                  Matthew Purver},
  title        = {Anisotropy Is Inherent to Self-Attention in Transformers},
  booktitle    = {Proceedings of the 18th Conference of the European Chapter of the
                  Association for Computational Linguistics, {EACL} 2024 - Volume 1:
                  Long Papers, St. Julian's, Malta, March 17-22, 2024},
  pages        = {35--48},
  publisher    = {Association for Computational Linguistics},
  year         = {2024},
  url          = {https://aclanthology.org/2024.eacl-long.3},
  bibsource    = {dblp computer science bibliography, https://dblp.org}
}

@inproceedings{DBLP:conf/acl/GroeneveldBWBKT24,
  author       = {Dirk Groeneveld and
                  Iz Beltagy and
                  Evan Pete Walsh and
                  Akshita Bhagia and
                  Rodney Kinney and
                  Oyvind Tafjord and
                  Ananya Harsh Jha and
                  Hamish Ivison and
                  Ian Magnusson and
                  Yizhong Wang and
                  Shane Arora and
                  David Atkinson and
                  Russell Authur and
                  Khyathi Raghavi Chandu and
                  Arman Cohan and
                  Jennifer Dumas and
                  Yanai Elazar and
                  Yuling Gu and
                  Jack Hessel and
                  Tushar Khot and
                  William Merrill and
                  Jacob Morrison and
                  Niklas Muennighoff and
                  Aakanksha Naik and
                  Crystal Nam and
                  Matthew E. Peters and
                  Valentina Pyatkin and
                  Abhilasha Ravichander and
                  Dustin Schwenk and
                  Saurabh Shah and
                  Will Smith and
                  Emma Strubell and
                  Nishant Subramani and
                  Mitchell Wortsman and
                  Pradeep Dasigi and
                  Nathan Lambert and
                  Kyle Richardson and
                  Luke Zettlemoyer and
                  Jesse Dodge and
                  Kyle Lo and
                  Luca Soldaini and
                  Noah A. Smith and
                  Hannaneh Hajishirzi},
  editor       = {Lun{-}Wei Ku and
                  Andre Martins and
                  Vivek Srikumar},
  title        = {OLMo: Accelerating the Science of Language Models},
  booktitle    = {Proceedings of the 62nd Annual Meeting of the Association for Computational
                  Linguistics (Volume 1: Long Papers), {ACL} 2024, Bangkok, Thailand,
                  August 11-16, 2024},
  pages        = {15789--15809},
  publisher    = {Association for Computational Linguistics},
  year         = {2024},
  url          = {https://doi.org/10.18653/v1/2024.acl-long.841},
  doi          = {10.18653/V1/2024.ACL-LONG.841},
  bibsource    = {dblp computer science bibliography, https://dblp.org}
}

@inproceedings{DBLP:conf/nips/ZhangS19a,
  author       = {Biao Zhang and
                  Rico Sennrich},
  editor       = {Hanna M. Wallach and
                  Hugo Larochelle and
                  Alina Beygelzimer and
                  Florence d'Alch{\'{e}}{-}Buc and
                  Emily B. Fox and
                  Roman Garnett},
  title        = {Root Mean Square Layer Normalization},
  booktitle    = {Advances in Neural Information Processing Systems 32: Annual Conference
                  on Neural Information Processing Systems 2019, NeurIPS 2019, December
                  8-14, 2019, Vancouver, BC, Canada},
  pages        = {12360--12371},
  year         = {2019},
  url          = {https://proceedings.neurips.cc/paper/2019/hash/1e8a19426224ca89e83cef47f1e7f53b-Abstract.html},
  bibsource    = {dblp computer science bibliography, https://dblp.org}
}

@misc{DBLP:journals/corr/abs-2403-08295,
  title={Gemma: Open Models Based on Gemini Research and Technology}, 
      author={Gemma Team and Thomas Mesnard and Cassidy Hardin and Robert Dadashi and Surya Bhupatiraju and Shreya Pathak and Laurent Sifre and Morgane Rivière and Mihir Sanjay Kale and Juliette Love and Pouya Tafti and Léonard Hussenot and Pier Giuseppe Sessa and Aakanksha Chowdhery and Adam Roberts and Aditya Barua and Alex Botev and Alex Castro-Ros and Ambrose Slone and Amélie Héliou and Andrea Tacchetti and Anna Bulanova and Antonia Paterson and Beth Tsai and Bobak Shahriari and Charline Le Lan and Christopher A. Choquette-Choo and Clément Crepy and Daniel Cer and Daphne Ippolito and David Reid and Elena Buchatskaya and Eric Ni and Eric Noland and Geng Yan and George Tucker and George-Christian Muraru and Grigory Rozhdestvenskiy and Henryk Michalewski and Ian Tenney and Ivan Grishchenko and Jacob Austin and James Keeling and Jane Labanowski and Jean-Baptiste Lespiau and Jeff Stanway and Jenny Brennan and Jeremy Chen and Johan Ferret and Justin Chiu and Justin Mao-Jones and Katherine Lee and Kathy Yu and Katie Millican and Lars Lowe Sjoesund and Lisa Lee and Lucas Dixon and Machel Reid and Maciej Mikuła and Mateo Wirth and Michael Sharman and Nikolai Chinaev and Nithum Thain and Olivier Bachem and Oscar Chang and Oscar Wahltinez and Paige Bailey and Paul Michel and Petko Yotov and Rahma Chaabouni and Ramona Comanescu and Reena Jana and Rohan Anil and Ross McIlroy and Ruibo Liu and Ryan Mullins and Samuel L Smith and Sebastian Borgeaud and Sertan Girgin and Sholto Douglas and Shree Pandya and Siamak Shakeri and Soham De and Ted Klimenko and Tom Hennigan and Vlad Feinberg and Wojciech Stokowiec and Yu-hui Chen and Zafarali Ahmed and Zhitao Gong and Tris Warkentin and Ludovic Peran and Minh Giang and Clément Farabet and Oriol Vinyals and Jeff Dean and Koray Kavukcuoglu and Demis Hassabis and Zoubin Ghahramani and Douglas Eck and Joelle Barral and Fernando Pereira and Eli Collins and Armand Joulin and Noah Fiedel and Evan Senter and Alek Andreev and Kathleen Kenealy},
      year={2024},
      eprint={2403.08295},
      archivePrefix={arXiv},
      primaryClass={cs.CL},
      url={https://arxiv.org/abs/2403.08295}, 
}

@inproceedings{DBLP:conf/nips/ZhouXGM0W20,
  author       = {Wangchunshu Zhou and
                  Canwen Xu and
                  Tao Ge and
                  Julian J. McAuley and
                  Ke Xu and
                  Furu Wei},
  editor       = {Hugo Larochelle and
                  Marc'Aurelio Ranzato and
                  Raia Hadsell and
                  Maria{-}Florina Balcan and
                  Hsuan{-}Tien Lin},
  title        = {{BERT} Loses Patience: Fast and Robust Inference with Early Exit},
  booktitle    = {Advances in Neural Information Processing Systems 33: Annual Conference
                  on Neural Information Processing Systems 2020, NeurIPS 2020, December
                  6-12, 2020, virtual},
  year         = {2020},
  url          = {https://proceedings.neurips.cc/paper/2020/hash/d4dd111a4fd973394238aca5c05bebe3-Abstract.html},
  bibsource    = {dblp computer science bibliography, https://dblp.org}
}

@inproceedings{DBLP:conf/acl/XinTLYL20,
  author       = {Ji Xin and
                  Raphael Tang and
                  Jaejun Lee and
                  Yaoliang Yu and
                  Jimmy Lin},
  editor       = {Dan Jurafsky and
                  Joyce Chai and
                  Natalie Schluter and
                  Joel R. Tetreault},
  title        = {DeeBERT: Dynamic Early Exiting for Accelerating {BERT} Inference},
  booktitle    = {Proceedings of the 58th Annual Meeting of the Association for Computational
                  Linguistics, {ACL} 2020, Online, July 5-10, 2020},
  pages        = {2246--2251},
  publisher    = {Association for Computational Linguistics},
  year         = {2020},
  url          = {https://doi.org/10.18653/v1/2020.acl-main.204},
  doi          = {10.18653/V1/2020.ACL-MAIN.204},
  bibsource    = {dblp computer science bibliography, https://dblp.org}
}

@inproceedings{DBLP:conf/acl/LiuZWZDJ20,
  author       = {Weijie Liu and
                  Peng Zhou and
                  Zhiruo Wang and
                  Zhe Zhao and
                  Haotang Deng and
                  Qi Ju},
  editor       = {Dan Jurafsky and
                  Joyce Chai and
                  Natalie Schluter and
                  Joel R. Tetreault},
  title        = {FastBERT: a Self-distilling {BERT} with Adaptive Inference Time},
  booktitle    = {Proceedings of the 58th Annual Meeting of the Association for Computational
                  Linguistics, {ACL} 2020, Online, July 5-10, 2020},
  pages        = {6035--6044},
  publisher    = {Association for Computational Linguistics},
  year         = {2020},
  url          = {https://doi.org/10.18653/v1/2020.acl-main.537},
  doi          = {10.18653/V1/2020.ACL-MAIN.537},
  bibsource    = {dblp computer science bibliography, https://dblp.org}
}

@inproceedings{DBLP:conf/nips/SchusterFG0B0TM22,
  author       = {Tal Schuster and
                  Adam Fisch and
                  Jai Gupta and
                  Mostafa Dehghani and
                  Dara Bahri and
                  Vinh Tran and
                  Yi Tay and
                  Donald Metzler},
  editor       = {Sanmi Koyejo and
                  S. Mohamed and
                  A. Agarwal and
                  Danielle Belgrave and
                  K. Cho and
                  A. Oh},
  title        = {Confident Adaptive Language Modeling},
  booktitle    = {Advances in Neural Information Processing Systems 35: Annual Conference
                  on Neural Information Processing Systems 2022, NeurIPS 2022, New Orleans,
                  LA, USA, November 28 - December 9, 2022},
  year         = {2022},
  url          = {http://papers.nips.cc/paper\_files/paper/2022/hash/6fac9e316a4ae75ea244ddcef1982c71-Abstract-Conference.html},
  bibsource    = {dblp computer science bibliography, https://dblp.org}
}

@misc{javaheripi2023phi2,
  author       = {Javaheripi, Mojan and Bubeck, S{\'e}bastien and Abdin, Marah and Aneja, Jyoti and Mendes, Caio C{\'e}sar Teodoro and Chen, Weizhu and Del Giorno, Allie and Eldan, Ronen and Gopi, Sivakanth},
  title        = {{Phi-2}: The Surprising Power of Small Language Models},
  howpublished = {Microsoft Research Blog},
  month        = jul,
  year         = {2023},
  url          = {https://www.microsoft.com/research/blog/phi-2-the-surprising-power-of-small-language-models/}
}

@inproceedings{DBLP:conf/iclr/DehghaniGVUK19,
  author       = {Mostafa Dehghani and
                  Stephan Gouws and
                  Oriol Vinyals and
                  Jakob Uszkoreit and
                  Lukasz Kaiser},
  title        = {Universal Transformers},
  booktitle    = {7th International Conference on Learning Representations, {ICLR} 2019,
                  New Orleans, LA, USA, May 6-9, 2019},
  publisher    = {OpenReview.net},
  year         = {2019},
  url          = {https://openreview.net/forum?id=HyzdRiR9Y7},
  bibsource    = {dblp computer science bibliography, https://dblp.org}
}

@inproceedings{DBLP:conf/iclr/LanCGGSS20,
  author       = {Zhenzhong Lan and
                  Mingda Chen and
                  Sebastian Goodman and
                  Kevin Gimpel and
                  Piyush Sharma and
                  Radu Soricut},
  title        = {{ALBERT:} {A} Lite {BERT} for Self-supervised Learning of Language
                  Representations},
  booktitle    = {8th International Conference on Learning Representations, {ICLR} 2020,
                  Addis Ababa, Ethiopia, April 26-30, 2020},
  publisher    = {OpenReview.net},
  year         = {2020},
  url          = {https://openreview.net/forum?id=H1eA7AEtvS},
  bibsource    = {dblp computer science bibliography, https://dblp.org}
}

@article{DBLP:journals/csl/SajjadDDN23,
  author       = {Hassan Sajjad and
                  Fahim Dalvi and
                  Nadir Durrani and
                  Preslav Nakov},
  title        = {On the effect of dropping layers of pre-trained transformer models},
  journal      = {Comput. Speech Lang.},
  volume       = {77},
  pages        = {101429},
  year         = {2023},
  url          = {https://doi.org/10.1016/j.csl.2022.101429},
  doi          = {10.1016/J.CSL.2022.101429},
  bibsource    = {dblp computer science bibliography, https://dblp.org}
}

@inproceedings{DBLP:conf/iclr/JiangZZ25,
  author       = {Jiachen Jiang and
                  Jinxin Zhou and
                  Zhihui Zhu},
  title        = {Tracing Representation Progression: Analyzing and Enhancing Layer-Wise
                  Similarity},
  booktitle    = {The Thirteenth International Conference on Learning Representations,
                  {ICLR} 2025, Singapore, April 24-28, 2025},
  publisher    = {OpenReview.net},
  year         = {2025},
  url          = {https://openreview.net/forum?id=vVxeFSR4fU},
  bibsource    = {dblp computer science bibliography, https://dblp.org}
}

@misc{achiam2023gpt,
  title={Gpt-4 technical report},
  author={Achiam, Josh and Adler, Steven and Agarwal, Sandhini and Ahmad, Lama and Akkaya, Ilge and Aleman, Florencia Leoni and Almeida, Diogo and Altenschmidt, Janko and Altman, Sam and Anadkat, Shyamal and others},
  journal={arXiv preprint arXiv:2303.08774},
  year={2023}
}

@misc{grattafiori2024llama,
  title={The Llama 3 Herd of Models}, 
      author={Aaron Grattafiori and Abhimanyu Dubey and Abhinav Jauhri and Abhinav Pandey and Abhishek Kadian and Ahmad Al-Dahle and Aiesha Letman and Akhil Mathur and Alan Schelten and Alex Vaughan and Amy Yang and Angela Fan and Anirudh Goyal and Anthony Hartshorn and Aobo Yang and Archi Mitra and Archie Sravankumar and Artem Korenev and Arthur Hinsvark and Arun Rao and Aston Zhang and Aurelien Rodriguez and Austen Gregerson and Ava Spataru and Baptiste Roziere and Bethany Biron and Binh Tang and Bobbie Chern and Charlotte Caucheteux and Chaya Nayak and Chloe Bi and Chris Marra and Chris McConnell and Christian Keller and Christophe Touret and Chunyang Wu and Corinne Wong and Cristian Canton Ferrer and Cyrus Nikolaidis and Damien Allonsius and Daniel Song and Danielle Pintz and Danny Livshits and Danny Wyatt and David Esiobu and Dhruv Choudhary and Dhruv Mahajan and Diego Garcia-Olano and Diego Perino and Dieuwke Hupkes and Egor Lakomkin and Ehab AlBadawy and Elina Lobanova and Emily Dinan and Eric Michael Smith and Filip Radenovic and Francisco Guzmán and Frank Zhang and Gabriel Synnaeve and Gabrielle Lee and Georgia Lewis Anderson and Govind Thattai and Graeme Nail and Gregoire Mialon and Guan Pang and Guillem Cucurell and Hailey Nguyen and Hannah Korevaar and Hu Xu and Hugo Touvron and Iliyan Zarov and Imanol Arrieta Ibarra and Isabel Kloumann and Ishan Misra and Ivan Evtimov and Jack Zhang and Jade Copet and Jaewon Lee and Jan Geffert and Jana Vranes and Jason Park and Jay Mahadeokar and Jeet Shah and Jelmer van der Linde and Jennifer Billock and Jenny Hong and Jenya Lee and Jeremy Fu and Jianfeng Chi and Jianyu Huang and Jiawen Liu and Jie Wang and Jiecao Yu and Joanna Bitton and Joe Spisak and Jongsoo Park and Joseph Rocca and Joshua Johnstun and Joshua Saxe and Junteng Jia and Kalyan Vasuden Alwala and Karthik Prasad and Kartikeya Upasani and Kate Plawiak and Ke Li and Kenneth Heafield and Kevin Stone and Khalid El-Arini and Krithika Iyer and Kshitiz Malik and Kuenley Chiu and Kunal Bhalla and Kushal Lakhotia and Lauren Rantala-Yeary and Laurens van der Maaten and Lawrence Chen and Liang Tan and Liz Jenkins and Louis Martin and Lovish Madaan and Lubo Malo and Lukas Blecher and Lukas Landzaat and Luke de Oliveira and Madeline Muzzi and Mahesh Pasupuleti and Mannat Singh and Manohar Paluri and Marcin Kardas and Maria Tsimpoukelli and Mathew Oldham and Mathieu Rita and Maya Pavlova and Melanie Kambadur and Mike Lewis and Min Si and Mitesh Kumar Singh and Mona Hassan and Naman Goyal and Narjes Torabi and Nikolay Bashlykov and Nikolay Bogoychev and Niladri Chatterji and Ning Zhang and Olivier Duchenne and Onur Çelebi and Patrick Alrassy and Pengchuan Zhang and Pengwei Li and Petar Vasic and Peter Weng and Prajjwal Bhargava and Pratik Dubal and Praveen Krishnan and Punit Singh Koura and Puxin Xu and Qing He and Qingxiao Dong and Ragavan Srinivasan and Raj Ganapathy and Ramon Calderer and Ricardo Silveira Cabral and Robert Stojnic and Roberta Raileanu and Rohan Maheswari and Rohit Girdhar and Rohit Patel and Romain Sauvestre and Ronnie Polidoro and Roshan Sumbaly and Ross Taylor and Ruan Silva and Rui Hou and Rui Wang and Saghar Hosseini and Sahana Chennabasappa and Sanjay Singh and Sean Bell and Seohyun Sonia Kim and Sergey Edunov and Shaoliang Nie and Sharan Narang and Sharath Raparthy and Sheng Shen and Shengye Wan and Shruti Bhosale and Shun Zhang and Simon Vandenhende and Soumya Batra and Spencer Whitman and Sten Sootla and Stephane Collot and Suchin Gururangan and Sydney Borodinsky and Tamar Herman and Tara Fowler and Tarek Sheasha and Thomas Georgiou and Thomas Scialom and Tobias Speckbacher and Todor Mihaylov and Tong Xiao and Ujjwal Karn and Vedanuj Goswami and Vibhor Gupta and Vignesh Ramanathan and Viktor Kerkez and Vincent Gonguet and Virginie Do and Vish Vogeti and Vítor Albiero and Vladan Petrovic and Weiwei Chu and Wenhan Xiong and Wenyin Fu and Whitney Meers and Xavier Martinet and Xiaodong Wang and Xiaofang Wang and Xiaoqing Ellen Tan and Xide Xia and Xinfeng Xie and Xuchao Jia and Xuewei Wang and Yaelle Goldschlag and Yashesh Gaur and Yasmine Babaei and Yi Wen and Yiwen Song and Yuchen Zhang and Yue Li and Yuning Mao and Zacharie Delpierre Coudert and Zheng Yan and Zhengxing Chen and Zoe Papakipos and Aaditya Singh and Aayushi Srivastava and Abha Jain and Adam Kelsey and Adam Shajnfeld and Adithya Gangidi and Adolfo Victoria and Ahuva Goldstand and Ajay Menon and Ajay Sharma and Alex Boesenberg and Alexei Baevski and Allie Feinstein and Amanda Kallet and Amit Sangani and Amos Teo and Anam Yunus and Andrei Lupu and Andres Alvarado and Andrew Caples and Andrew Gu and Andrew Ho and Andrew Poulton and Andrew Ryan and Ankit Ramchandani and Annie Dong and Annie Franco and Anuj Goyal and Aparajita Saraf and Arkabandhu Chowdhury and Ashley Gabriel and Ashwin Bharambe and Assaf Eisenman and Azadeh Yazdan and Beau James and Ben Maurer and Benjamin Leonhardi and Bernie Huang and Beth Loyd and Beto De Paola and Bhargavi Paranjape and Bing Liu and Bo Wu and Boyu Ni and Braden Hancock and Bram Wasti and Brandon Spence and Brani Stojkovic and Brian Gamido and Britt Montalvo and Carl Parker and Carly Burton and Catalina Mejia and Ce Liu and Changhan Wang and Changkyu Kim and Chao Zhou and Chester Hu and Ching-Hsiang Chu and Chris Cai and Chris Tindal and Christoph Feichtenhofer and Cynthia Gao and Damon Civin and Dana Beaty and Daniel Kreymer and Daniel Li and David Adkins and David Xu and Davide Testuggine and Delia David and Devi Parikh and Diana Liskovich and Didem Foss and Dingkang Wang and Duc Le and Dustin Holland and Edward Dowling and Eissa Jamil and Elaine Montgomery and Eleonora Presani and Emily Hahn and Emily Wood and Eric-Tuan Le and Erik Brinkman and Esteban Arcaute and Evan Dunbar and Evan Smothers and Fei Sun and Felix Kreuk and Feng Tian and Filippos Kokkinos and Firat Ozgenel and Francesco Caggioni and Frank Kanayet and Frank Seide and Gabriela Medina Florez and Gabriella Schwarz and Gada Badeer and Georgia Swee and Gil Halpern and Grant Herman and Grigory Sizov and Guangyi and Zhang and Guna Lakshminarayanan and Hakan Inan and Hamid Shojanazeri and Han Zou and Hannah Wang and Hanwen Zha and Haroun Habeeb and Harrison Rudolph and Helen Suk and Henry Aspegren and Hunter Goldman and Hongyuan Zhan and Ibrahim Damlaj and Igor Molybog and Igor Tufanov and Ilias Leontiadis and Irina-Elena Veliche and Itai Gat and Jake Weissman and James Geboski and James Kohli and Janice Lam and Japhet Asher and Jean-Baptiste Gaya and Jeff Marcus and Jeff Tang and Jennifer Chan and Jenny Zhen and Jeremy Reizenstein and Jeremy Teboul and Jessica Zhong and Jian Jin and Jingyi Yang and Joe Cummings and Jon Carvill and Jon Shepard and Jonathan McPhie and Jonathan Torres and Josh Ginsburg and Junjie Wang and Kai Wu and Kam Hou U and Karan Saxena and Kartikay Khandelwal and Katayoun Zand and Kathy Matosich and Kaushik Veeraraghavan and Kelly Michelena and Keqian Li and Kiran Jagadeesh and Kun Huang and Kunal Chawla and Kyle Huang and Lailin Chen and Lakshya Garg and Lavender A and Leandro Silva and Lee Bell and Lei Zhang and Liangpeng Guo and Licheng Yu and Liron Moshkovich and Luca Wehrstedt and Madian Khabsa and Manav Avalani and Manish Bhatt and Martynas Mankus and Matan Hasson and Matthew Lennie and Matthias Reso and Maxim Groshev and Maxim Naumov and Maya Lathi and Meghan Keneally and Miao Liu and Michael L. Seltzer and Michal Valko and Michelle Restrepo and Mihir Patel and Mik Vyatskov and Mikayel Samvelyan and Mike Clark and Mike Macey and Mike Wang and Miquel Jubert Hermoso and Mo Metanat and Mohammad Rastegari and Munish Bansal and Nandhini Santhanam and Natascha Parks and Natasha White and Navyata Bawa and Nayan Singhal and Nick Egebo and Nicolas Usunier and Nikhil Mehta and Nikolay Pavlovich Laptev and Ning Dong and Norman Cheng and Oleg Chernoguz and Olivia Hart and Omkar Salpekar and Ozlem Kalinli and Parkin Kent and Parth Parekh and Paul Saab and Pavan Balaji and Pedro Rittner and Philip Bontrager and Pierre Roux and Piotr Dollar and Polina Zvyagina and Prashant Ratanchandani and Pritish Yuvraj and Qian Liang and Rachad Alao and Rachel Rodriguez and Rafi Ayub and Raghotham Murthy and Raghu Nayani and Rahul Mitra and Rangaprabhu Parthasarathy and Raymond Li and Rebekkah Hogan and Robin Battey and Rocky Wang and Russ Howes and Ruty Rinott and Sachin Mehta and Sachin Siby and Sai Jayesh Bondu and Samyak Datta and Sara Chugh and Sara Hunt and Sargun Dhillon and Sasha Sidorov and Satadru Pan and Saurabh Mahajan and Saurabh Verma and Seiji Yamamoto and Sharadh Ramaswamy and Shaun Lindsay and Shaun Lindsay and Sheng Feng and Shenghao Lin and Shengxin Cindy Zha and Shishir Patil and Shiva Shankar and Shuqiang Zhang and Shuqiang Zhang and Sinong Wang and Sneha Agarwal and Soji Sajuyigbe and Soumith Chintala and Stephanie Max and Stephen Chen and Steve Kehoe and Steve Satterfield and Sudarshan Govindaprasad and Sumit Gupta and Summer Deng and Sungmin Cho and Sunny Virk and Suraj Subramanian and Sy Choudhury and Sydney Goldman and Tal Remez and Tamar Glaser and Tamara Best and Thilo Koehler and Thomas Robinson and Tianhe Li and Tianjun Zhang and Tim Matthews and Timothy Chou and Tzook Shaked and Varun Vontimitta and Victoria Ajayi and Victoria Montanez and Vijai Mohan and Vinay Satish Kumar and Vishal Mangla and Vlad Ionescu and Vlad Poenaru and Vlad Tiberiu Mihailescu and Vladimir Ivanov and Wei Li and Wenchen Wang and Wenwen Jiang and Wes Bouaziz and Will Constable and Xiaocheng Tang and Xiaojian Wu and Xiaolan Wang and Xilun Wu and Xinbo Gao and Yaniv Kleinman and Yanjun Chen and Ye Hu and Ye Jia and Ye Qi and Yenda Li and Yilin Zhang and Ying Zhang and Yossi Adi and Youngjin Nam and Yu and Wang and Yu Zhao and Yuchen Hao and Yundi Qian and Yunlu Li and Yuzi He and Zach Rait and Zachary DeVito and Zef Rosnbrick and Zhaoduo Wen and Zhenyu Yang and Zhiwei Zhao and Zhiyu Ma},
      year={2024},
      eprint={2407.21783},
      archivePrefix={arXiv},
      primaryClass={cs.AI},
      url={https://arxiv.org/abs/2407.21783}, 
}

@misc{gao-2024-eval-harness,
    author = {Gao, Leo and Tow, Jonathan and Abbasi, Baber and Biderman, Stella and Black, Sid and DiPofi, Anthony and Foster, Charles and Golding, Laurence and Hsu, Jeffrey and Le Noac'h, Alain and Li, Haonan and McDonell, Kyle and Muennighoff, Niklas and Ociepa, Chris and Phang, Jason and Reynolds, Laria and Schoelkopf, Hailey and Skowron, Aviya and Sutawika, Lintang and Tang, Eric and Thite, Anish and Wang, Ben and Wang, Kevin and Zou, Andy},
    title = {The Language Model Evaluation Harness},
    year = {2024},
    publisher = {Zenodo},
    version = {v0.4.8},
    url = {https://zenodo.org/records/14970487}
}

@misc{DBLP:journals/corr/abs-2401-02954,
  title={DeepSeek LLM: Scaling Open-Source Language Models with Longtermism}, 
      author={DeepSeek-AI and : and Xiao Bi and Deli Chen and Guanting Chen and Shanhuang Chen and Damai Dai and Chengqi Deng and Honghui Ding and Kai Dong and Qiushi Du and Zhe Fu and Huazuo Gao and Kaige Gao and Wenjun Gao and Ruiqi Ge and Kang Guan and Daya Guo and Jianzhong Guo and Guangbo Hao and Zhewen Hao and Ying He and Wenjie Hu and Panpan Huang and Erhang Li and Guowei Li and Jiashi Li and Yao Li and Y. K. Li and Wenfeng Liang and Fangyun Lin and A. X. Liu and Bo Liu and Wen Liu and Xiaodong Liu and Xin Liu and Yiyuan Liu and Haoyu Lu and Shanghao Lu and Fuli Luo and Shirong Ma and Xiaotao Nie and Tian Pei and Yishi Piao and Junjie Qiu and Hui Qu and Tongzheng Ren and Zehui Ren and Chong Ruan and Zhangli Sha and Zhihong Shao and Junxiao Song and Xuecheng Su and Jingxiang Sun and Yaofeng Sun and Minghui Tang and Bingxuan Wang and Peiyi Wang and Shiyu Wang and Yaohui Wang and Yongji Wang and Tong Wu and Y. Wu and Xin Xie and Zhenda Xie and Ziwei Xie and Yiliang Xiong and Hanwei Xu and R. X. Xu and Yanhong Xu and Dejian Yang and Yuxiang You and Shuiping Yu and Xingkai Yu and B. Zhang and Haowei Zhang and Lecong Zhang and Liyue Zhang and Mingchuan Zhang and Minghua Zhang and Wentao Zhang and Yichao Zhang and Chenggang Zhao and Yao Zhao and Shangyan Zhou and Shunfeng Zhou and Qihao Zhu and Yuheng Zou},
      year={2024},
      eprint={2401.02954},
      archivePrefix={arXiv},
      primaryClass={cs.CL},
      url={https://arxiv.org/abs/2401.02954}, 
}

@misc{yang2025qwen3technicalreport,
      title={Qwen3 Technical Report}, 
      author={An Yang and Anfeng Li and Baosong Yang and Beichen Zhang and Binyuan Hui and Bo Zheng and Bowen Yu and Chang Gao and Chengen Huang and Chenxu Lv and Chujie Zheng and Dayiheng Liu and Fan Zhou and Fei Huang and Feng Hu and Hao Ge and Haoran Wei and Huan Lin and Jialong Tang and Jian Yang and Jianhong Tu and Jianwei Zhang and Jianxin Yang and Jiaxi Yang and Jing Zhou and Jingren Zhou and Junyang Lin and Kai Dang and Keqin Bao and Kexin Yang and Le Yu and Lianghao Deng and Mei Li and Mingfeng Xue and Mingze Li and Pei Zhang and Peng Wang and Qin Zhu and Rui Men and Ruize Gao and Shixuan Liu and Shuang Luo and Tianhao Li and Tianyi Tang and Wenbiao Yin and Xingzhang Ren and Xinyu Wang and Xinyu Zhang and Xuancheng Ren and Yang Fan and Yang Su and Yichang Zhang and Yinger Zhang and Yu Wan and Yuqiong Liu and Zekun Wang and Zeyu Cui and Zhenru Zhang and Zhipeng Zhou and Zihan Qiu},
      year={2025},
      eprint={2505.09388},
      archivePrefix={arXiv},
      primaryClass={cs.CL},
      url={https://arxiv.org/abs/2505.09388}, 
}

@inproceedings{DBLP:conf/acl/ZellersHBFC19,
  author       = {Rowan Zellers and
                  Ari Holtzman and
                  Yonatan Bisk and
                  Ali Farhadi and
                  Yejin Choi},
  editor       = {Anna Korhonen and
                  David R. Traum and
                  Llu{\'{\i}}s M{\`{a}}rquez},
  title        = {HellaSwag: Can a Machine Really Finish Your Sentence?},
  booktitle    = {Proceedings of the 57th Conference of the Association for Computational
                  Linguistics, {ACL} 2019, Florence, Italy, July 28- August 2, 2019,
                  Volume 1: Long Papers},
  pages        = {4791--4800},
  publisher    = {Association for Computational Linguistics},
  year         = {2019},
  url          = {https://doi.org/10.18653/v1/p19-1472},
  doi          = {10.18653/V1/P19-1472},
  bibsource    = {dblp computer science bibliography, https://dblp.org}
}

@inproceedings{DBLP:conf/aaai/BiskZLGC20,
  author       = {Yonatan Bisk and
                  Rowan Zellers and
                  Ronan Le Bras and
                  Jianfeng Gao and
                  Yejin Choi},
  title        = {{PIQA:} Reasoning about Physical Commonsense in Natural Language},
  booktitle    = {The Thirty-Fourth {AAAI} Conference on Artificial Intelligence, {AAAI}
                  2020, The Thirty-Second Innovative Applications of Artificial Intelligence
                  Conference, {IAAI} 2020, The Tenth {AAAI} Symposium on Educational
                  Advances in Artificial Intelligence, {EAAI} 2020, New York, NY, USA,
                  February 7-12, 2020},
  pages        = {7432--7439},
  publisher    = {{AAAI} Press},
  year         = {2020},
  url          = {https://doi.org/10.1609/aaai.v34i05.6239},
  doi          = {10.1609/AAAI.V34I05.6239},
  bibsource    = {dblp computer science bibliography, https://dblp.org}
}

@misc{belrose2023eliciting,
  title={Eliciting Latent Predictions from Transformers with the Tuned Lens}, 
      author={Nora Belrose and Zach Furman and Logan Smith and Danny Halawi and Igor Ostrovsky and Lev McKinney and Stella Biderman and Jacob Steinhardt},
      year={2023},
      eprint={2303.08112},
      archivePrefix={arXiv},
      primaryClass={cs.LG},
      url={https://arxiv.org/abs/2303.08112}, 
}

@inproceedings{
kobayashi2024analyzing,
title={Analyzing Feed-Forward Blocks in Transformers through the Lens of Attention Maps},
author={Goro Kobayashi and Tatsuki Kuribayashi and Sho Yokoi and Kentaro Inui},
booktitle={The Twelfth International Conference on Learning Representations},
year={2024},
url={https://openreview.net/forum?id=mYWsyTuiRp}
}

@inproceedings{
tigges2024llm,
title={{LLM} Circuit Analyses Are Consistent Across Training and Scale},
author={Curt Tigges and Michael Hanna and Qinan Yu and Stella Biderman},
booktitle={The Thirty-eighth Annual Conference on Neural Information Processing Systems},
year={2024},
url={https://openreview.net/forum?id=3Ds5vNudIE}
}

@inproceedings{
stolfo2024confidence,
title={Confidence Regulation Neurons in Language Models},
author={Alessandro Stolfo and Ben Peng Wu and Wes Gurnee and Yonatan Belinkov and Xingyi Song and Mrinmaya Sachan and Neel Nanda},
booktitle={The Thirty-eighth Annual Conference on Neural Information Processing Systems},
year={2024},
url={https://openreview.net/forum?id=0og7nmvDbe}
}

@inproceedings{
song2024unraveling,
title={Unraveling the Gradient Descent Dynamics of Transformers},
author={Bingqing Song and Boran Han and Shuai Zhang and Jie Ding and Mingyi Hong},
booktitle={The Thirty-eighth Annual Conference on Neural Information Processing Systems},
year={2024},
url={https://openreview.net/forum?id=XswQeLjJo5}
}

@inproceedings{aghajanyan-etal-2021-intrinsic,
    title = "Intrinsic Dimensionality Explains the Effectiveness of Language Model Fine-Tuning",
    author = "Aghajanyan, Armen  and
      Gupta, Sonal  and
      Zettlemoyer, Luke",
    editor = "Zong, Chengqing  and
      Xia, Fei  and
      Li, Wenjie  and
      Navigli, Roberto",
    booktitle = "Proceedings of the 59th Annual Meeting of the Association for Computational Linguistics and the 11th International Joint Conference on Natural Language Processing (Volume 1: Long Papers)",
    month = aug,
    year = "2021",
    address = "Online",
    publisher = "Association for Computational Linguistics",
    url = "https://aclanthology.org/2021.acl-long.568/",
    doi = "10.18653/v1/2021.acl-long.568",
    pages = "7319--7328"
}

@inproceedings{DBLP:journals/corr/abs-2502-10341,
  title={Organize the Web: Constructing Domains Enhances Pre-Training Data Curation},
author={Alexander Wettig and Kyle Lo and Sewon Min and Hannaneh Hajishirzi and Danqi Chen and Luca Soldaini},
booktitle={Forty-second International Conference on Machine Learning},
year={2025},
url={https://openreview.net/forum?id=boSqwdvJVC}
}

@inproceedings{DBLP:conf/aclnut/WelblLG17,
  author       = {Johannes Welbl and
                  Nelson F. Liu and
                  Matt Gardner},
  editor       = {Leon Derczynski and
                  Wei Xu and
                  Alan Ritter and
                  Tim Baldwin},
  title        = {Crowdsourcing Multiple Choice Science Questions},
  booktitle    = {Proceedings of the 3rd Workshop on Noisy User-generated Text, NUT@EMNLP
                  2017, Copenhagen, Denmark, September 7, 2017},
  pages        = {94--106},
  publisher    = {Association for Computational Linguistics},
  year         = {2017},
  url          = {https://doi.org/10.18653/v1/w17-4413},
  doi          = {10.18653/V1/W17-4413},
  bibsource    = {dblp computer science bibliography, https://dblp.org}
}

@inproceedings{DBLP:conf/emnlp/ZellersBSC18,
  author       = {Rowan Zellers and
                  Yonatan Bisk and
                  Roy Schwartz and
                  Yejin Choi},
  editor       = {Ellen Riloff and
                  David Chiang and
                  Julia Hockenmaier and
                  Jun'ichi Tsujii},
  title        = {{SWAG:} {A} Large-Scale Adversarial Dataset for Grounded Commonsense
                  Inference},
  booktitle    = {Proceedings of the 2018 Conference on Empirical Methods in Natural
                  Language Processing, Brussels, Belgium, October 31 - November 4, 2018},
  pages        = {93--104},
  publisher    = {Association for Computational Linguistics},
  year         = {2018},
  url          = {https://doi.org/10.18653/v1/d18-1009},
  doi          = {10.18653/V1/D18-1009},
  bibsource    = {dblp computer science bibliography, https://dblp.org}
}

@inproceedings{DBLP:conf/acl/RazzhigaevMGGOD24,
  author       = {Anton Razzhigaev and
                  Matvey Mikhalchuk and
                  Elizaveta Goncharova and
                  Nikolai Gerasimenko and
                  Ivan V. Oseledets and
                  Denis Dimitrov and
                  Andrey Kuznetsov},
  editor       = {Lun{-}Wei Ku and
                  Andre Martins and
                  Vivek Srikumar},
  title        = {Your Transformer is Secretly Linear},
  booktitle    = {Proceedings of the 62nd Annual Meeting of the Association for Computational
                  Linguistics (Volume 1: Long Papers), {ACL} 2024, Bangkok, Thailand,
                  August 11-16, 2024},
  pages        = {5376--5384},
  publisher    = {Association for Computational Linguistics},
  year         = {2024},
  url          = {https://doi.org/10.18653/v1/2024.acl-long.293},
  doi          = {10.18653/V1/2024.ACL-LONG.293},
  bibsource    = {dblp computer science bibliography, https://dblp.org}
}

@inproceedings{DBLP:conf/coling/DasJSMPY25,
  author       = {Souvik Das and
                  Lifeng Jin and
                  Linfeng Song and
                  Haitao Mi and
                  Baolin Peng and
                  Dong Yu},
  editor       = {Owen Rambow and
                  Leo Wanner and
                  Marianna Apidianaki and
                  Hend Al{-}Khalifa and
                  Barbara Di Eugenio and
                  Steven Schockaert},
  title        = {Entropy Guided Extrapolative Decoding to Improve Factuality in Large
                  Language Models},
  booktitle    = {Proceedings of the 31st International Conference on Computational
                  Linguistics, {COLING} 2025, Abu Dhabi, UAE, January 19-24, 2025},
  pages        = {6589--6600},
  publisher    = {Association for Computational Linguistics},
  year         = {2025},
  url          = {https://aclanthology.org/2025.coling-main.439/},
  bibsource    = {dblp computer science bibliography, https://dblp.org}
}

@inproceedings{DBLP:journals/corr/abs-2411-12768,
  title={{CROW}: Eliminating Backdoors from Large Language Models via Internal Consistency Regularization},
author={Nay Myat Min and Long H. Pham and Yige Li and Jun Sun},
booktitle={Forty-second International Conference on Machine Learning},
year={2025},
url={https://openreview.net/forum?id=ZGtcgeCpWB}
}

@inproceedings{DBLP:conf/nips/WangIDHKCWMSBH23,
  author       = {Yizhong Wang and
                  Hamish Ivison and
                  Pradeep Dasigi and
                  Jack Hessel and
                  Tushar Khot and
                  Khyathi Raghavi Chandu and
                  David Wadden and
                  Kelsey MacMillan and
                  Noah A. Smith and
                  Iz Beltagy and
                  Hannaneh Hajishirzi},
  editor       = {Alice Oh and
                  Tristan Naumann and
                  Amir Globerson and
                  Kate Saenko and
                  Moritz Hardt and
                  Sergey Levine},
  title        = {How Far Can Camels Go? Exploring the State of Instruction Tuning on
                  Open Resources},
  booktitle    = {Advances in Neural Information Processing Systems 36: Annual Conference
                  on Neural Information Processing Systems 2023, NeurIPS 2023, New Orleans,
                  LA, USA, December 10 - 16, 2023},
  year         = {2023},
  url          = {http://papers.nips.cc/paper\_files/paper/2023/hash/ec6413875e4ab08d7bc4d8e225263398-Abstract-Datasets\_and\_Benchmarks.html},
  bibsource    = {dblp computer science bibliography, https://dblp.org}
}

@inproceedings{DBLP:conf/nips/ZhouTQYJX0024,
  author       = {Hang Zhou and
                  Yehui Tang and
                  Haochen Qin and
                  Yujie Yang and
                  Renren Jin and
                  Deyi Xiong and
                  Kai Han and
                  Yunhe Wang},
  editor       = {Amir Globersons and
                  Lester Mackey and
                  Danielle Belgrave and
                  Angela Fan and
                  Ulrich Paquet and
                  Jakub M. Tomczak and
                  Cheng Zhang},
  title        = {Star-Agents: Automatic Data Optimization with {LLM} Agents for Instruction
                  Tuning},
  booktitle    = {Advances in Neural Information Processing Systems 38: Annual Conference
                  on Neural Information Processing Systems 2024, NeurIPS 2024, Vancouver,
                  BC, Canada, December 10 - 15, 2024},
  year         = {2024},
  url          = {http://papers.nips.cc/paper\_files/paper/2024/hash/0852b88e96d973bd4e21b673f51621d0-Abstract-Conference.html},
  bibsource    = {dblp computer science bibliography, https://dblp.org}
}

@inproceedings{DBLP:conf/nips/ZhengC00WZL0LXZ23,
  author       = {Lianmin Zheng and
                  Wei{-}Lin Chiang and
                  Ying Sheng and
                  Siyuan Zhuang and
                  Zhanghao Wu and
                  Yonghao Zhuang and
                  Zi Lin and
                  Zhuohan Li and
                  Dacheng Li and
                  Eric P. Xing and
                  Hao Zhang and
                  Joseph E. Gonzalez and
                  Ion Stoica},
  editor       = {Alice Oh and
                  Tristan Naumann and
                  Amir Globerson and
                  Kate Saenko and
                  Moritz Hardt and
                  Sergey Levine},
  title        = {Judging LLM-as-a-Judge with MT-Bench and Chatbot Arena},
  booktitle    = {Advances in Neural Information Processing Systems 36: Annual Conference
                  on Neural Information Processing Systems 2023, NeurIPS 2023, New Orleans,
                  LA, USA, December 10 - 16, 2023},
  year         = {2023},
  url          = {http://papers.nips.cc/paper\_files/paper/2023/hash/91f18a1287b398d378ef22505bf41832-Abstract-Datasets\_and\_Benchmarks.html},
  bibsource    = {dblp computer science bibliography, https://dblp.org}
}

@inproceedings{DBLP:conf/iclr/XuSZG0FTLJ24,
  author       = {Can Xu and
                  Qingfeng Sun and
                  Kai Zheng and
                  Xiubo Geng and
                  Pu Zhao and
                  Jiazhan Feng and
                  Chongyang Tao and
                  Qingwei Lin and
                  Daxin Jiang},
  title        = {WizardLM: Empowering Large Pre-Trained Language Models to Follow Complex
                  Instructions},
  booktitle    = {The Twelfth International Conference on Learning Representations,
                  {ICLR} 2024, Vienna, Austria, May 7-11, 2024},
  publisher    = {OpenReview.net},
  year         = {2024},
  url          = {https://openreview.net/forum?id=CfXh93NDgH},
  bibsource    = {dblp computer science bibliography, https://dblp.org}
}

@inproceedings{DBLP:conf/emnlp/WolfDSCDMCRLFDS20,
  author       = {Thomas Wolf and
                  Lysandre Debut and
                  Victor Sanh and
                  Julien Chaumond and
                  Clement Delangue and
                  Anthony Moi and
                  Pierric Cistac and
                  Tim Rault and
                  R{\'{e}}mi Louf and
                  Morgan Funtowicz and
                  Joe Davison and
                  Sam Shleifer and
                  Patrick von Platen and
                  Clara Ma and
                  Yacine Jernite and
                  Julien Plu and
                  Canwen Xu and
                  Teven Le Scao and
                  Sylvain Gugger and
                  Mariama Drame and
                  Quentin Lhoest and
                  Alexander M. Rush},
  editor       = {Qun Liu and
                  David Schlangen},
  title        = {Transformers: State-of-the-Art Natural Language Processing},
  booktitle    = {Proceedings of the 2020 Conference on Empirical Methods in Natural
                  Language Processing: System Demonstrations, {EMNLP} 2020 - Demos,
                  Online, November 16-20, 2020},
  pages        = {38--45},
  publisher    = {Association for Computational Linguistics},
  year         = {2020},
  url          = {https://doi.org/10.18653/v1/2020.emnlp-demos.6},
  doi          = {10.18653/V1/2020.EMNLP-DEMOS.6},
  bibsource    = {dblp computer science bibliography, https://dblp.org}
}

@inproceedings{DBLP:conf/icml/KediaZKJGL24,
  author       = {Akhil Kedia and
                  Mohd Abbas Zaidi and
                  Sushil Khyalia and
                  Jungho Jung and
                  Harshith Goka and
                  Haejun Lee},
  title        = {Transformers Get Stable: An End-to-End Signal Propagation Theory for
                  Language Models},
  booktitle    = {Forty-first International Conference on Machine Learning, {ICML} 2024,
                  Vienna, Austria, July 21-27, 2024},
  publisher    = {OpenReview.net},
  year         = {2024},
  url          = {https://openreview.net/forum?id=30waYPIZUA},
  bibsource    = {dblp computer science bibliography, https://dblp.org}
}

@article{DBLP:journals/corr/abs-2304-14802,
  author       = {Shufang Xie and
                  Huishuai Zhang and
                  Junliang Guo and
                  Xu Tan and
                  Jiang Bian and
                  Hany Hassan Awadalla and
                  Arul Menezes and
                  Tao Qin and
                  Rui Yan},
  title        = {ResiDual: Transformer with Dual Residual Connections},
  journal      = {CoRR},
  volume       = {abs/2304.14802},
  year         = {2023},
  url          = {https://doi.org/10.48550/arXiv.2304.14802},
  doi          = {10.48550/ARXIV.2304.14802},
  eprinttype    = {arXiv},
  eprint       = {2304.14802},
  bibsource    = {dblp computer science bibliography, https://dblp.org}
}
